\documentclass[10pt]{article} % For LaTeX2e
\usepackage[accepted]{tmlr}
\usepackage{amsmath,amsfonts,bm}

\def\eqref#1{equation~\ref{#1}}
\def\1{\bm{1}}

\DeclareMathAlphabet{\mathsfit}{\encodingdefault}{\sfdefault}{m}{sl}
\SetMathAlphabet{\mathsfit}{bold}{\encodingdefault}{\sfdefault}{bx}{n}

\usepackage{hyperref}
\usepackage{url}
\usepackage{subfiles}
\usepackage{graphicx}
\usepackage{subfigure}
\usepackage{comment}
\usepackage{booktabs}
\usepackage{float}
\usepackage{algorithm}
\usepackage{algorithmicx}
\usepackage{algpseudocode}
\usepackage{graphbox}
\usepackage{color}
\usepackage{caption}
\usepackage{soul}
\usepackage{color, xcolor}
\setstcolor{red}
\usepackage{ulem}
\title{Revisiting Discrete Soft Actor-Critic}

\author{\name Haibin Zhou \email haibinzhou@tencent.com \\
      \addr Tencent Inc. 
      \AND
        \name Tong Wei \email wt22@mails.tsinghua.edu.cn \\
      \addr Tsinghua University
      \AND
        \name Zichuan Lin \email zichuanlin@tencent.com \\
      \addr Tencent Inc.
      \AND
        \name Junyou Li \email junyouli@tencent.com \\
      \addr Tencent Inc.
      \AND
      \name Junliang Xing \email jlxing@tsinghua.edu.cn \\
      \addr Tsinghua University
      \AND
      \name Yuanchun Shi \email shiyc@tsinghua.edu.cn \\
      \addr Tsinghua University
      \AND
      \name Li Shen \email  mathshenli@gmail.com\\
      \addr Sun Yat-sen University
      \AND
        \name Chao Yu \email yuchao3@mail.sysu.edu.cn \\
      \addr Sun Yat-sen University
      \AND
        \name Deheng Ye\thanks{Corresponding author.} 
        \email dericye@tencent.com \\
      \addr Tencent
    }
\def\month{11}  % Insert correct month for camera-ready version
\def\year{2024} % Insert correct year for camera-ready version
\def\openreview{\url{https://openreview.net/forum?id=EUF2R6VBeU}} % Insert correct link to OpenReview for camera-ready version

\begin{document}

\maketitle

\begin{abstract}
%We adapt the ideas underlying the success of Deep Q-Learning to the continuous action domain. We present an actor-critic, model-free algorithm based on the deterministic policy gradient that can operate over continuous action spaces. Using the same learning algorithm, network architecture and hyper-parameters, our algorithm robustly solves more than 20 simulated physics tasks, including classic problems such as cartpole swing-up, dexterous manipulation, legged locomotion and car driving. Our algorithm is able to find policies whose performance is competitive with those found by a planning algorithm with full access to the dynamics of the domain and its derivatives. We further demonstrate that for many of the tasks the algorithm can learn policies end-to-end: directly from raw pixel inputs.

%We adapt the ideas underlying the success of soft actor-critic to the continuous action domain.

%Action space matters in reinforcement learning (RL). 
%We study the adaption of soft actor-critic (SAC), which is considered as a state-of-the-art reinforcement learning (RL) algorithm, from continuous action space to discrete action space. 
We study the adaptation of Soft Actor-Critic (SAC), which is considered as a state-of-the-art reinforcement learning  algorithm, from continuous action space to discrete action space. 
We revisit vanilla discrete SAC, i.e., SAC for discrete action space, and provide an in-depth understanding of its Q value underestimation and performance instability issues when applied to discrete settings. 
We thereby propose Stable Discrete SAC (SD-SAC), an algorithm that leverages entropy-penalty and double average Q-learning with Q-clip to address these issues.
%We examine the performance of our optimized discrete SAC 
Extensive experiments on typical benchmarks with discrete action space, including Atari games and a large-scale MOBA game, show the efficacy of SD-SAC. 
Our code is at: \url{https://github.com/coldsummerday/SD-SAC.git}.
%Our code is at: \url{https://github.com/revisiting-sac/Revisiting-Discrete-SAC.git}.
%in comparison to rainbow and vanilla discrete SAC.
%Soft Actor-Critic (SAC) is an off-policy actor-critic reinforcement learning algorithm, which is considered state-of-the-art in continuous action space. However, extending SAC to discrete actions still lacks an accepted practice. We revisit vanilla discrete SAC standard practice and expose it's two failure modes on Atari games. We explain why standard design choices are problematic in these cases and propose alternative choices of entropy-penalty and double average q-learning with q-clip can prevent the failure modes. We examined the performance of discrete SAC with alternative choices on both Atari games and Honor of Kings 1v1. Results demonstrate the efficacy of the proposed method in comparison to rainbow and vanilla discrete SAC.
\end{abstract}

\section{Introduction}
%Model-free deep reinforcement learning (RL) algorithms have been applied in a range of challenging domains, including games\citep{mnih2013playing,silver2016mastering}, robotic control\citep{gu2017deep,lillicrap2015continuous} and designing chip placement\citep{mirhoseini2020chip}. 
In the conventional model-free reinforcement learning (RL) paradigm, an agent can be trained by learning an approximator of action-value (Q) function \citep{mnih2015human,bellemare2017distributional}. 
The class of actor-critic algorithms \citep{mnih2016a3c,fujimoto2018td3} evaluates the policy function by approximating the value function. 
Motivated by maximum-entropy RL \citep{ziebart2008maximum,rawlik2012stochastic,abdolmaleki2018maximum}, soft actor-critic (SAC) \citep{haarnoja2018sacv1} introduces action entropy in the framework of actor-critic to balance exploitation and exploration. It has achieved remarkable performance in a range of environments with continuous action spaces \citep{haarnoja2018sacapps}, and is considered as the state-of-the-art algorithm for domains with continuous action space, e.g., Mujoco \citep{todorov2012mujoco}. 

%The success of SAC in continuous domains can be understood with the following three points: 1) With entropy regularization, the policy is trained to maximize a trade-off between expected return and entropy, a measure of randomness in the policy. 2) To avoid overestimation, SAC uses clipped double-Q-learning and target policy smoothing tricks to stabilize the q value for optimization. And the soft-q-value is distilled to the policy distribution through KL divergence. 3) The balance between expected return and entropy is regulated by temperature $\alpha$. SAC automatically adjusts the temperature $\alpha$ by a priori target entropy using the Lagrangian function.

%\textcolor{red}{We failed to motivate why we need a discrete sac. Or, why do we want to extend it to discrete action space? We have to say something here.} 

However, while SAC solves problems with continuous action space, it cannot be directly applied to discrete domains since it relies on the reparameterization of Gaussian policies to sample actions, in which the action in discrete domains is categorical. Soft-DQN \citep{VieillardPG20Munchausen} provides a simple way to discretize SAC by adopting the maximum-entropy RL to DQN \citep{mnih2013playing}. However, Soft-DQN utilizes only a Q-value parametrization to bypass the policy parameterization.
%instead of using policy parameters and minimizing the KL-divergence between the policy distribution and Q-value.} 
Another discretization of the continuous action output and Q value in vanilla SAC is suggested by previous work \citep{christodoulou2019sacd} to adapt SAC to discrete domains, resulting in discrete SAC (DSAC).  
However, it is counter-intuitive that the empirical experiments in subsequent efforts \citep{xu2021targetannealing} indicate that discrete SAC performs poorly in discrete domains, e.g., Atari games. %, which serve as a typical benchmark for RL research with discrete action space.  
We believe that the idea of maximum entropy RL applies to both discrete and continuous domains. 
However, extending the maximum-entropy-based SAC algorithm to discrete domains still lacks a commonly accepted practice in the community. 
Therefore, in this paper, similar to the motivation of DDPG (deep deterministic policy gradient)  \citep{lillicrap2015continuous}, which adapts DQN (deep Q networks) \citep{mnih2013playing} from discrete action space to continuous action space, we aim to optimize SAC algorithm for discrete domains.

% \textcolor{blue}{
% In many tasks, like DOTA et al, discrete action space can potentially achieve better exploration to bypass suboptimal policies\citep{tang2020discretizing}. However, SAC cannot be straight-forwardly applied to discrete domains since it relies on the reparameterization of gaussian policies to sample action, in which the action in discrete domains is categorical. Extending SAC with maximum entropy reinforcement learning to discrete actions to accommodate more domains still lacks an accepted practice. Simple discretization of the continuous action output and q-value in a vanilla SAC is an obvious strategy suggested by \citep{christodoulou2019sacd} to adapt SAC to discrete domains named discrete SAC. It is counter-intuitive that previous work\citep{xu2021targetannealing} shows that discrete SAC does not perform well in discrete domains.
% }

% However, extending SAC with maximum entropy reinforcement learning to discrete actions to accommodate more domains still lacks an accepted practice. Simple discretization of the continuous action output and q-value in a vanilla SAC is an obvious strategy suggested by \citep{christodoulou2019sacd} to adapt SAC to discrete domains named discrete SAC, but this approach has not been as successful as SAC in discrete action spaces\citep{xu2021targetannealing} \textcolor{red}{ what does this sentence mean?  discrete -> continuous?}.

Previous studies \citep{xu2021targetannealing,wang2020metasac} have analyzed the reasons for the performance disparity of SAC between continuous and discrete domains. Reviewing from the perspective of automating entropy adjustment, an unreasonable setting of target-entropy for temperature $\alpha$ may break the SAC value–entropy trade-off \citep{wang2020metasac,xu2021targetannealing}. 
Furthermore, the function approximation errors of Q-value are known to lead to estimation bias and hurt performance in actor-critic methods~\citep{fujimoto2018td3}.
To avoid overestimation bias, both discrete SAC and continuous SAC resort to clipped double Q-learning \citep{fujimoto2018td3} for actor-critic algorithms. 
On the contrary, using the lower bound approximation to the critic can lead to underestimation bias, which makes the policy fall into pessimistic underexplored, as pointed by \citep{ciosek2019oac,pan2020softmax}, mainly when the reward is sparse. 
However, existing works only focus on continuous domains \citep{ciosek2019oac,pan2020softmax}, while SAC for discrete cases remains less explored. 

In addition to the abovementioned issues, we conjecture that discrete SAC fails also due to the absence of policy update constraints. 
Intuitively, the unstable training causes a shift in the Q function distribution and policy entropy, which generates a rapidly changing target for the critic network due to the soft Q-learning objective. 
Meanwhile, the critic network in SAC needs time to adapt to the oscillating target process, exacerbating policy instability.
%\textcolor{red}{This may lead to potential issues for the trust region policy updates objective in discrete SAC. - wrong sentence}. 
%Therefore, we suggest an extra entropy-penalty on the policy optimization objective to constrain policy update, which ensures the stability of the discrete sac training process. 
%On the other hand, in terms of the Q-value estimation, previous works \citep{fujimoto2018td3,duan2021distributionalsac} have focused on overestimation bias on Q-function-based methods in continuous domains, while an overestimation bias phenomenon in discrete SAC can also exist, as SAC can enter a pessimistic exploration when rewards become sparse. 
%Therefore, we propose double average Q-learning and Q-approximation-clip to confine the Q value within a reasonable range.

%To verify the aforementioned conjecture, 
To address the above challenges, we first design test cases to replicate the failure modes of vanilla discrete SAC, exposing its inherent weaknesses regarding training instability and Q-value underestimation.   % does not achieve the desired convergence in discrete action spaces.  
Then, accordingly, we propose Stable Discrete SAC (SD-SAC) 
to stabilize the training. We develop an entropy penalty on the policy optimization objective to constrain policy updates. We also develop double average Q-learning with Q-clip to confine the Q value within a reasonable range. 
We use Atari games (the default testbed for the RL algorithm for discrete action space) to verify the effectiveness of our optimizations. 
%Given the recent concerns in reproducibility \citep{henderson2018matters}, we run our experiments across multiple seeds with fair evaluation metrics. 
%Not only did we verify our contributions on Atari games, 
We also deploy our method to the Honor of Kings 1v1 game, a large-scale MOBA game used extensively in recent RL advances \citep{ye2020supervised,ye2020mastering,ye2020towards,wei2022hokenv}, to demonstrate the scale-up capacity of our Stable Discrete SAC. 

To sum up, our contributions are: %the key contributions of this paper are featured as follows:
\begin{itemize}
    \item We pinpoint two failure modes of discrete SAC, regarding training instability and underestimated Q values. We find that the underlying causes are the environment's deceptive rewards and SAC's double Q learning respectively.

    %\item  Policy instability with discrete sac on training process due to lacking constraints for trust region policy update. We propose entropy-penalty to improve the unstable performance of discrete sac.
    \item To alleviate the training instability issue, we propose entropy-penalty to constrain the policy update in discrete SAC.  

    %\item We pay attention to the underestimation bias of discrete sac and analyze the negative impact on pessimistic underexplore caused by underestimation bias. To Address this issue, double average q-learning with q-approximation-clip is suggested to estimate state-action value. 
    \item To deal with the underestimation bias of Q value in discrete SAC, we propose double average Q-learning with Q-clip to estimate the state-action value.
 \end{itemize}
 With the above contributions, we have obtained the Stable Discrete SAC (SD-SAC) algorithm. Extensive experiments on Atari games and a large-scale MOBA game show SD-SAC's superiority compared to baselines, with a 68\% improvement of normalized scores in Atari and around 100\% ELO increase in the Honor of Kings 1v1 game environment.

%The remaining contents of this paper are organized as follows. Related works and SAC background are overviewed in Section \ref{section_related_works} and Section \ref{section_preliminaries}. Analysis of failure modes of vanilla discrete SAC is presented in Section \ref{section_failure_modes}. We present improvements for SAC failure modes and experimental results in Section \ref{section_improvement}. Section \ref{section_experiments} demonstrates experimental results and a discussion of these results associated with discussions The paper ends in Section \ref{section_discussion} with conclusion and future work propositions.

%We aim to answer: Can maximum entropy RL be naturally applied to discrete action space?  If not, what are the barriers that affects the performance of discrete SAC? Consequently, can we optimize SAC?

\section{Related Work} \label{section_related_works}
%In this section, 
% We review recent efforts on algorithmic improvements to soft actor-critic. %The most relevant efforts are on Q estimation and performance stability, summarized below. 

%\textcolor{red}{this review is all of a mess.... }

%Therefore, we summarize the representative works of this two lines below. 

% \subsection{Adaption of Action Space}
\textbf{Adaptation of Action Space}. 
The most relevant works to this paper are vanilla discrete SAC \citep{christodoulou2019sacd}, TES-SAC \citep{xu2021targetannealing} and Soft-DQN \citep{VieillardPG20Munchausen}. Discrete SAC replaces the Gaussian policy with a categorical one and discretizes the Q-value output to adapt SAC from continuous to discrete action space. However, as we will point out, a direct discretization of SAC will have specific failure modes with poor performance. 
%TES-SAC point out that it is counter-intuitively that SAC does not work well for discrete action space. They propose a new scheduling method for the target entropy parameters in discrete SAC. However, they contend that the failure modes of discrete SAC are due to unreasonable target-entropy parameters. In contrast, we point out that poor adaption of discrete SAC results from policy instability and underestimated Q value. 
TES-SAC proposes a new scheduling method for the target entropy parameters in discrete SAC. 
% Soft-DQN has discretized SAC by adopting the maximum-entropy RL to DQN. Unlike SAC, which uses both policy and Q value parameters simultaneously, Soft-DQN utilizes only a Q value parametrization and directly applies a softmax operation to the Q-values to take action.
Soft-DQN has discretized SAC by adopting the maximum-entropy RL to DQN, utilizing only a Q value parametrization and directly applies a softmax operation to the Q-values to take action.
%\subsection{Estimating Q-value of SAC} 

% \subsection{Q Estimation}
\textbf{Q Estimation}.
Previous works \citep{fujimoto2018td3,ciosek2019oac,pan2020softmax,duan2021distributionalsac} have already expressed concerns about the estimation bias of Q value for SAC. SD3 \citep{pan2020softmax} proposes to reduce the Kurtosis distribution of Q approximately by using the softmax operator on the original Q value output to reduce the overestimation bias. OAC \citep{ciosek2019oac} constrains the Q value approximation objective by calculating the upper and lower boundaries of two Q-networks. Distributional SAC \citep{duan2021distributionalsac} replaces the Q learning target with the expected reward sum obtained from the current state to the end of the episode and uses a multi-frame estimates target to reduce overestimation. 
% These methods unavoidably increase the complexity cost while obtaining an accurate Q-value overestimation with continuous action spaces. 
% However, little research is on discrete settings. By comparison, we propose an approximation method by replacing double average Q outputs as the learning target and clipping the current Q value with the target network. We prevent both overestimation and underestimation with little extra computational cost.
Maxmin Q-learning \citep{LanPFW20maxmin} controls estimation bias by minimizing the complete ensemble in the target. MME \citep{HanS21MME} extends max-min operation to the entropy framework to adapt to SAC. REM \citep{AgarwalS020REM} ensembles Q-value estimations with a random convex combination to enhance generalization in the offline setting. REDQ \citep{chen2021REDQ} reduces the estimation bias by minimizing a random subset of Q-functions. AEQ \citep{GongLYZL23AEQ} adjusts the estimation bias by using the mean of Q-functions minus their standard deviation. However, little research is on discrete settings. Our approach focuses on reducing the underestimation bias for the double Q-estimators to enhance exploration.

%\subsection{Performance Stability of SAC}
% \subsection{Performance Stability}
\textbf{Performance Stability}.
Flow-SAC \citep{ward2019improving} applies a technique called normalizing flows policy on continuous SAC
%. With normalizing flows policy, modeling expressively is higher than with Gaussian policy, 
leading to the finer transformation that improves training stability when exploring complex states. However, applying normalizing flows to discrete domains will cause a degeneracy problem \citep{horvat2021denoising}, making it difficult to transfer to discrete actions. SAC-AWMP \citep{hou2020off} improves the stability of the final policy by using a weighted mixture to combine multiple policies. 
Based on this method, the cost of network parameters and inference speed is significantly increased. ISAC \citep{banerjee2022improved} increases SAC stability by mixing prioritized and on-policy samples, enabling the actor to repeat learns states with drastic changes. Repeatedly learning priority samples, however, runs the risk of settling into a local optimum. By comparison, our method improves policy stability in case of drastic state changes with an entropy constraint.

\section{Preliminaries} \label{section_preliminaries}
% This section briefly overviews the symbol definitions of SAC for discrete action space. 

%\subsection{Notation}
%Reinforcement learning (RL) aims to learn optimal policies for an agent acting in an environment with a scalar reward signal. Formally, a Markov decision process (MDP) defined by $(S,A,r,p)$, where $S$ represents the state space, $A$ is the action space, $p(\hat{s}|s,a)$ is the probability of a state transition from current state $s\in S$.  The agent chooses the action $a\in A$ to translate to the next state $\hat{s} \in S$ and receives signal reward $r(s,a)$. A parameterized policy $\pi(a|s)$ can be used to generate action given a state. The task of reinforcement learning is to optimize a policy, and then use the optimized policy $\pi^{*}$ interact with the environment to maximize expected return $R =\sum_{t=0}^{T} \underset{\substack{s_{t} \sim p \\ a_{t} \sim \pi}}{\mathbb{E}}[r(s_{t}, a_{t})]$

%\subsection{Discrete Soft Actor-Critic}
This section briefly overviews the symbol definitions of SAC for discrete action space. Following the maximum entropy framework, SAC adds an entropy term $\mathcal{H}(\pi(\cdot \mid s))$ as regularization to the policy gradient objective:

\begin{equation}
\pi^{*}=\underset{\pi}{\operatorname{argmax}}\sum_{t=0}^{T} \left[\gamma^t \underset{\substack{s_{t} \sim p \\ a_{t} \sim \pi}}{\mathbb{E}}[r(s_{t}, a_{t})+\alpha \mathcal{H}(\pi(\cdot \mid s_t))]\right],
\end{equation}
\begin{equation}
\begin{aligned}
\mathcal{H}(\pi(\cdot \mid s)) &=-\sum_a \pi(a \mid s) \log \pi(a \mid s) =\underset{a \sim \pi(\cdot \mid s)}{\mathbb{E}}[-\log \pi(a \mid s)]
\end{aligned}
\end{equation}
where $\pi$ is a policy, $\pi^{*}$ is the optimal policy, and $\alpha$ is the temperature parameter that determines the relative importance of the entropy term versus the reward $r$, thus controls the stochasticity of the optimal policy.
% \begin{equation}
% \begin{aligned}
% \mathbb{H}(\pi(\cdot \mid s)) &=-\int_{a \in \mathcal{A}} \pi(a \mid s) \log \pi(a \mid s) \mathrm{d} a \\
% &=\underset{a \sim \pi(\cdot \mid s)}{\mathbb{E}}[-\log \pi(a \mid s)].
% \end{aligned}
% \end{equation}

%To maximize the objective, SAC iterates between: 1) updating the soft Q-function $Q(s,a)$ to match a soft bellman backup target; 2) minimizing the KL divergence between policy $\pi$ and the soft Q-function.

\textbf{Soft Bellman Backup}
The soft Q-function, parametrized by $\theta$, is updated by reducing the soft Bellman error as described in the next subsection:
\begin{equation}
J_{Q}(\theta)=\frac{1}{2}\left(r(s_{t},a_{t})+\gamma V(s_{t+1})-Q_{\theta}(s_{t}, a_{t})\right)^{2},
\end{equation}
where $V(s_{t})$ defines the soft state value function, which represents the expected reward estimate that policy obtains from the current state to the end of the trajectory.
\begin{equation}
V(s_{t})=\mathbb{E}_{a_{t} \sim \pi}[Q_{\theta}(s_{t}, a_{t})-\alpha \log (\pi(a_{t} \mid s_{t}))].
\end{equation}
Soft actor-critic minimizes soft Q-function with final soft Bellman error:
% \begin{equation}
% \resizebox{0.9\hsize}{!}{$
% J_{Q}(\theta)  =\mathbb{E}_{(s_{t}, a_{t}) \sim D}[\frac{1}{2}(Q_{\theta}(s_{t}, a_{t})-(r(s_{t}, a_{t}) + \gamma \mathbb{E}_{s_{t+1}} \sim p(s_{t}, a_{t})[V(s_{t+1})]))^{2}]
% $}
% \label{equation_q}
% \end{equation}
% \begin{equation}
% \resizebox{0.9\hsize}{!}{$
% J_{Q}(\theta)  =\mathbb{E}_{(s_{t}, a_{t}) \sim D}[\frac{1}{2}(Q_{\theta}(s_{t}, a_{t})-(r(s_{t}, a_{t}) + \gamma \mathbb{E}_{s_{t+1} \sim p(\cdot \mid s_{t}, a_{t})}[V(s_{t+1})]))^{2}]
% $}
% \label{equation_q}
% \end{equation}
\begin{equation}
    \begin{aligned}
    J_{Q}(\theta)  =\mathbb{E}_{(s_{t}, a_{t}) \sim D}[\frac{1}{2}(Q_{\theta}(s_{t}, a_{t})-
    (r(s_{t}, a_{t}) + \gamma \mathbb{E}_{s_{t+1} \sim p(\cdot \mid s_{t}, a_{t})}[V(s_{t+1})]))^{2}],
    \end{aligned}
    \label{equation_q}
\end{equation}
where $D$ is a replay buffer replenished by rollouts of the policy $\pi$ interacting with the environment. In the implementation, SAC \citep{haarnoja2018sacv1} uses the minimum of two delayed-update target-critic network outputs as the soft bellman learning objective to reduce overestimation. The formula is expressed as
% \begin{equation}
% \resizebox{0.9\hsize}{!}{$
%  V(s_{t+1}) =  \min_{i=1,2}\mathbb{E}_{a_{t} \sim \pi}[Q_{\theta_{i}^{\prime}}(s_{t+1}, a_{t+1})-\alpha \log (\pi(a_{t+1} \mid s_{t+1}))]
%  $},
% \end{equation}
\begin{equation}
\begin{aligned}
 V(s_{t+1})=\min_{i=1,2}\mathbb{E}_{a_{t} \sim \pi}[Q_{\theta_{i}^{\prime}}(s_{t+1}, a_{t+1})-\alpha \log (\pi(a_{t+1} \mid s_{t+1}))],
\end{aligned}
\end{equation}
where $Q_{\theta_{i}^{\prime}}$ represents  $i$-th target-critic network.

\textbf{Policy Update Iteration}
The policy, parameterized by $\phi$, distills the softmax policy induced by the soft Q-function. The discrete SAC policy directly maximizes the probability of discrete actions, in contrast to the continuous SAC policy, which optimizes the two parameters of the Gaussian distribution. Then, the discrete SAC policy is updated by minimizing KL divergence between the policy distribution and the soft Q-function.
\begin{equation}
\pi_{\phi_{new}}=\underset{\pi_{\phi_{old}} \in \Pi}{\operatorname{argmin}} D_{\mathrm{KL}}\left(\pi_{\phi_{old}}(. \mid s_{t}) \| \frac{\exp (\frac{1}{\alpha} Q^{\pi_{\phi_{old}}}(s_{t}, .))}{Z^{\pi_{\phi_{old}}}(s_{t})}\right).
\end{equation}
Note that the partition function ${Z^{\pi_{\phi_{old}}}(s_{t})}$ is a normalization term that can be ignored since it does not affect the gradient concerning the new policy. The resulting optimization objective of the policy is as follows:
\begin{equation}
J_{\pi}(\phi)=\mathbb{E}_{s_{t} \sim D}[\mathbb{E}_{a_{t} \sim \pi_{\phi}}[\alpha \log (\pi_{\phi}(a_{t} \mid s_{t}))-Q_{\theta}(s_{t}, a_{t})]].
\label{equation_policy_update}
\end{equation}

\textbf{Automating Entropy Adjustment}
The entropy parameter temperature $\alpha$ regulates the value-entropy balance in soft Q learning. The SAC paper proposes using the temperature Lagrange term to tune the temperature $\alpha$ automatically. The following equation can be regarded as the optimization objective satisfying an entropy constraint.
\begin{equation}
    \begin{aligned}
    &\max _{\pi_{0: T}}\mathbb{{E}_{\rho_{\pi}}}\left[\sum_{t=0}^{T}r(s_{t}, a_{t})\right]    \quad \text {s.t. } \mathbb{{E}}_{({s}_{t}, {a}_{t}) \sim \rho_{\pi}}[-\log (\pi_{t}(a_{t} \mid s_{t}))] \geq \mathcal{H} , \forall t,
    \end{aligned}
\end{equation}
% \begin{equation}
%     \resizebox{0.8\hsize}{!}{$
%     \max _{\pi_{0: T}}\mathbb{{E}_{\rho_{\pi}}}[\sum_{t=0}^{T} r(\mathbf{s}_{t}, \mathbf{a}_{t})] \text { s.t. } \mathbb{{E}_{(\mathbf{s}_{t}, \mathbf{a}_{t}) \sim \rho_{\pi}}}[-\log (\pi_{t}(\mathbf{a}_{t} \mid \mathbf{s}_{t}))] \geq \mathcal{H} , \forall t
%     $}
% \end{equation}
where $\mathcal{H}$ is the desired minimum expected entropy. Optimizing the Lagrangian term $\alpha$ involves minimizing:
\begin{equation}
J(\alpha)=\mathbb{E}_{(a \mid s) \sim \pi_{t}}[- \alpha \log \pi_{t}(a_{t} \mid s_{t})- \alpha{\mathcal{H}}].
\end{equation}

By setting a loose upper limit on the target entropy $\mathcal{H}$, SAC achieves automatic adjustment of temperature $\alpha$. Typically, the target entropy is set to $0.98 * -log(\frac{1}{dim(Actions)})$ for discrete\citep{christodoulou2019sacd} and $-dim(Actions)$ for continuous actions\citep{haarnoja2018sacapps}.

\section{Failure Modes of Vanilla Discrete SAC}  \label{section_failure_modes}
We start by outlining the failure modes of the vanilla discrete SAC and then analyze under what circumstances the standard choices of vanilla discrete SAC perform poorly.

\subsection{Unstable Coupling Training} \label{subsection_failure_mode_policy}
The first failure mode comes from the instability caused by fluctuations in Q function distribution and policy entropy during training.
The maximum entropy mechanism in SAC effectively balances exploration and exploitation. However, due to the existence of entropy term in the soft Bellman error, and the mechanism in discrete SAC that aligns the policy with the Q function, the policy update iteration (Eq.~\ref{equation_policy_update}) is strongly coupled with Q-learning iteration (Eq.~\ref{equation_q}).
%it also makes policy update iteration coupled with Q-learning iteration while implicitly adding the policy term (entropy) to soft Bellman error \zhihui{``while'' is not clear, do you mean like ``due to''?}. 

In environments with deceptive rewards \citep{hong2018diversity}, the agent can gain substantial returns in the early stages of training through short-term rewards, causing the Q value of specific actions to rise rapidly and the Q function distribution to become sharper. The coupling learning paradigm of discrete SAC leads to a sharper policy distribution and, thus, a decline in entropy. Consequently, the Q learning target becomes unstable, which can, in turn, deteriorate the policy learning. As a result, the agent falls into local optima and struggles to discover alternative strategies with larger long-term payoffs.
%if abrupt state changes result in policy instability and Q learning target variations as the result of entropy chattering. 
% state distribution changes -> policy changes -> entropy rises -> Q target change
To illustrate this issue more concretely, we take the training process of discrete SAC in the Atari game Asterix as an example.
%consider the following Atari example:

% \begin{figure}[!t]

% \includegraphics[width=0.2\textwidth]{images/analysis/fail_mode_1/Asterix_simi.jpeg}
% \includegraphics[width=0.2\textwidth]{images/analysis/fail_mode_1/Asterix_entropy.jpeg}

% \includegraphics[width=0.2\textwidth]{images/analysis/fail_mode_1/Asterix_q1.jpeg}
% \includegraphics[width=0.2\textwidth]{images/analysis/fail_mode_1/Asterix_reward.jpeg}

% \caption{Measuring cosine similarity of states, policy action entropy, estimation of q-value and final evaluation reward on Atari Game  Asterix environment with baseline-SAC-discrete over 10 million time steps}
% \label{fig_cos_simi_entropy_analysis}
% \end{figure}

\begin{figure} [!t]
    \centering
    \subfigure[Gameplay screenshot of the Atari Game Asterix]{
		\includegraphics[width=0.75\textwidth]{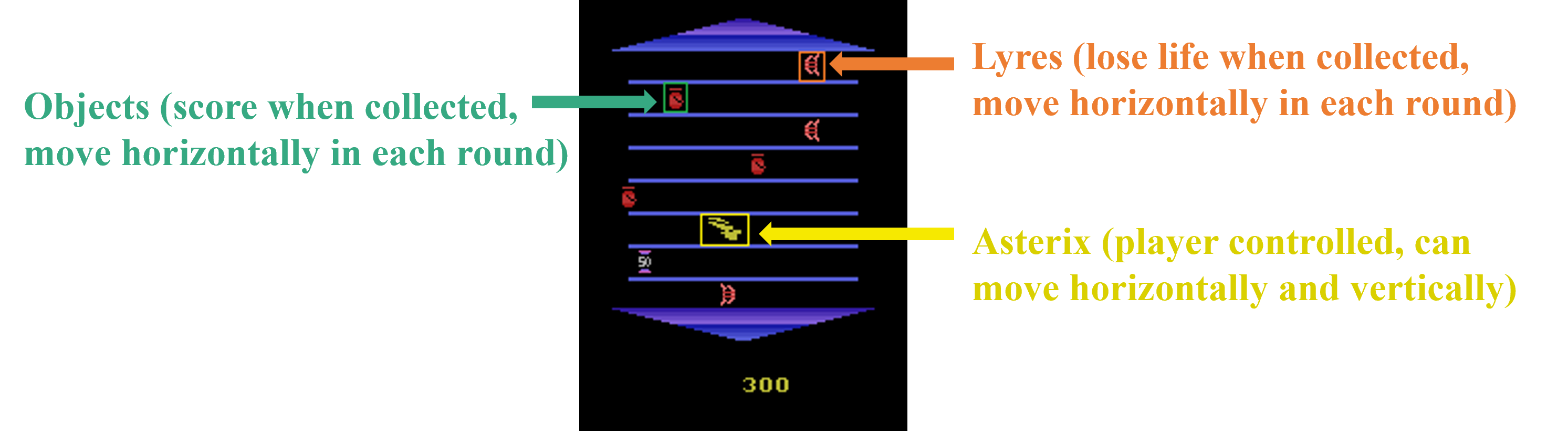}
		\label{subfig_asterix_example}
		}
	\subfigure[Deceptive rewards in Asterix]{
		\includegraphics[width=0.16\textwidth]{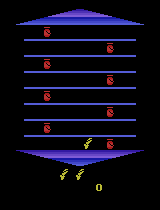}
		\label{subfig_asterix_init}
		}
    \caption{Gameplay screenshot of the Atari Game Asterix, including the player-controlled Asterix (yellow box), scoring objects (green box) and life-losing lyres (orange box) that appear in rounds. Deceptive rewards appear in the early stage of game when there are only scoring objects.}
    \label{fig_asterix_example} 
\end{figure}

\begin{figure} [!t]
    \centering
	\subfigure[Q Function Variance]{
		\includegraphics[width=0.31\textwidth]{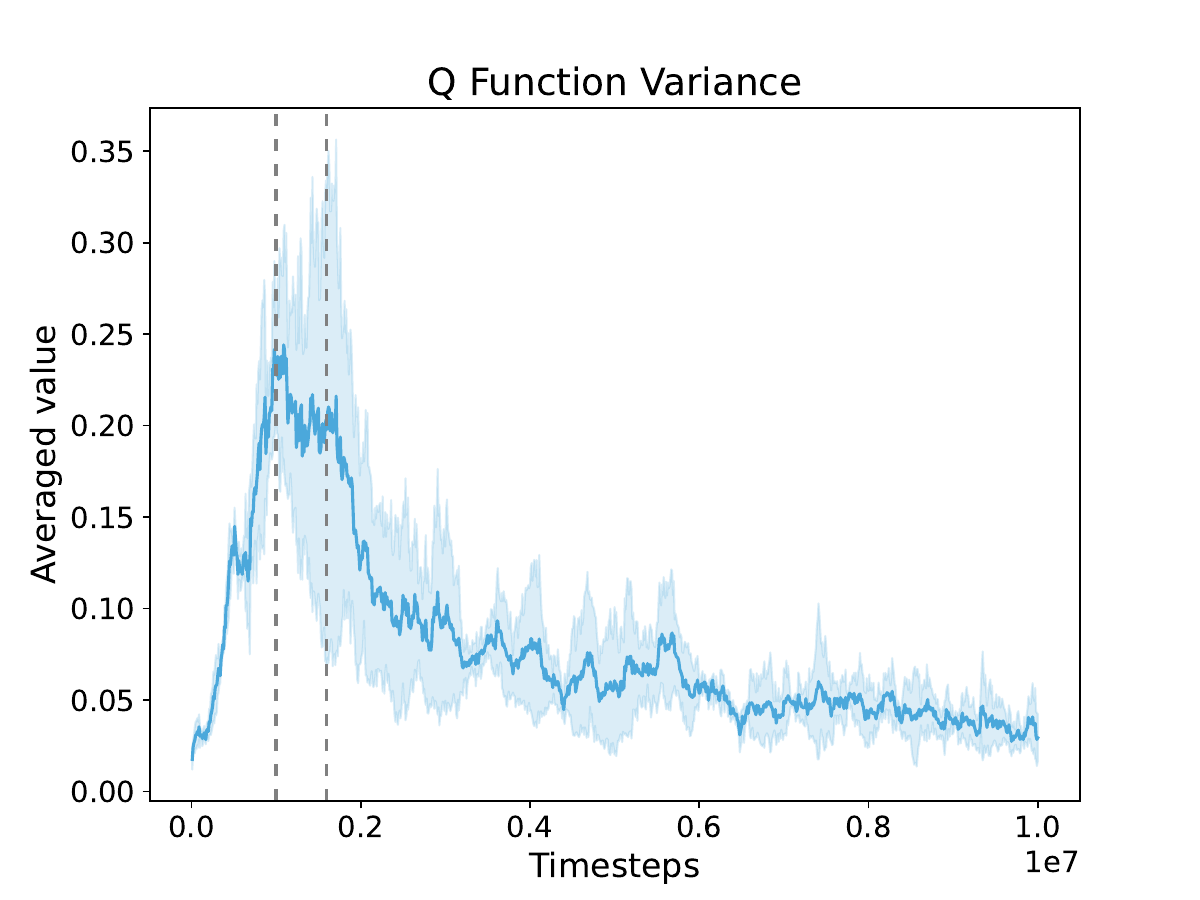}
		\label{subfig_cos_simi_entropy_analysis_qvar}
		}
	\subfigure[Q-value]{
		\includegraphics[width=0.31\textwidth]{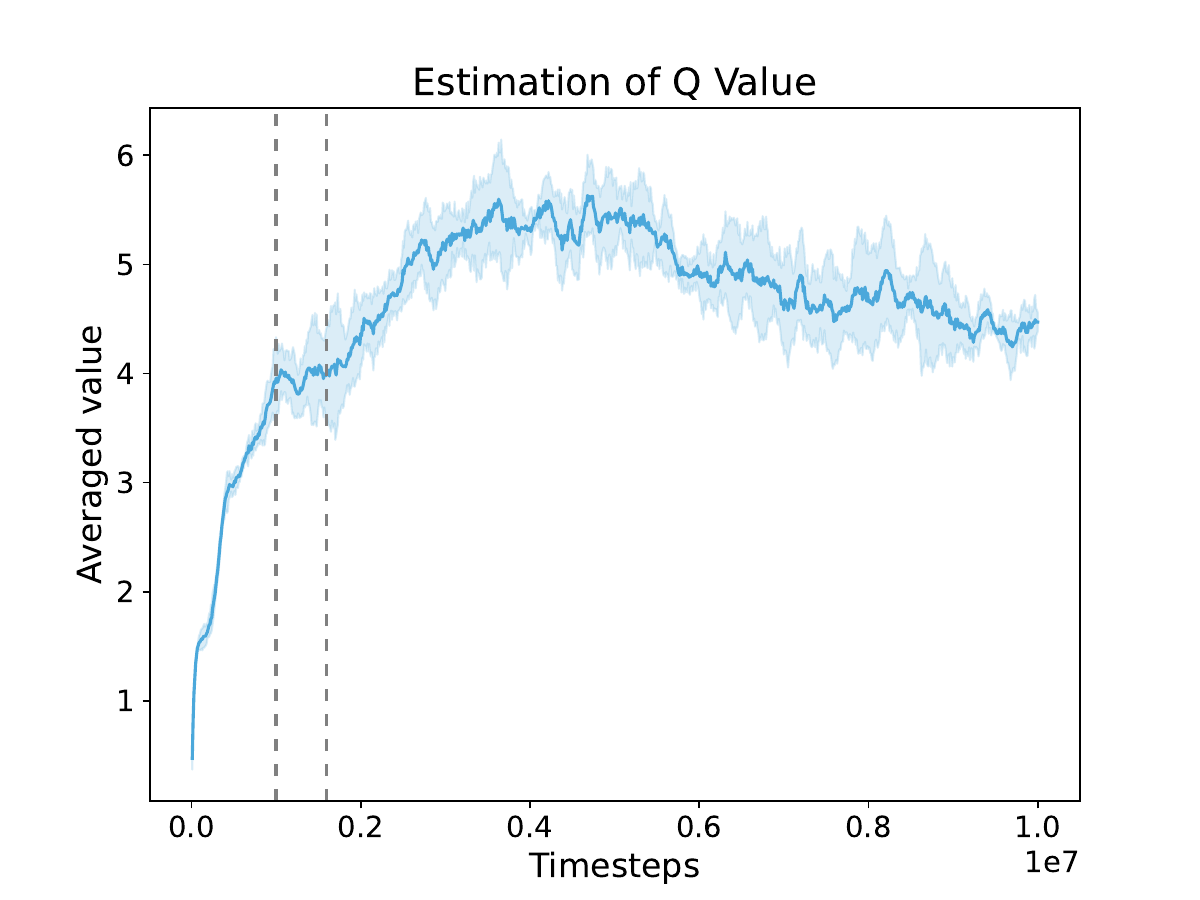}
		\label{subfig_cos_simi_entropy_analysis_q}
		}
	\subfigure[Entropy]{
		\includegraphics[width=0.31\textwidth]{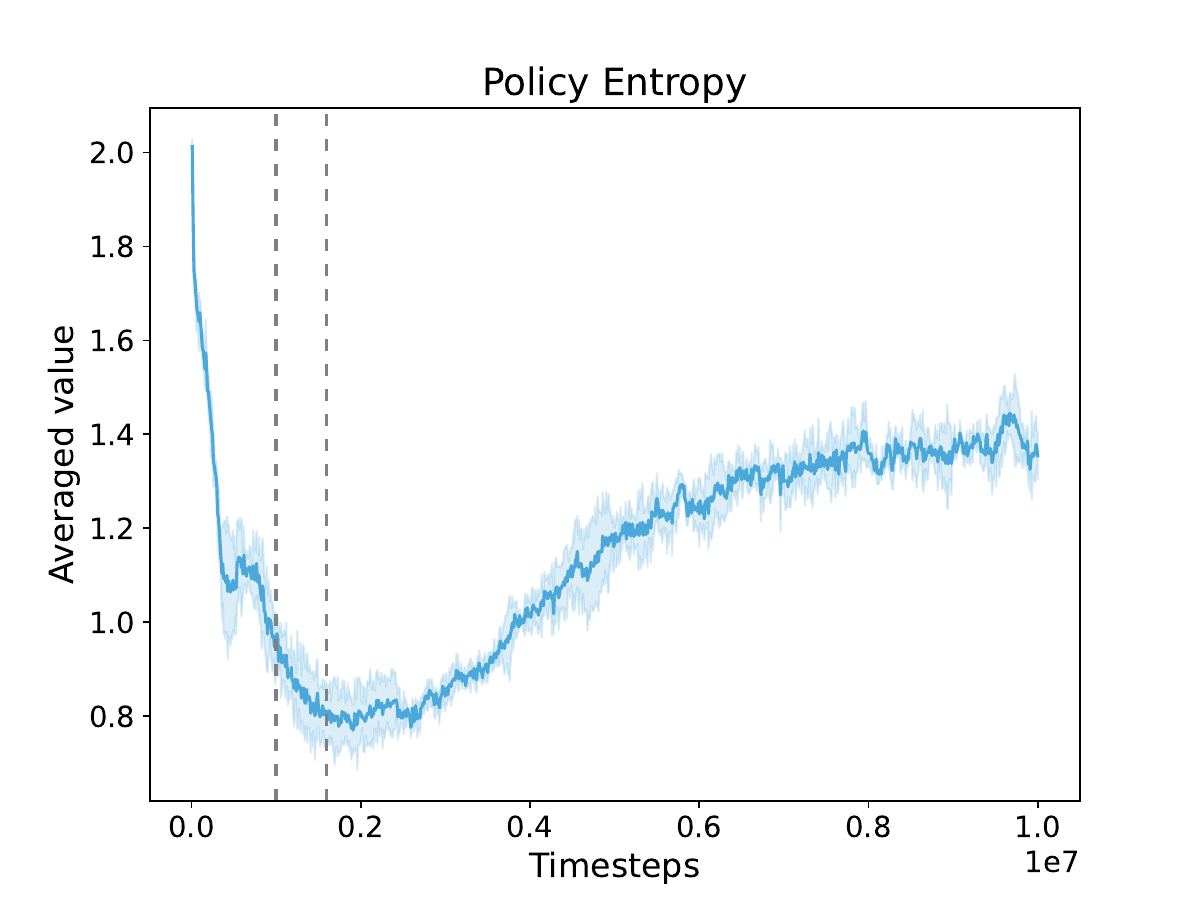}
		\label{subfig_cos_simi_entropy_analysis_entropy}
		}
        \\
  	\subfigure[Episode Length]{
		\includegraphics[width=0.31\textwidth]{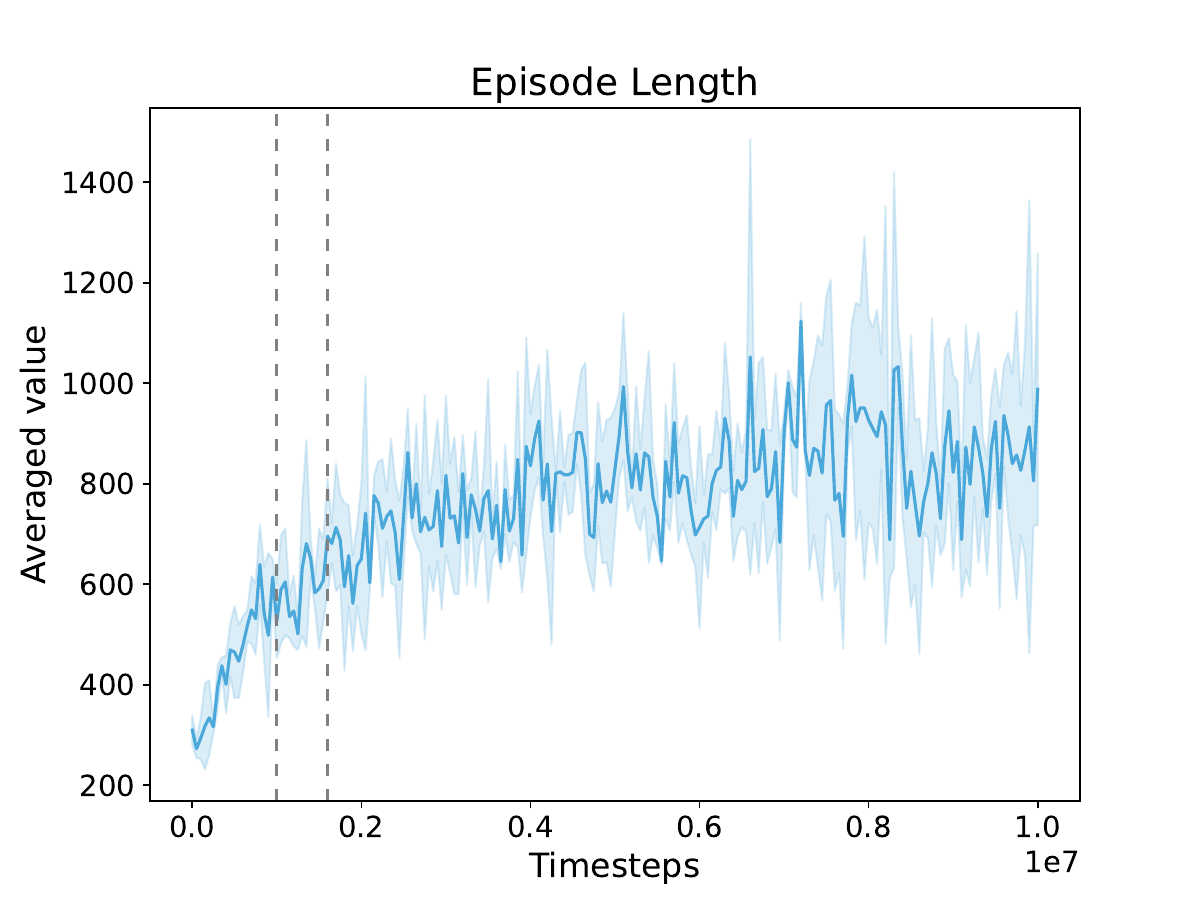}
		\label{subfig_cos_simi_entropy_analysis_length}
		}
        \subfigure[Steps with Rewards]{
		\includegraphics[width=0.31\textwidth]{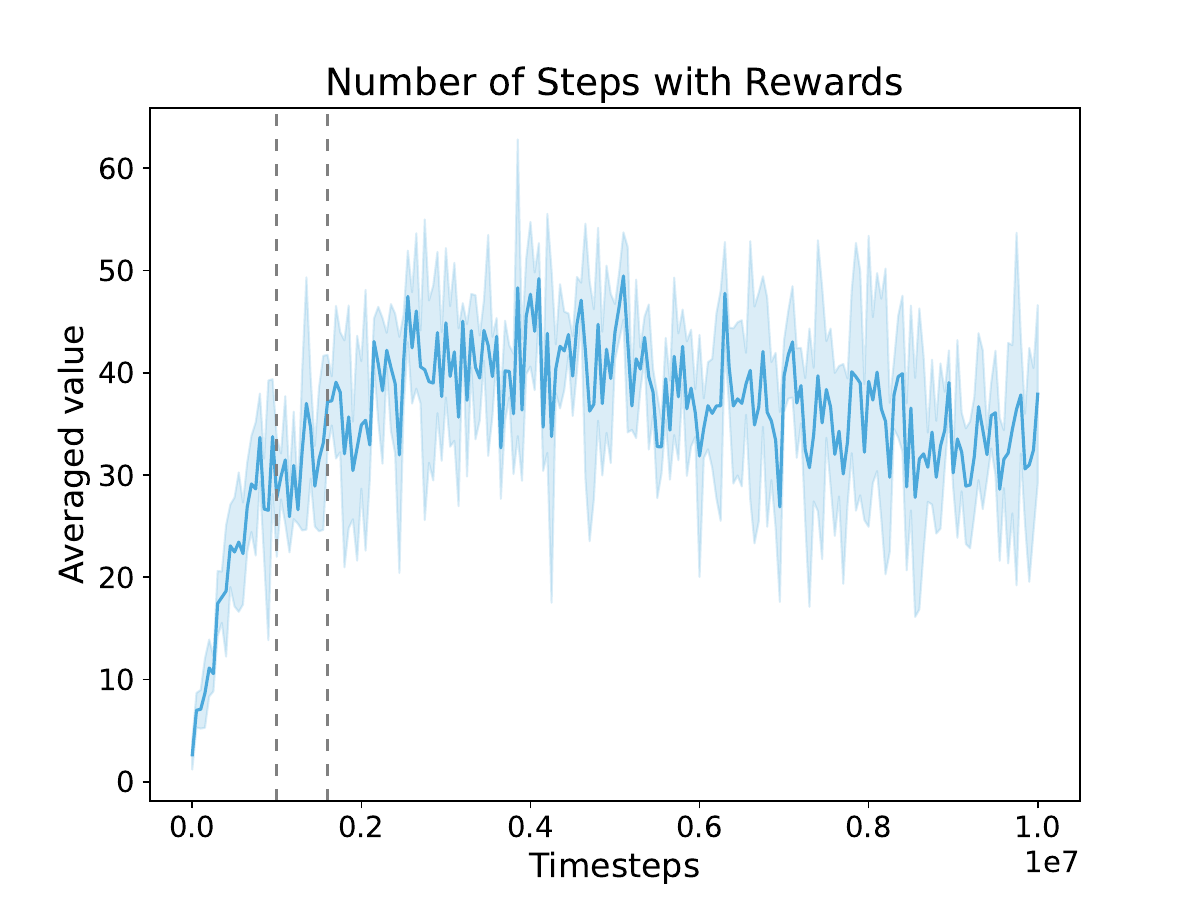}
		\label{subfig_cos_simi_entropy_analysis_rwstp}
		}
        \subfigure[Score]{
		\includegraphics[width=0.31\textwidth]{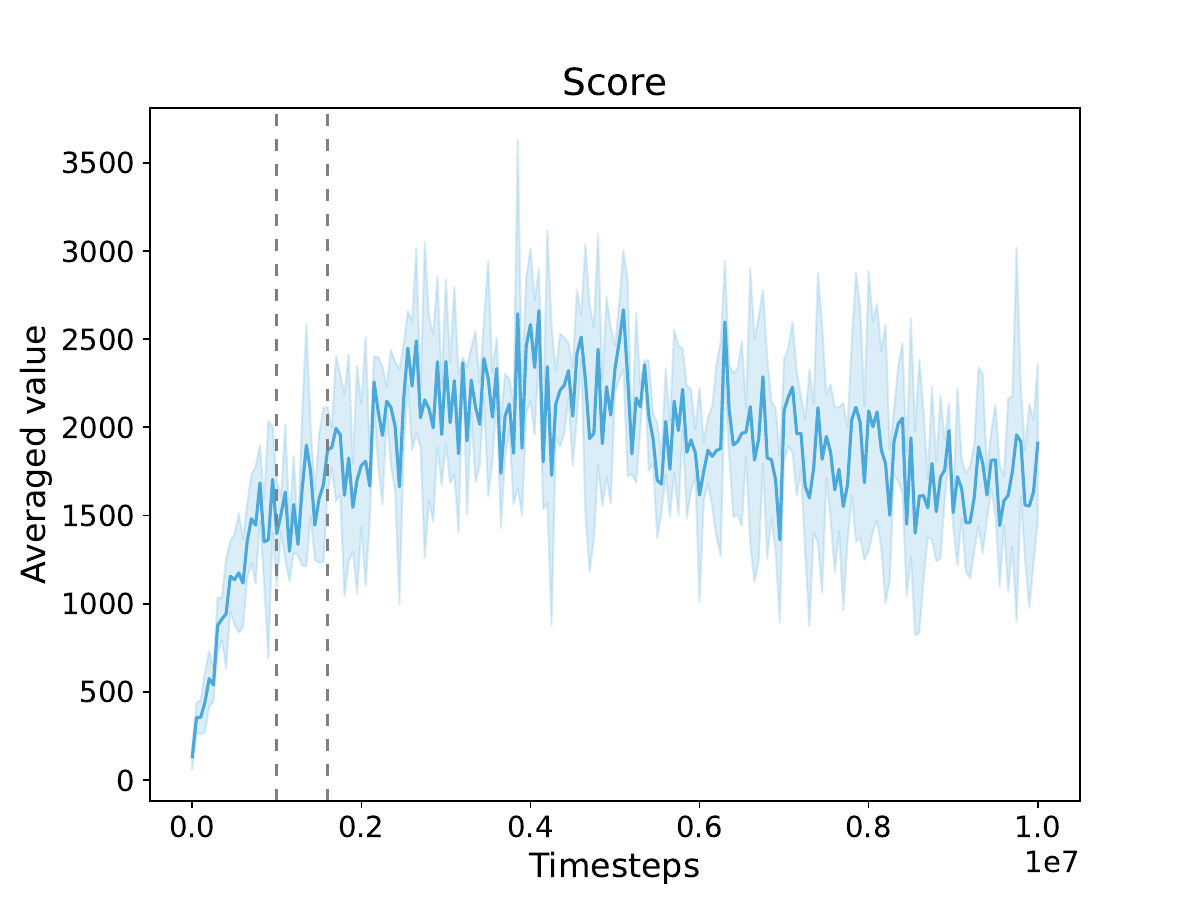}
		\label{subfig_cos_simi_entropy_analysis_return}
		}
	\caption{Measuring Q variance, estimation of Q-value, policy entropy, episode length, steps with rewards, and score on Atari Game Asterix with discrete SAC over 10 million timesteps.}
	\label{fig_cos_simi_entropy_analysis} 
\end{figure}

% The procedure can be described as following: 
%\zhihui{use ``;'' after each step; better to reference back to each subfigure} 
As shown in Fig.~\ref{subfig_asterix_example}, the player controls Asterix, which can move horizontally and vertically. In each round, horizontally moving objects appear. Asterix scores points by collecting objects and loses a life when collecting a lyre. In the early stage of the game, rounds often appear where there are only scoring objects and no life-losing lyres (Fig.~\ref{subfig_asterix_init}), allowing the agent to score quickly by collecting objects, resulting in deceptive rewards. These rewards make the Q function sharper, thereby reducing the entropy of the policy. In Fig.~\ref{subfig_cos_simi_entropy_analysis_qvar}, we sample a fixed set of states, and measure the variance of Q function across different actions for these states. We find that the Q function variance increases rapidly, indicating that the Q function becomes sharp quickly. Policy entropy also decreased during this period.
%the changed state manifested as an increase in the state cosine similarity \zhihui{what is state cosine similarity and how to compute it? probably use ``cosine similarity between xxx and xxx''}. 

As the learning process continues, the policy entropy drops rapidly, and the action probabilities become deterministic quickly (Fig.~\ref{subfig_cos_simi_entropy_analysis_entropy}). The agent can collect objects but struggles to avoid obstacles effectively. After the policy entropy reaches its lowest point at round 2 million steps, neither the episode length (Fig.~\ref{subfig_cos_simi_entropy_analysis_length}) nor the number of steps with rewards (Fig.~\ref{subfig_cos_simi_entropy_analysis_rwstp}) increases significantly.
%due to the significant reward brought by state changes, and action probabilities become deterministic. 
At the same time, the drastic change of policy entropy misleads the learning process, and thus, both Q-value and policy fall into local optimum (Fig.~\ref{subfig_cos_simi_entropy_analysis_entropy} and Fig.~\ref{subfig_cos_simi_entropy_analysis_q}).
%converge due to the determinism of policy (Fig.~\ref{fig_cos_simi_entropy_analysis}(c)).
%the combinatory impact of state distribution changes and entropy decrease. 
Since both policy and Q-value converge to the local optimum, it becomes hard for the policy to explore efficiently in the later training stage. Even the policy entropy re-rises in the later stage (Fig.~\ref{subfig_cos_simi_entropy_analysis_entropy})), the performance of policy does not improve anymore (Fig.~\ref{subfig_cos_simi_entropy_analysis_return}). Similar situations also occur in other Atari environments, and we provide more examples in Appendix \ref{appendix_environments}.
%\zhihui{``the''?} Policy \zhihui{``takes''?} suboptimal actions due to the misleading Q-value drop. Eventually, the policy becomes random again \zhihui{why?}, which is reflected as the re-rising of the entropy curve. 
%5) The interaction of policy randomness and the decrease of the Q-value curve \zhihui{``curve'' will not lead to local optima} leads to learning falling \zhihui{drives the policy network} into a local optimum, and the performance is no longer improved \zhihui{upon the interaction of xxx and xxx, xxx}.

To better understand why this undesirable behavior occurs, we inspect the gradient of the soft Bellman object calculated by the formula \ref{equation_q}.
\begin{equation}\label{eq:gradient}
\begin{aligned}
\hat{\nabla}_{\theta} J_{Q}(\theta)=\nabla_{\theta} Q_{\theta}(a_{t}, s_{t})(Q_{\theta}(s_{t}, a_{t})-
(r(s_{t}, a_{t})+\gamma(Q_{\theta}(s_{t+1}, a_{t+1})-
\alpha \log (\pi_{\phi}(a_{t+1} \mid s_{t+1})))).
\end{aligned}
\end{equation}
As shown in Eq.~\ref{eq:gradient}, the improvement of $Q_{\theta}(s_{t}, a_{t})$ relies on the Q-estimation of the following states and policy entropy. 
However, a sharper Q function causes the drastically shifting entropy, increasing the uncertainty of gradient updates and misleading the learning of the Q-network.
Since the soft Q-network induces the policy, the policy can also become misleading and hurt performance.
%Randomness in the policy is caused by uncertainty in $Q_{\theta}(\mathbf{s}_{t}, \mathbf{a}_{t})$, which in turn causes entropy instability. 
To mitigate this phenomenon, the key is to ensure the smoothness of policy entropy change to maintain stable training. 
In the next section, we will introduce how to constrain the policy's randomness to ensure smooth policy changes.

%\subsection{Pessimistic Exploration on Sparse Reward}
\subsection{Pessimistic Exploration}

% However, the overestimation bias in continuous domains does not come from the max operator. 
% Therefore, in this case, algorithms in continuous domains are inherently less prone to overestimation compared to DQN-based algorithms.
% Therefore, we need to revisit the use of double Q-learning in discrete domains for actor-critic algorithms.
The second failure mode comes from pessimistic exploration due to the double Q-learning mechanism. 
% In continuous domains, it is believed that the Q-value always suffer from the overestimation bias. 
% Previous work of SAC adopt clipped double Q-learning to mitigate the overestimation issues~\citep{haarnoja2018sacv1}.
To address the issue of overestimation in DQN, double Q-learning was proposed. This approach mitigates the problem by employing two independent Q-networks, and using the minimum value between them as the final Q-value. 
The concept was initially introduced by Double DQN~\citep{van2016ddqn} in the discrete domain. In the continuous domain, TD3~\citep{fujimoto2018td3} and SAC~\citep{haarnoja2018sacv1} also adopt clipped double Q-learning to mitigate overestimation, making it a favored technique across various reinforcement learning algorithms.

% The double Q-learning trick has been widely used in value-based or actor-critic RL algorithms for both discrete (e.g., Double DQN~\citep{van2016ddqn}) and continuous (e.g., SAC~\citep{haarnoja2018sacv1}) domains. 
% Due to the max operator, DQN tends to suffer from overestimation bias in discrete domains. Double DQN uses the double Q-learning trick to mitigate this issue.
% In continuous domains, inspired by Double DQN, TD3~\citep{fujimoto2018td3} and SAC adopt clipped double Q-learning to mitigate overestimation. 
% However, in SAC, the learning objective does not involve either max operator or Q-value maximization which are considered as the source of overestimation bias in \citep{doubledqn???,fujimoto2018td3}. 
Empirical results demonstrate that the clipped double Q-learning trick can boost SAC performance in continuous domains, but its impact remains unclear in discrete domains. Therefore, we need to revisit clipped double Q-learning for discrete SAC. 

In our experiments, in discrete domains, we find that discrete SAC tends to suffer from underestimation bias instead of overestimation bias. 
%While SAC focuses primarily on overestimation bias in the continuous domain, we discover that SAC suffers from underestimating bias in the discrete domain.
This underestimation bias can cause pessimistic exploration, especially in the case of sparse reward.
Here, we illustrate how the popularly used clipped double Q-learning trick causes underestimation bias and how the policy used with this trick tends to converge to suboptimal actions for discrete action spaces. 
%when the standard discrete SAC clipped double Q-learning as suggested in \citep{fujimoto2018td3}. 
%This predicament occurs because the current practice of SAC selects the pessimistic estimate of future in the event of sparse reward, which exacerbates underestimation bias. 
Our work complements previous work with a more in-depth analysis of clipped double Q-learning. 
We demonstrate the existence of underestimation bias and then illustrate its impact on Atari games. %in which the underestimation bias issue can be exacerbated in the case of sparse reward.

To analyze the estimated bias $\epsilon$, we introduce the mathematical expression of the soft state-value function:
% \begin{equation}
%     \begin{aligned}
%     V(s_{t}) = r_{s_{t},a_{t}} + \gamma \mathbb{E}_{s_{t+1},a_{t+1}} [Q_{\theta}(s_{t+1},a_{t+1})-\alpha \log (\pi(a_{t+1} \mid s_{t+1}))],
%     \end{aligned}
% \end{equation}
\begin{equation}
    \begin{aligned}
    V(s_{t}) =  \mathbb{E}_{a_{t} \sim \pi}[Q(s_{t},a_{t})-\alpha \log (\pi(a_{t} \mid s_{t}))],
    \end{aligned}
\end{equation}
where $Q(s_t, a_t)$ represents the true Q-value. In practice, SAC uses the clipped double Q-learning trick. The learning target of soft state-value function can be written as: %  (which by calculating the true discounted calculation return sum from the next state to the end of the trajectory).
%Given current soft Q-function parameters $\theta$, let $\theta^{\prime}$ define the parameters from the critic update by clipped double Q-learning.
\begin{equation}
    \begin{aligned}
    V_{appox}(s_{t}) =  \mathbb{E}_{a_{t} \sim \pi}  \min _{i=1,2}[Q_{\theta^{\prime}_{i}}(s_{t},a_{t}) -\alpha \log (\pi(a_{t} \mid s_{t}))] ,
    \end{aligned}
\end{equation}
where $Q_{\theta_{i}^{\prime}}$ represents estimation of target-critic networks parameterized by $\theta_{i}^{\prime}$. 
The estimated bias for $Q^{\prime}_{\theta_i}$ can be calculated as $\epsilon_i = Q_{\theta^{\prime}_{i}}(s,a) - Q(s,a)$. 
On the one hand, when $\epsilon_1 > \epsilon_2 > 0$, the clipped double Q-learning trick can help mitigate overestimation error due to the $min$ operation. 
%takes $Q_{\theta_{2}}$ to mitigate overestimation error. 
On the other hand, when $\epsilon_1 < \epsilon_2 < 0$ or $\epsilon_1 < 0 < \epsilon_2$, the clipped double Q-learning trick will lead to underestimation (i.e., $V_{appox} < V$) and consequently result in pessimistic exploration~\citep{pan2020softmax,ciosek2019oac}.
\begin{figure} [!t]
    \centering
    \subfigure[Discrete SAC\label{fig_underestimation_a}]{
        \begin{minipage}[b]{0.31\textwidth}
        \includegraphics[width=0.98\textwidth]{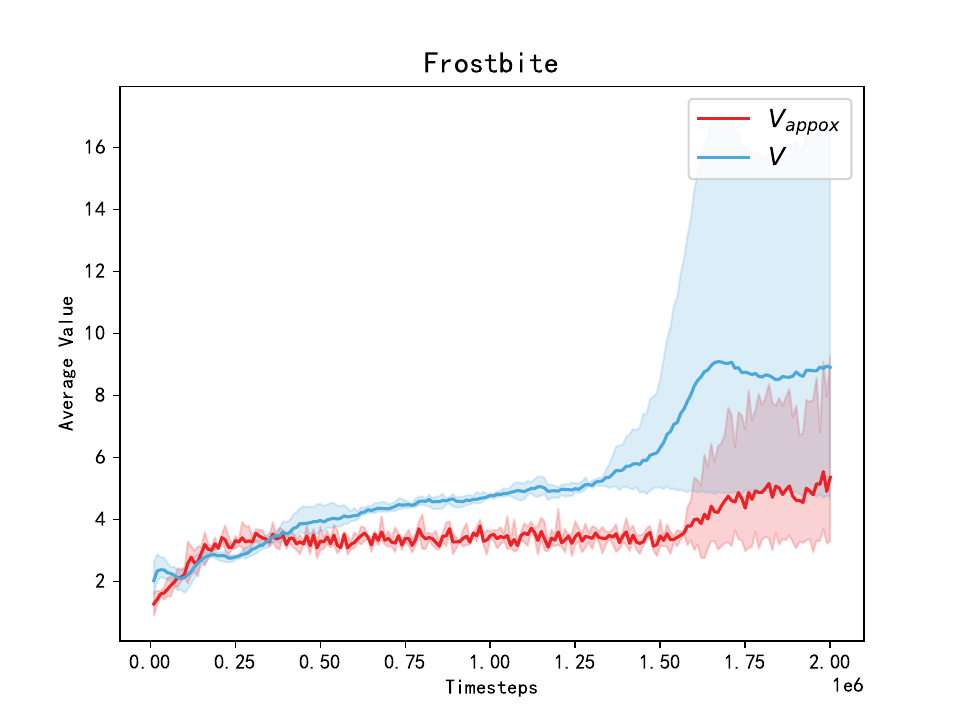}\\
	\includegraphics[width=0.98\textwidth]{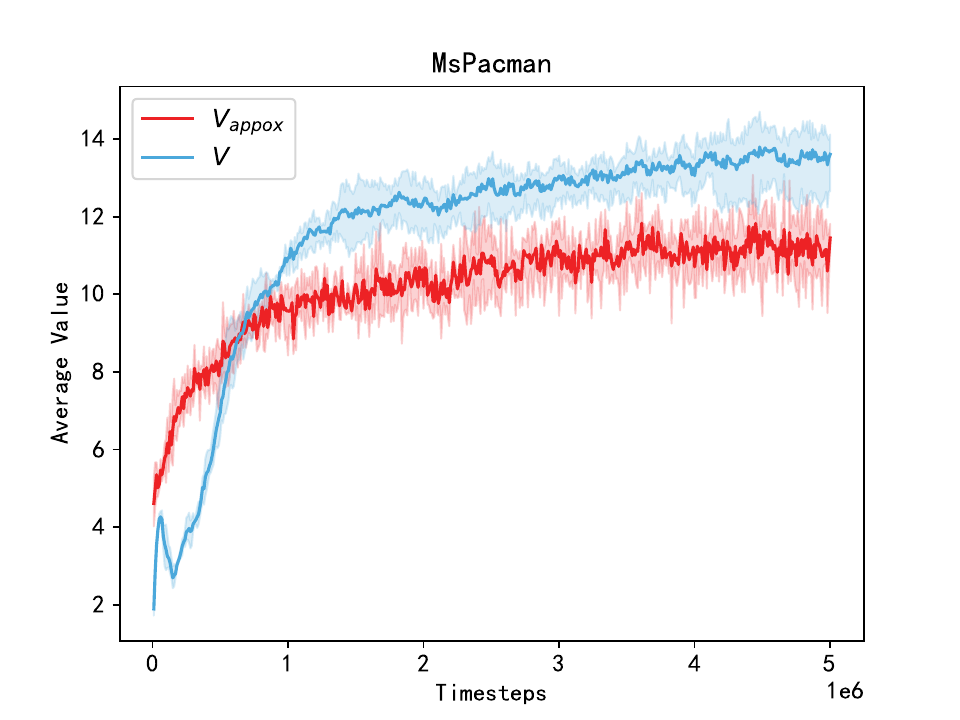}
        \end{minipage}
    }
    \subfigure[Single Q\label{fig_underestimation_b}]{
        \begin{minipage}[b]{0.31\textwidth}
        \includegraphics[width=0.98\textwidth]{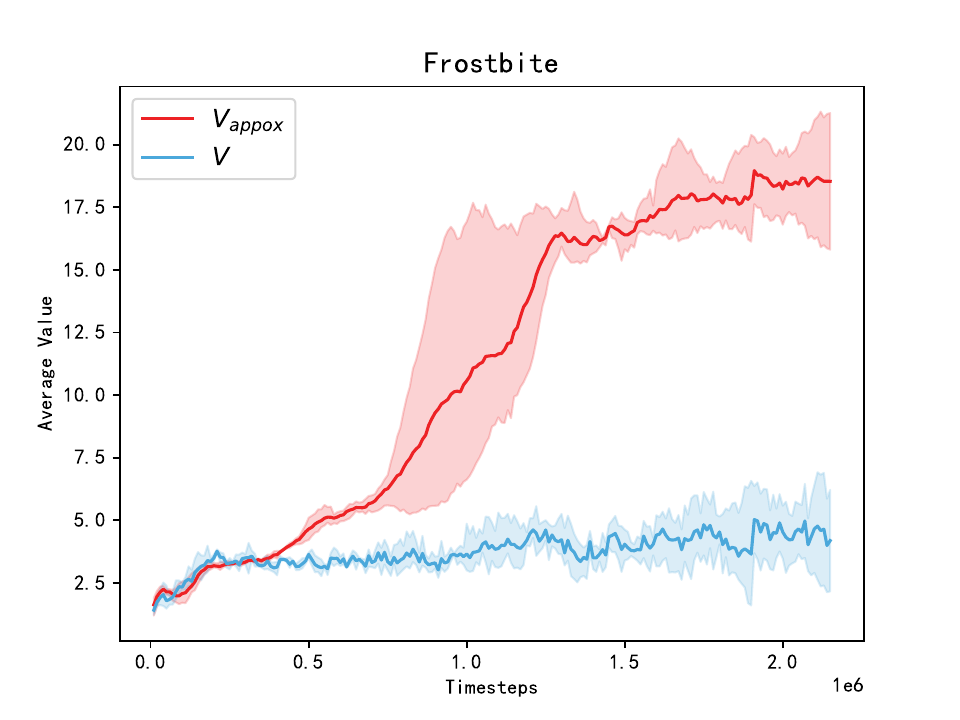}\\
	\includegraphics[width=0.98\textwidth]{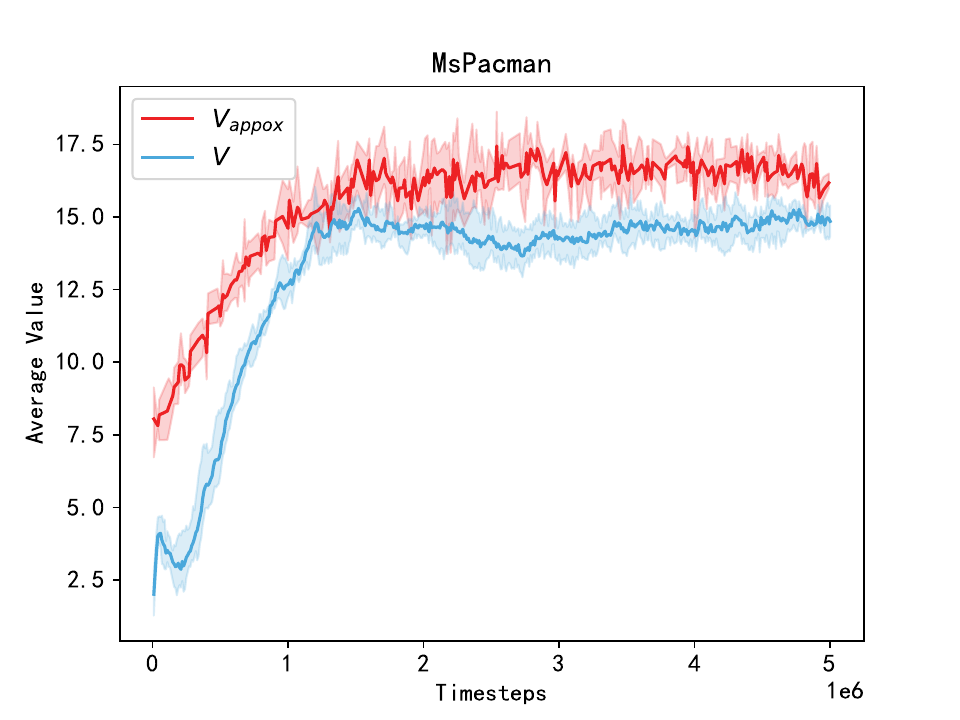}
        \end{minipage}
    }
    \subfigure[Score\label{fig_underestimation_c}]{
        \begin{minipage}[b]{0.31\textwidth}
        \includegraphics[width=0.98\textwidth]{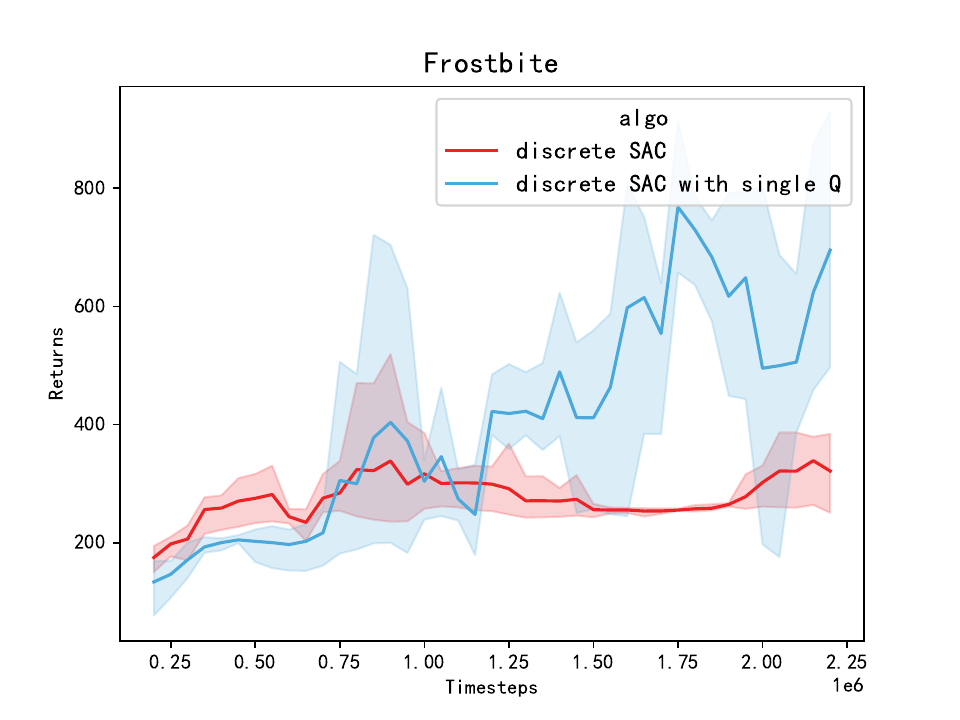}\\
	\includegraphics[width=0.98\textwidth]{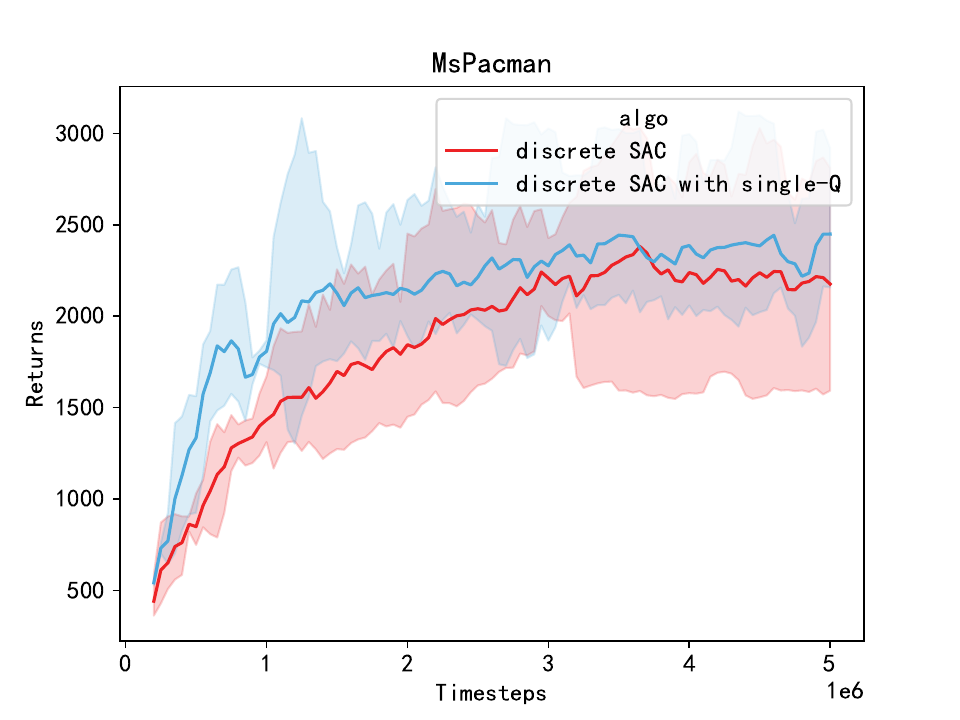}
        \end{minipage}
    }
\caption{The results of Atari game Frostbite/MsPacman environment over 2/5 million time steps: a) Measuring Q-value estimates of discrete SAC; b) Measuring Q-value estimates of discrete SAC with single Q; c) Score comparison between discrete SAC and discrete SAC with single Q.}
\label{fig_underestimation}
\end{figure}

Does this theoretical underestimate occur in practice for discrete SAC and hurt the performance? We answer this question by showing the influence of the clipped double Q-learning trick for discrete SAC in Atari games, as shown in Fig.~\ref{fig_underestimation}. 
Here, we compare the true value to the estimated value. The results are averaged over three independent experiments with different random seeds. 
We find that, in Fig.~\ref{fig_underestimation_a}, the approximate values are lower than the true value over time, demonstrating the underestimation bias issue. 
At the same time, we also run experiments for discrete SAC with a single Q (DSAC-S), which uses a single Q-value for bootstrapping instead of clipped double Q-values. As shown in Fig.~\ref{fig_underestimation_b}, without the clipped double Q-learning trick, the estimated value of DSAC-S is higher than the true value and thus has an overestimation bias. 
However, in Fig.~\ref{fig_underestimation_c}, we discover that even though DSAC-S suffers from overestimation bias, it performs much better than discrete SAC which adopts the clipped double Q-learning mechanism. This indicates that the clipped double Q-learning trick can lead to pessimistic exploration issues and hurt the agent's performance.

%The phenomenon here is different from that in continuous domains where the Q-value is known to suffer from overestimation bias~\citep{fujimoto2018td3}. 

%\textcolor{blue}{We need to show how underestimation bias leads to pessimistic exploration.}
%Even some researches~\citep{ciosek2019oac} have shown 
%The idea of underexploration is also 
%average Q-value estimate from three experiments using various seeds and an estimate of the true value. The true value is estimated using the average discounted return over 3 experiments following the current policy, starting from states sampled from the replay buffer.
%\textcolor{red}{Has this been solved or alleviated by existing work? If not, say something like: exsiting works have attempted to xxxx. However, this issue lying in discrete SAC has remained to be unsolved? -- which is where our work comes in. }

%\textcolor{blue}{we need to figure to show underestimation bias -> pessimistic exploration!}

%Additionally, we examine the circumstances in which underestimating affects performance. When \textcolor{red}{environment reward} become sparse, the soft Bellman value degenerates into the sum of the lower bound of the target-critic network and the entropy. Underestimation of critic leads to suboptimal Q-value, further confusing the gradient update of policy. Then policy becomes a random policy, and entropy term increases to maintain the training stability of Q-value update iteration. After several training iterations, performance drop due to the policy \textcolor{red}{becomes randomly}.

 \section{Improvements of SAC Failure Modes} \label{section_improvement}
 
We provide two simple alternatives, which are the surrogate objective with entropy-penalty and double average Q-learning with Q-clip,
to avoid the two failure modes of discrete SAC discussed in Section \ref{section_failure_modes}. Combining these two modifications, we propose stable discrete SAC (SD-SAC).
% While both alternative design choices have been previously proposed in the RL literature, they have been largely ignored in combination with the discrete sac algorithm.

\subsection{Entropy-Penalty}

The drastic change of Q function distribution and entropy affects the optimization of the Q-value. Due to the mutual coupling of the Q function and policy training in discrete SAC, we optimize policy entropy to alleviate the unstable effect on training caused by a sharp Q function distribution and a rapid drop in entropy. Simply removing the entropy term will injure the exploration ability under the framework of maximum entropy RL. 
An intuitive solution is to introduce an entropy penalty in the objective of policy to avoid entropy chattering. 
We will introduce how to incorporate the entropy penalty in the learning process for the discrete SAC algorithm.

Recall the objective of policy in discrete SAC as in Eq.~\ref{equation_policy_update}. 
% \begin{equation*}
% J_{\pi}(\phi)=\mathbb{E}_{s_{t} \sim D}[\mathbb{E}_{a_{t} \sim \pi_{\phi}}[\alpha \log (\pi_{\phi}(a_{t} \mid s_{t}))-Q_{\theta}(s_{t}, a_{t})]]
% \end{equation*}
For a mini-batch transition data pair $(s_{t}, a_{t},r_{r},s_{t+1})$ sampled from the replay buffer, we add an extra entropy term $\mathcal{H}_{\pi_{old}}$ to the transition tuple which reflects the randomness of policy (i.e., $(s_{t}, a_{t},r,s_{t+1},\mathcal{H}_{\pi_{old}})$), where $\pi_{old}$ denotes the policy used for data sampling. 
%The action distribution $pi$ for the current state is recalculated by current policy during the training phase. At this point, we measure the randomness of $pi$ and then explicitly add the entropy distances between collecting data phase and training phase to policy update objective.
We calculate the entropy penalty by measuring the distance between $\mathcal{H}_{\pi_{old}}$ and $\mathcal{H}_\pi$.
Formally, the objective of the policy is as follows:

\begin{equation}
\begin{aligned}
%J_{\pi}(\phi)&=\mathbb{E}_{s_{t} \sim D}[\mathbb{E}_{a_{t} \sim \pi_{\phi}}[\alpha \log (\pi_{\phi}(a_{t} \mid s_{t}))-Q_{\theta}(s_{t}, a_{t})]] + \beta \cdot \frac{1}{2}(\mathbb{H}_{\pi_{old}} - \mathbb{H}_{\pi})^{2},
J_{\pi}(\phi)&=\mathbb{E}_{s_{t} \sim D}[\mathbb{E}_{a_{t} \sim \pi_{\phi}}[\alpha \log (\pi_{\phi}(a_{t} \mid s_{t}))-Q_{\theta}(s_{t}, a_{t})]] \\&+ \beta \cdot \frac{1}{2}  
        \mathbb{E}_{s_{t} \sim D} ([ \mathbb{E}_{a_{t} \sim \pi_{\phi_{old}}} [-  \log(\pi_{\phi_{old}})  ] - \mathbb{E}_{a_{t} \sim \pi_{\phi}} [ - \log(\pi_{\phi}) ]
    )^{2},
%\beta \cdot \frac{1}{2}(\mathbb{H}_{\pi_{old}} - \mathbb{H}_{\pi})^{2},
\end{aligned}
\end{equation}

% \begin{equation}
%     \mathbb{H}_{\pi_{old}} = \mathbb{E}_{s_{t} \sim D} [ - \pi_{old} \textup{log} \pi_{old} ]
%     \beta \cdot \frac{1}{2}( 
%         \mathbb{E}_{s_{t} \sim D} [ - \pi_{old} \textup{log} \pi_{old} ] - \mathbb{E}_{s_{t} \sim D} [ - \pi_{\phi} \textup{log} \pi_{\phi} ]
%     )^{2}
% \end{equation}
where $\mathbb{E}_{a_{t} \sim \pi_{\phi_{old}}} [-  \log(\pi_{\phi_{old}})  ]$ represents policy entropy of $\pi_{\phi_{old}}$, $\mathbb{E}_{a_{t} \sim \pi_{\phi}} [ - \log(\pi_{\phi}) ]$ represents policy entropy of $\pi_{\phi}$, and $\beta$ denotes a coefficient for the penalty term and is set to 0.5 in this paper. 
By constraining the policy objective with this penalty term, we increase the stability of the policy learning process.
%Bear in mind that the entropy penalty is only used to prevent drastic changes in policy, rather than limiting the process of policy from random to deterministic. Our proposed method can improve training stability while maintaining the final effect of the policy.

% \begin{figure}[!t]

% \subfigure{
% 		\includegraphics[width=0.15\textwidth]{images/analysis/improvement_of_fail_mode_1/Asterix_entropy.jpeg}
% }
% \subfigure{
% 		\includegraphics[width=0.15\textwidth]{images/analysis/improvement_of_fail_mode_1/Asterix_q1.jpeg}
% }
% \subfigure{
% 		\includegraphics[width=0.15\textwidth]{images/analysis/improvement_of_fail_mode_1/Asterix_reward.jpeg}
% }

% \caption{Measuring policy action entropy, estimation of q-value and final evaluation reward on Atati Game Asterix environment compared between baseline-SAC-discrete and SAC-discrete with entropy-penalty over 10 million time steps}
% \label{fig_entropy_improvements_analysis}
% \end{figure}
\begin{figure} [!t]
    \centering
        \subfigure[Q Function Variance]{
		\includegraphics[width=0.40\textwidth]{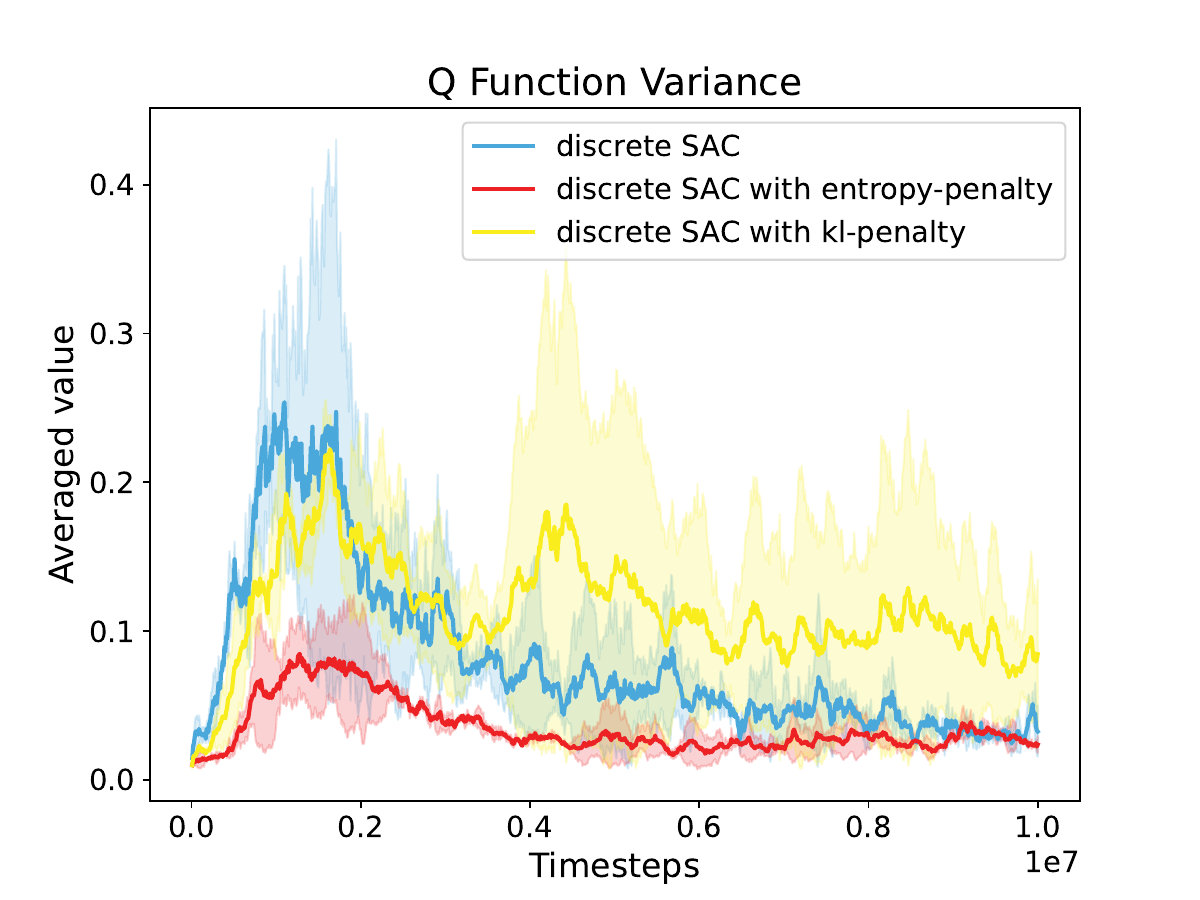}
		\label{subfig_improve_qvar}
		}
	\subfigure[Entropy]{
		\includegraphics[width=0.40\textwidth]{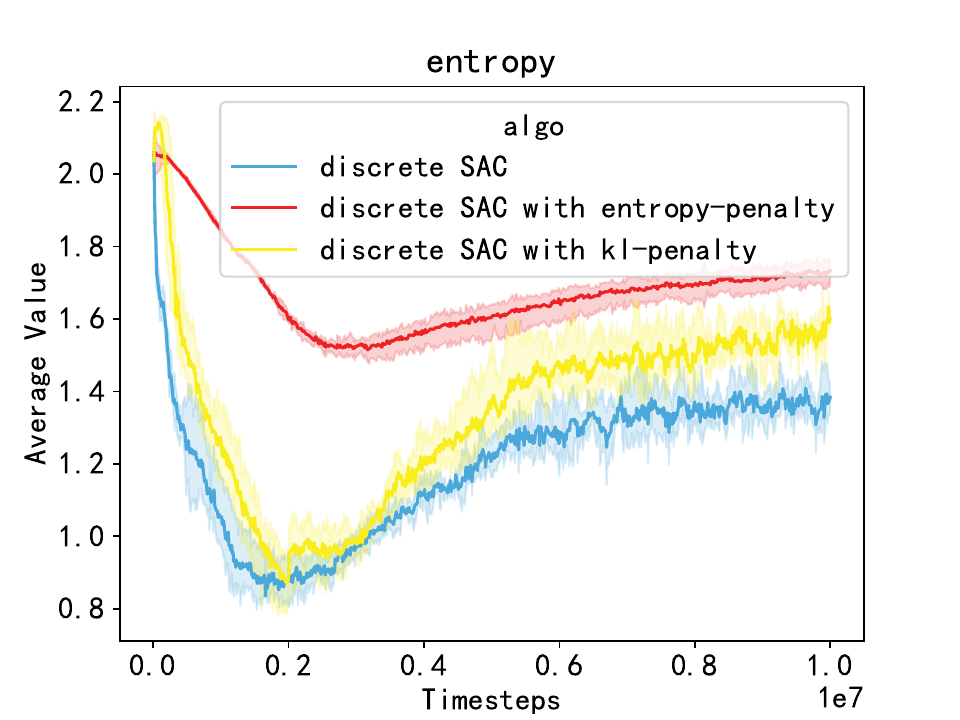}
		\label{subfig_improve_entropy}
		}
	\subfigure[Q-value]{
		\includegraphics[width=0.40\textwidth]{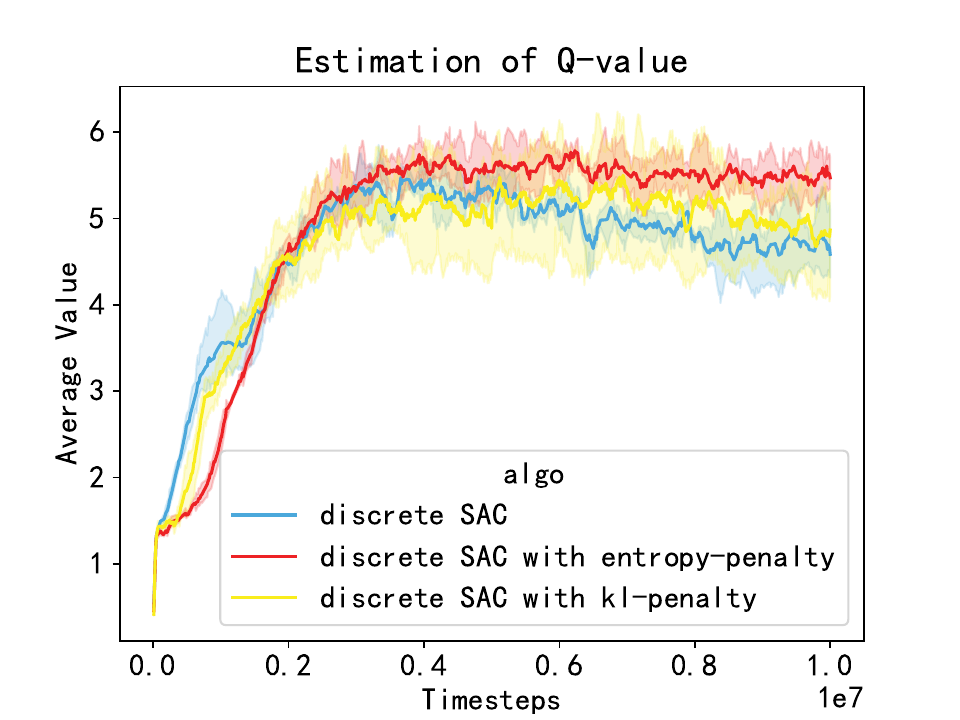}
		\label{subfig_improve_qvalue}
		}
	\subfigure[Score]{
		\includegraphics[width=0.40\textwidth]{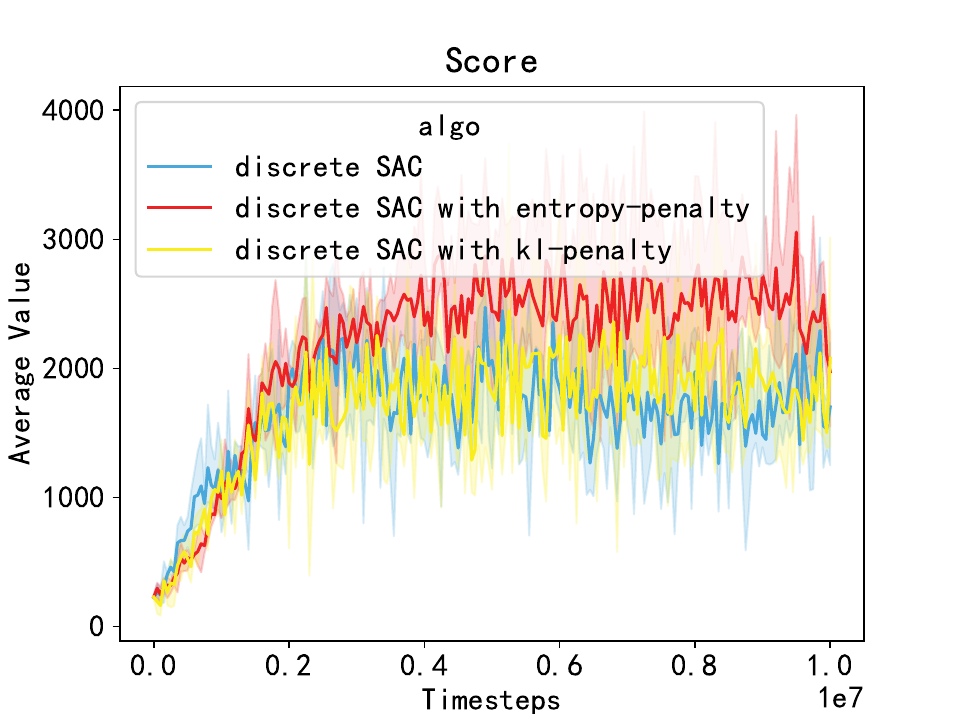}
	   \label{subfig_improve_return}	
        }
	\caption{Measuring Q function variance, policy action entropy, estimation of Q-value, and score on Atari game Asterix compared between discrete SAC, discrete SAC with KL-penalty and discrete SAC with entropy-penalty over 10 million time steps.}
	\label{fig_entropy_improvements_analysis} 
\end{figure}

% A reliable off-policy training process can be described as: 1) The critic updates the estimate with the reward when a stochastic policy explored the reward state. 2) At this point, the policy's behavior becomes deterministic, and the entropy is steadily declining. Additionally, the Q-value is progressively rising. 3) Finally, the test reward gradually increases over iterations.

% Figure \ref{fig_entropy_improvements_analysis} shows the training curves of how the entropy penalty mitigates the failure mode of policy drastic change. In entropy curve Figure \ref{subfig_improve_entropy}, the addition of the entropy-penalty constrains the limit of the randomness of the policy. The learning curve of the Q value grows steadily as the result of the relative stability of the entropy term. Finally, the SAC with entropy-penalty continuously increasing during the training period as opposed to the baseline, where the policy swings and the reward eventually reaches its maximum. The illustration in Figure \ref{fig_entropy_improvements_analysis} demonstrates the efficacy of the entropy-penalty mechanism. In the following section \ref{section_experiments}, we will present further experimental findings and analyses.
Fig.~\ref{fig_entropy_improvements_analysis} shows the training curves to demonstrate how the entropy penalty mitigates the failure mode of policy drastic change. 
In Fig.~\ref{subfig_improve_entropy}, the entropy of discrete SAC (the blue curve) drops quickly, and the policy falls into a local optimum at the early training stage. Later, the policy stops improving and even suffers from performance deterioration, as shown in the blue curves in Fig.~\ref{subfig_improve_qvalue} and Fig.~\ref{subfig_improve_return}. 

On the contrary, our proposed method (i.e., discrete SAC with entropy-penalty) demonstrates better stability than discrete SAC. 
As shown in Fig.~\ref{subfig_improve_qvar}, entropy penalty effectively constrains the sharpness of Q function, as a result, the policy changes smoothly during training (Fig.~\ref{subfig_improve_entropy}). Consequently, compared with discrete SAC, the policy in our approach can keep improving during the whole training stage and does not suffer from a performance drop at the later training stage (the red curves in Fig.~\ref{subfig_improve_qvalue} and Fig.~\ref{subfig_improve_return}). 

It is worth noting that, since the instability in training mainly manifests as the existence of policy entropy term in optimization, imposing constraints in the entropy space is more effective than constraints in the policy space. Other common methods, such as the KL penalty, limit the magnitude of policy updates and impose additional restrictions on policy updates. This is proved in experiments: KL penalty (the yellow curve) cannot effectively constrain the rise in Q variance (Fig.~\ref{subfig_improve_qvar}) and the decrease in entropy (Fig.~\ref{subfig_improve_entropy}). Consequently, the final Q-value and score of the KL penalty are lower than those with the entropy penalty, with a difference of 12\% and 23\%, respectively.
%The learning curve of the Q value grows steadily as a result of the relative stability of the entropy term. Finally, the discrete SAC with entropy-penalty continuously increases during the training period as opposed to the discrete SAC, where the policy swings and the reward no longer grows.

The entropy-penalty term $\frac{1}{2} \mathbb{E}_{s_{t} \sim D} ([ \mathbb{E}_{a_{t} \sim \pi_{\phi_{old}}} [-  \log(\pi_{\phi_{old}})  ] - \mathbb{E}_{a_{t} \sim \pi_{\phi}} [ - \log(\pi_{\phi})])^{2}$, in conjunction with the temperature $\alpha$, jointly regulates the exploration of policy. Different from other trust region methods such as KL constraint~\citep{bach2015trpo} or clipping surrogate objective~\citep{schulman2017ppo}, our method penalizes the change of action entropy between old and new policies to address policy instability during training. By adding regularization in entropy space instead of policy space, our method can mitigate the drastic changes of policy entropy while maintaining the inherent exploratory ability of discrete SAC (as shown in Fig.~\ref{subfig_improve_entropy}, the policy entropy changes smoothly. It keeps at a relatively high value to encourage exploration).
%aligning better with the optimization objectives of SAC with maximum entropy. 
%Additionally, the penalty on action entropy instead of policy update constraint allows SAC to retain its inherent exploratory ability (as depicted in Fig.~\ref{subfig_improve_entropy} , where entropy exhibits a stable ascending trend during exploration).}

\subsection{Double Average Q-learning with Q-clip}

While several approaches\citep{ciosek2019oac,pan2020softmax} have been proposed to reduce underestimation bias, they are not straightforward to be applied to discrete SAC due to the use of Gaussian distribution.
In this section, we introduce a novel variant of double Q-learning to mitigate the underestimation bias for discrete SAC.

In practice, discrete SAC uses clipped double q-learning with a pair of target critics ($Q_{\theta_{1}^{\prime}}$, $Q_{\theta_{2}^{\prime}}$), and the learning target of these two critics is:
%where actor $\pi$ is optimized with respect to double critic:
\begin{equation}
    \begin{aligned}
    y=r+\gamma \min _{i=1,2} Q_{\theta_{i}^{\prime}}(s^{\prime}, \pi(s^{\prime})).
    \end{aligned}
\end{equation}

When neural networks approximate the Q-function, there exists an unavoidable bias in the critics. Since policy is optimized concerning the low bound of double critics, for some states, we will have $Q_{\theta_{2}^{\prime}}(s, \pi_{\phi}(s))> Q_{true} >Q_{\theta_{1}^{\prime}}(s, \pi_{\phi}(s))$. This is problematic because $Q_{\theta_{1}^{\prime}}(s, \pi_{\phi}(s))$ will generally underestimate the true value, and this underestimated bias will be further exaggerated during the whole training phase, which results in pessimistic exploration. 

To address this problem, we propose to mitigate the underestimation bias by replacing the \textit{min} operator with \textit{avg} operator. 
%simply balance the under-biased estimate $Q_{\theta_{1}^{\prime}}$ by the less under-biased value estimate $Q_{\theta_{2}^{\prime}}$. 
This results in taking the average between the two estimates, which we refer to as \textit{double average Q-learning}:
\begin{equation}
\label{eq:avgq_target}
    \begin{aligned}
y=r+\gamma \cdot \textup{avg} ( Q_{\theta_{1}^{\prime}}(s^{\prime}, \pi(s^{\prime})) , Q_{\theta_{2}^{\prime}}(s^{\prime}, \pi(s^{\prime})) ).
    \end{aligned}
\end{equation}
By doing so, the other critic can mitigate the underestimated bias of the lower bound of double critics.  
%The risk of overestimation cannot be avoided by taking the average between the two estimates. we add dual clip function on estimation of Q-value to prevent massive overestimation.
To improve the stability of the Q-learning process, inspired by value clipping in PPO~\citep{schulman2017ppo}, we further add a clip operator on the Bellman error to prevent drastic updates of the Q-network. The modified Bellman loss of Q-network is as follows:
\begin{equation}
\label{eq:clipq}
%\small
    \begin{aligned}
    %Q_{\theta_{i}}(s,a) = \operatorname{clip}(Q_{\theta{i}^{\prime}}(s,a),-c, c)
    \mathcal{L}(\theta_i) = \textup{max} \left( 
    (Q_{\theta_i} - y)^2, 
    (Q_{\theta^{\prime}_{i}} + \textup{clip}(Q_{\theta_i} - Q_{\theta^{\prime}_{i}}, -c, c) )  - y)^2 
    \right),
    \end{aligned}
\end{equation}
where $Q_{\theta_{i}}$ represents the critic network's estimate, $Q_{\theta{i}^{\prime}}$ represents estimation of target-critic networks, and $c$ is the hyperparameter denoting the clip range. 
% We have a previous Q-value estimation $Q_{\theta{i}^{\prime}}(s,a)$ of the current state since the target-critic network is a delayed copy of critic network. 
% We may roughly estimate the acceptable range of the current Q-value using the prior estimate as a reference. 
This clipping operator prevents the Q-network from performing an incentive update beyond the clip range. In this way, the Q-learning process is more robust to the abrupt change in data distribution.
%The clipping mechanism avoids overestimation bias on single-step iteration process. 
Combining the clipping mechanism (Eq.~\ref{eq:clipq}) with double average Q-learning (Eq.~\ref{eq:avgq_target}), we refer to our proposed approach as \textit{double average Q-learning with Q-clip}.
% \begin{equation}
% \begin{aligned}
% J_{Q}(\theta_i)&=
% \mathbb{E}_{(s_{t}, a_{t}) \sim D} [
% \frac {1} {2} (Q_{\theta}(s_{t}, a_{t})-(r(s_{t}, a_{t}) + \gamma \cdot  \\ 
% &[mean(Q_{\theta_{1}^{\prime}}(s_{t+1}, \pi(s_{t+1}),Q_{\theta_{2}}(s_{t+1}, \pi(s_{t+1})) \\& -\alpha \log (\pi(a_{t+1} \mid s_{t+1}))])^{2}
% ] 
% %\\
% %& Q_{\theta}(s_{t},a_{t}) = \operatorname{clip}(Q_{\theta^{\prime}}(s,a),-c, c)
% \end{aligned}
% \end{equation}

% \begin{figure}[!t]

% \subfigure{
% 		\includegraphics[width=0.2\textwidth]{images/analysis/improvement_of_fail_mode_2/Frostbite_q1.jpeg}
% }
% \subfigure{
% 		\includegraphics[width=0.2\textwidth]{images/analysis/improvement_of_fail_mode_2/Frostbite_reward.jpeg}
% }

% \caption{Measuring estimation of q-value and final evaluation reward on Atati Game frostbite environment compared between baseline-SAC-discrete and SAC-discrete with double average q-learning with q-approximation-clip  over 10 million time steps}
% \label{fig_doubelavgq_improvement}
% \end{figure}
\begin{figure} [!t]
    \centering
        
    \subfigure[Discrete SAC \label{fig_doubelavgq_improvement_base}]{
        \begin{minipage}[b]{0.31\textwidth}
        \includegraphics[width=1\textwidth]{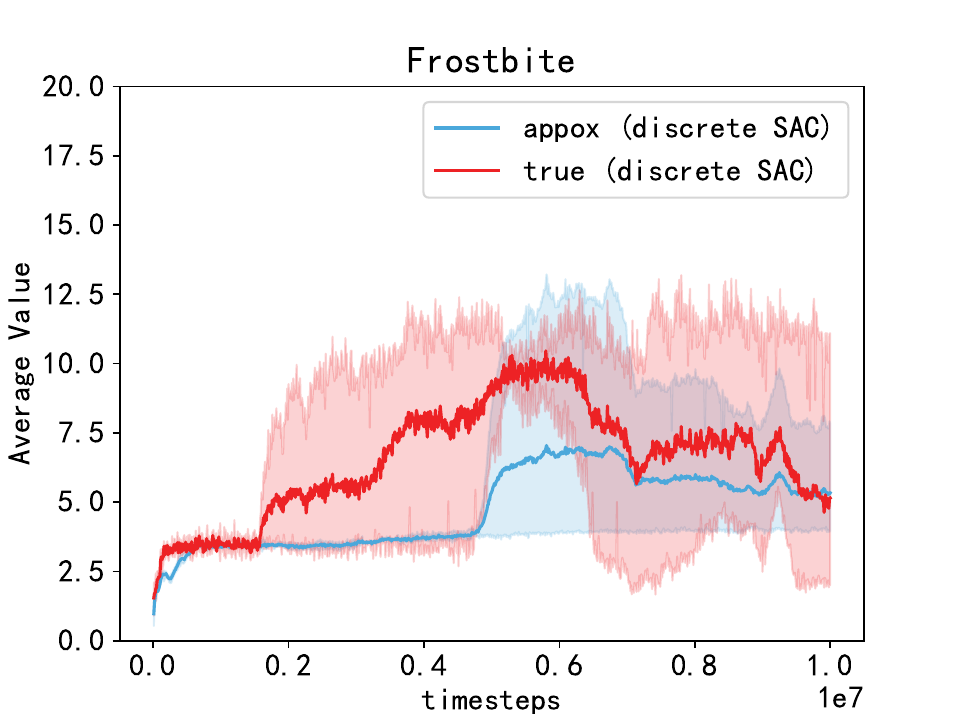}\\
	\includegraphics[width=1\textwidth]{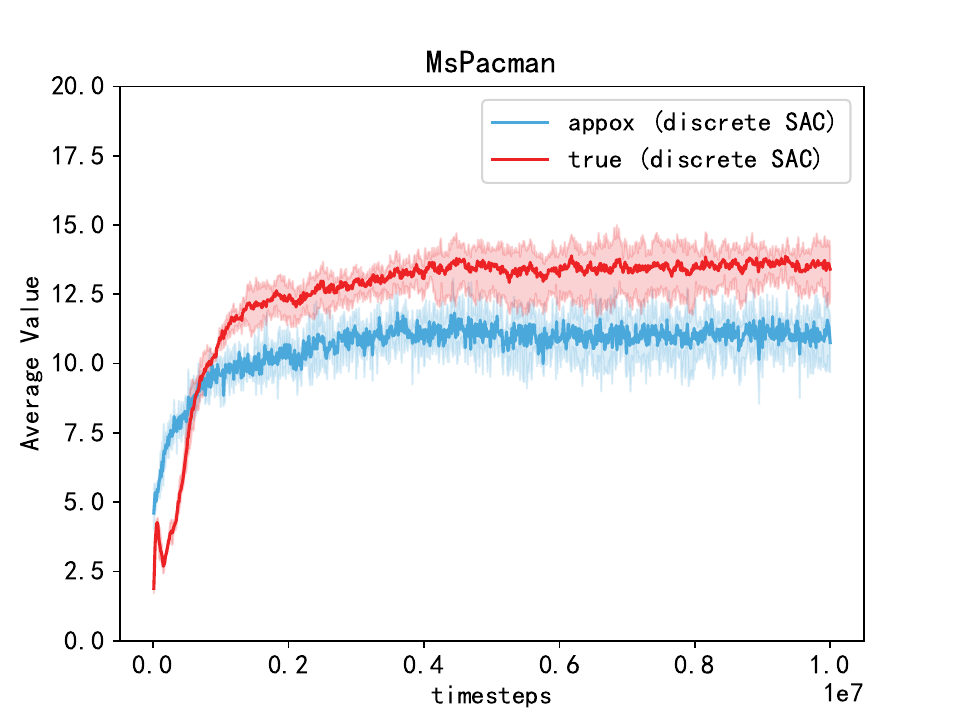}
        \end{minipage}
    }
    \subfigure[REDQ \label{fig_doubelavgq_improvement_redq}]{
        \begin{minipage}[b]{0.31\textwidth}
        \includegraphics[width=1\textwidth]{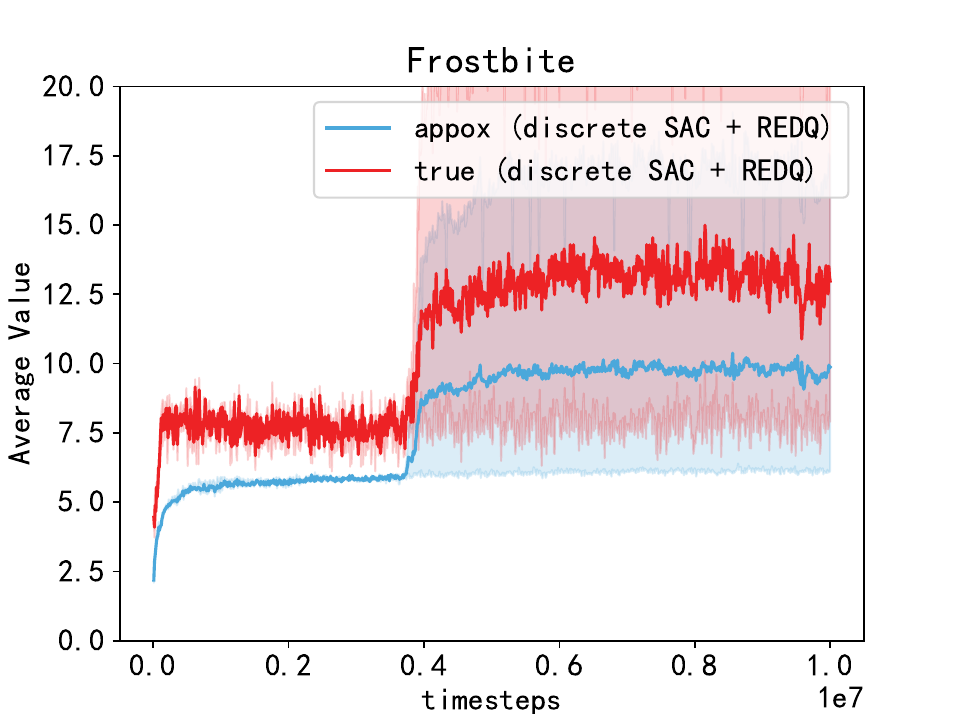}\\
	\includegraphics[width=1\textwidth]{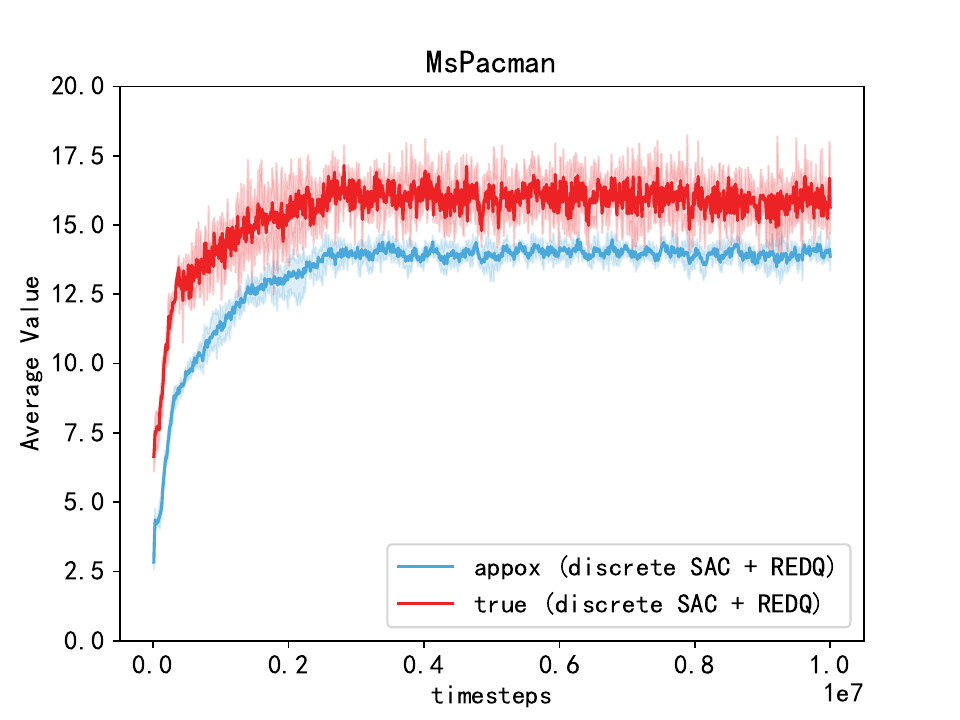}
        \end{minipage}
    }
    \subfigure[REM \label{fig_doubelavgq_improvement_rem}]{
        \begin{minipage}[b]{0.31\textwidth}
       \includegraphics[width=1\textwidth]{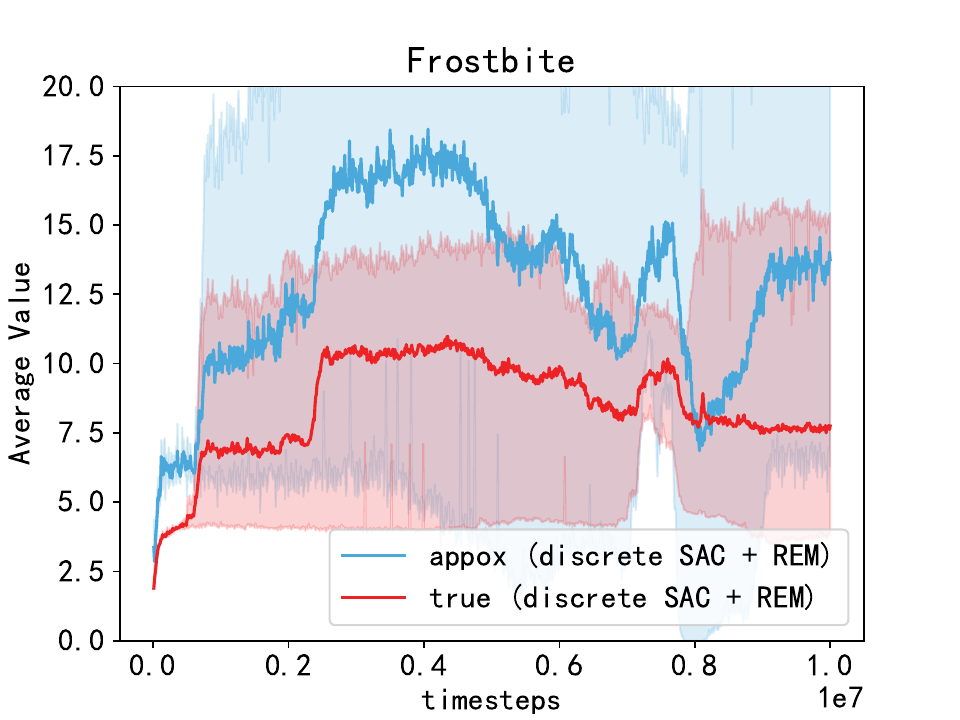}\\
	\includegraphics[width=1\textwidth]{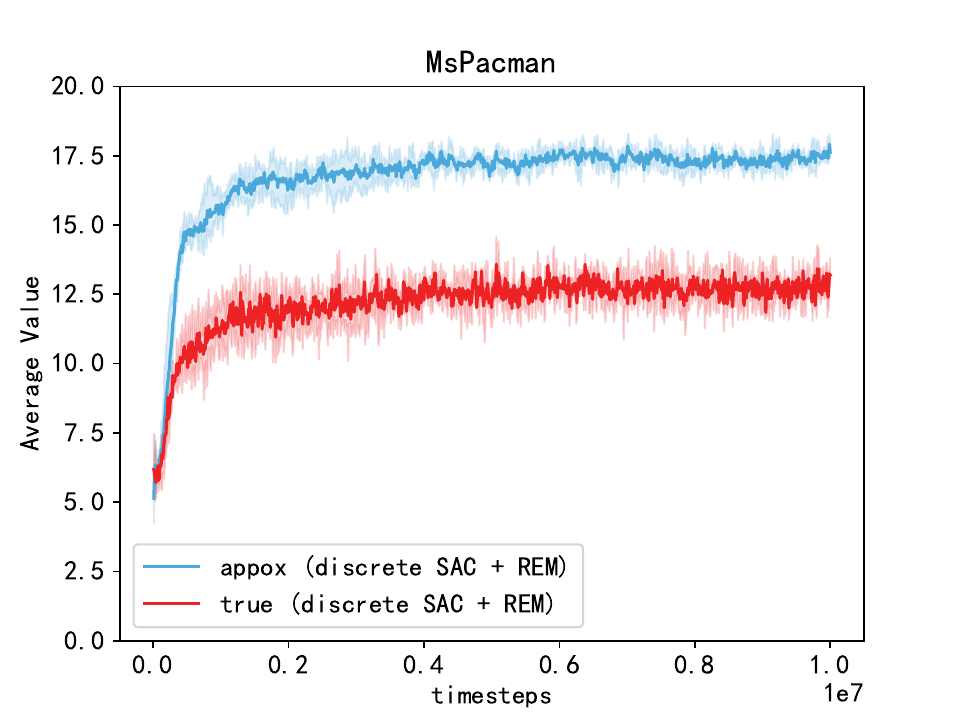}
        \end{minipage}
    }
    \subfigure[Ours \label{fig_doubelavgq_improvement_ours}]{
        \begin{minipage}[b]{0.31\textwidth}
        \includegraphics[width=1\textwidth]{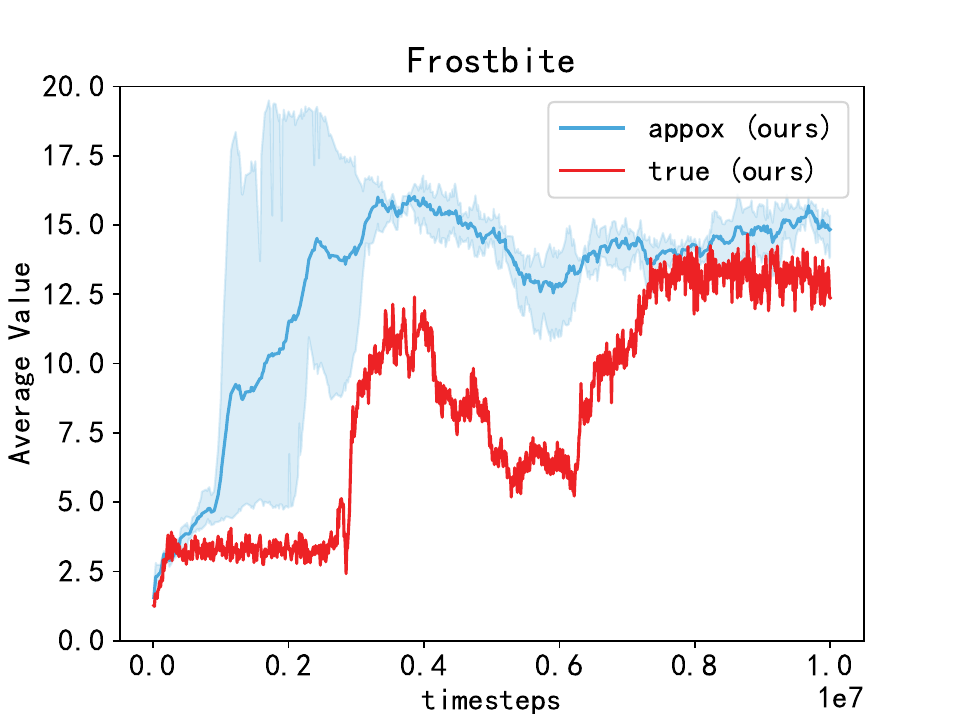}\\
	\includegraphics[width=1\textwidth]{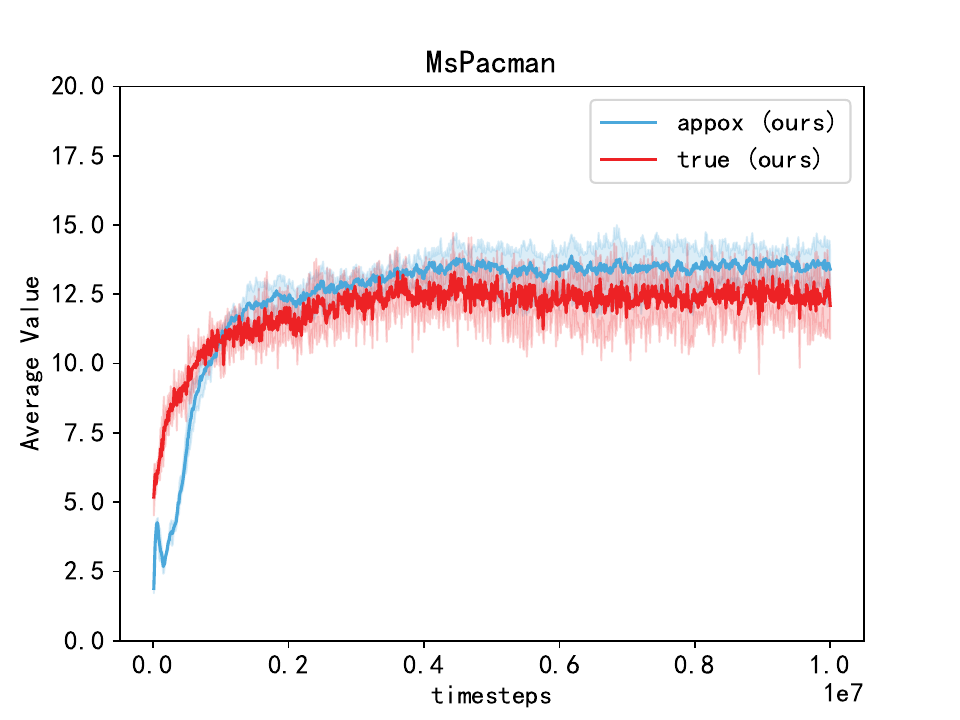}
        \end{minipage}
    }
    \subfigure[Score \label{fig_doubelavgq_improvement_score}]{
        \begin{minipage}[b]{0.31\textwidth}
        \includegraphics[width=1\textwidth]{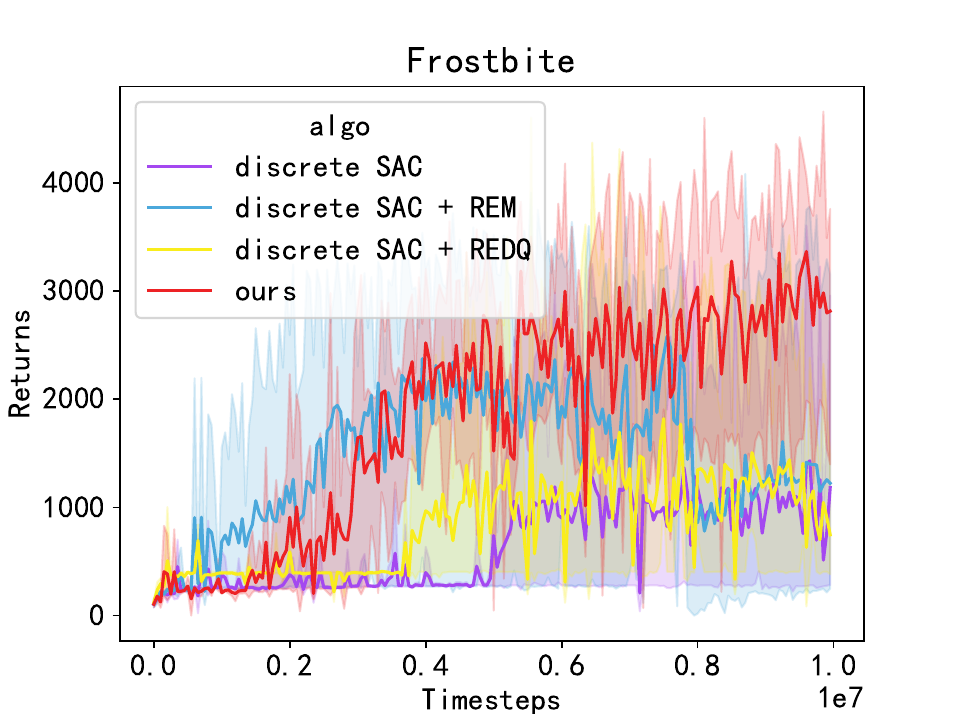}\\
	\includegraphics[width=1\textwidth]{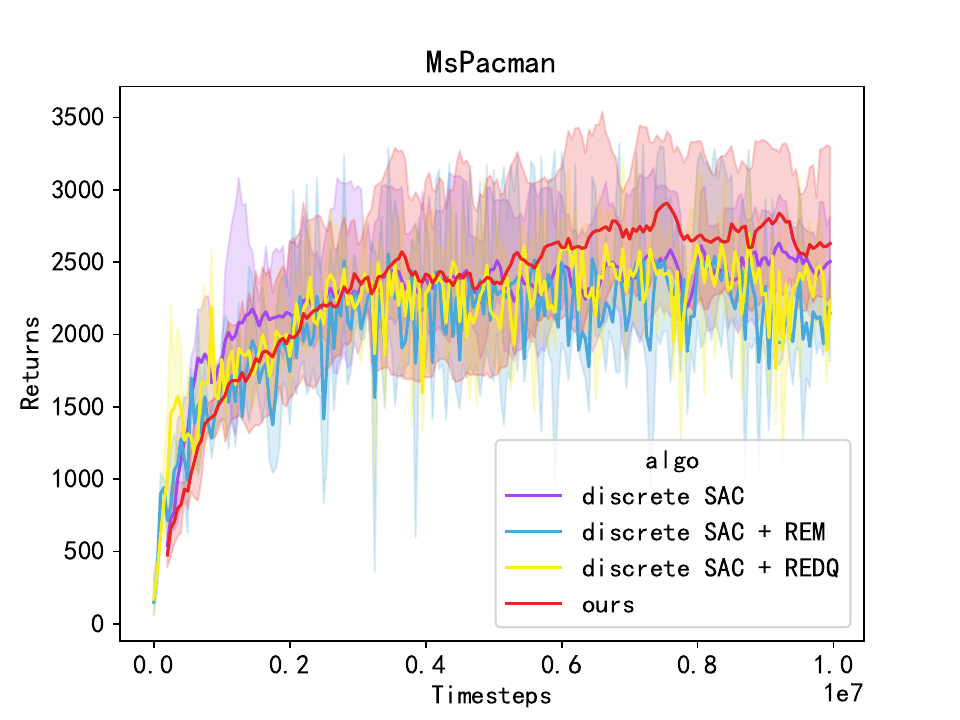}
        \end{minipage}
    }
	% \subfigure[Q-value estimate of discrete SAC]{
	% 	\includegraphics[width=0.25\textwidth]{images/analysis/improvement_of_fail_mode_2/Frostbite_base_q_estimation.pdf}
 %            \label{fig:aaa}
	% 	}	
	% \subfigure[Q-value estimate of ours]{
	% 	\includegraphics[width=0.25\textwidth]{images/analysis/improvement_of_fail_mode_2/Frostbite_avgq_q_estimation.pdf}
 %            \label{fig:bbb}
	% 	}
	% \subfigure[Score]{
	% 	\includegraphics[width=0.25\textwidth]{images/analysis/improvement_of_fail_mode_2/Frostbite_reward.pdf}
 %            \label{fig:ccc}
	% 	}

	\caption{Measuring estimation of Q-value and score on Atari Game Frostbite/MsPacman environment compared between discrete SAC, discrete SAC with REDQ, discrete SAC with REM, and ours (SD-SAC) over 10 million steps.}
	\label{fig_doubelavgq_improvement} 
\end{figure}

%Fig.~\ref{fig_doubelavgq_improvement} shows the Q value estimation and reward of double average Q learning with Q-clip. The state of pessimistic exploration is eliminated by double average Q learning, and rewards increase as a result of the optimistic Q value estimation.

Fig.~\ref{fig_doubelavgq_improvement} demonstrates the effectiveness of our approach. We compare the discrete SAC and various ensemble Q-estimation methods, including Randomized Ensembled Double Q-learning (REDQ) \cite{chen2021REDQ} and Random Ensemble Mixture (REM) \cite{AgarwalS020REM}, with our proposed method, SD-SAC. In Fig.~\ref{fig_doubelavgq_improvement_base}, the Q-value estimate of discrete SAC is underestimated than the true value. Therefore, the policy of discrete SAC suffers from pessimistic exploration and results in poor performance (purple curve in Fig.~\ref{fig_doubelavgq_improvement_score}). On the contrary, in Fig.~\ref{fig_doubelavgq_improvement_ours}, with double average Q-learning and Q-clip, the Q-value estimate eliminates underestimation bias and improves quickly at the early training stage. The improvement of Q-value carries over to the performance of policy. Consequently, our approach outperforms baseline discrete SAC by a large margin (Fig.~\ref{fig_doubelavgq_improvement_score}). The result also demonstrates that even though REDQ has less estimation bias in Fig.~\ref{fig_doubelavgq_improvement_redq}, it still suffers from underestimation bias, leading to suboptimal performance due to pessimistic exploration. Although REM addresses the underestimation issue in Fig.~\ref{fig_doubelavgq_improvement_rem}, the overestimation bias of REM significantly exceeds that of our proposed method, resulting in a rapid decline in performance at 8 million steps. In Fig.~\ref{fig_doubelavgq_improvement_ours}, we also notice that the Q-value overestimates the true value during the early training stage but finally converges to the true value after the training process. This encourages early exploration, which is consistent with the principle of optimism in the face of uncertainty~\citep{kearns2002near}.

\subsection{Psudocode}
Finally, we provide the pseudo code for SD-SAC (i.e., Stable Discrete SAC with entropy-penalty and double average Q-learning with Q-clip), as shown in Algorithm \ref{algorithmi-DSAC-ours}.

\begin{algorithm}[h]
	\caption{SD-SAC: Stable Discrete SAC with entropy-penalty and double average Q-learning with Q-clip}
	\begin{algorithmic}[]
        \State Input: $\theta_1, \theta_2, \phi$ \Comment{Initial parameters}
        \State Output: $\theta_1, \theta_2, \phi$ \Comment{Optimized parameters}
        \State Hyperparameters: $\gamma ,\beta,c,\tau $ 
        \State Initialise $Q_{\theta_1}: S \rightarrow \mathbb{R}^{|A|}, Q_{\theta_2}: S \rightarrow \mathbb{R}^{|A|}, \pi_\phi: S \rightarrow[0,1]^{|A|} $\Comment { Initialise local networks }
        \State Initialise ${Q}_{\theta^{\prime}_{1}}: S \rightarrow \mathbb{R}^{|A|}, {Q}_{\theta^{\prime}_{2}}: S \rightarrow \mathbb{R}^{|A|} \quad $ \Comment { Initialise target networks }
        \State ${\theta}_{1}^{\prime} \leftarrow \theta_1, {\theta}_{2}^{\prime} \leftarrow \theta_2 $ \Comment{ Equalise target and local network weights }
        \State $\mathcal{D} \leftarrow \emptyset $ \Comment { Initialize an empty replay buffer }
        \For {each iteration }
            \For {each environment step}
                \State $a_t \sim \pi_\phi\left(a_t \mid s_t\right)$\Comment{Sample action from the policy}
                \State $s_{t+1} \sim p\left(s_{t+1} \mid s_t, a_t\right)$\Comment{Sample transition from the environment}
                
                \State $\mathcal{H}_{\pi_{old}} \sim \underset{a \sim \pi_{\phi}(\cdot \mid s_{t})}{\mathbb{E}}[-\log \pi_{\phi}(a \mid s_{t})]$ \Comment{Calculate the entropy $\mathcal{H}_{\pi_{old}}$ of the current policy $\phi$}
                \State $\mathcal{D} \leftarrow \mathcal{D} \cup\left\{\left(s_t, a_t, r\left(s_t, a_t\right), s_{t+1},\mathcal{H}_{\pi_{old}}\right)\right\}$\Comment{Store the transition in the replay buffer}
                                
            \EndFor
            \For {each gradient step}
            \State $y\sim r\left(s_t, a_t\right) +\gamma \cdot \textup{avg} ( Q_{\theta^{\prime}_{1}}(s_{t+1}, \pi(s_{t+1})) , Q_{\theta^{\prime}_{2}}(s_{t+1}, \pi(s_{t+1})))$ \Comment{Double average Q-value estimation}
            \State $\mathcal{L}(\theta_i) \sim \textup{max} \left( 
    (Q_{\theta_i} - y)^2, (Q_{\theta^{\prime}_{i}} + \textup{clip}(Q_{\theta_i} - Q_{\theta^{\prime}_{i}}, -c, c) )  - y)^2 \right)$ for $i \in\{1,2\}$ \Comment{Clip the Q-value estimation from target critic network}
            \State $\theta_i \leftarrow \theta_i-\lambda_Q \hat{\nabla}_{\theta_i} \mathcal{L}(\theta_i)$ for $i \in\{1,2\}$ \Comment{Update the Q-function parameters}
            \State $\mathcal{H}_{\pi} \sim \underset{a \sim \pi_{\phi}(\cdot \mid s_{t})}{\mathbb{E}}[-\log \pi_{\phi}(a \mid s_{t})]$ \Comment{Calculate the entropy $\mathcal{H}_{\pi}$ of policy $\phi$}
            \State $J_{\pi}(\phi) \sim \mathbb{E}_{s_{t} \sim D}[\mathbb{E}_{a_{t} \sim \pi_{\phi}}[\alpha \log (\pi_{\phi}(a_{t} \mid s_{t}))-Q_{\theta}(s_{t}, a_{t})]] + \beta \cdot \frac{1}{2}(\mathcal{H}_{\pi_{old}} - \mathcal{H}_{\pi})^{2}$
            \State $\phi \sim \phi-\lambda_\pi \hat{\nabla}_\phi J_\pi(\phi)$ \Comment{Update policy weights}
            \State $\alpha \sim \alpha-\lambda \hat{\nabla}_\alpha J(\alpha)$ \Comment{Update temperature}
            \State ${Q}_{\theta^{\prime}_{i}} \leftarrow \tau Q_{\theta_{i}}+(1-\tau) {Q}_{\theta^{\prime}_{i}}$ for $i \in\{1,2\}$ \Comment{Update target network weights}
            \EndFor
        \EndFor
	\end{algorithmic}
	\label{algorithmi-DSAC-ours}
\end{algorithm}

\section{Experiments} \label{section_experiments}

\subsection{Experimental Setup}
To evaluate our algorithm, we compare our SD-SAC with most related baselines, i.e., discrete SAC \citep{christodoulou2019sacd}, TES-SAC \citep{xu2021targetannealing}, Soft-DQN \citep{VieillardPG20Munchausen} and Rainbow\citep{hessel2018rainbow} which is widely accepted algorithm in the discrete domain. We measure their performance in 20 Atari games chosen as the same as \citep{christodoulou2019sacd} for a fair comparison.
% After a policy is trained for every 50000 steps, its performance is immediately evaluated by running the corresponding deterministic policy for 10 episodes.
We evaluate for 10 episodes for every 50000 steps during training, and execute 3 random seeds for each algorithm for 10 million environment steps (or 40 million frames). 
For the baseline implementation of discrete-SAC, we use Tianshou \footnote{https://github.com/thu-ml/tianshou}. 
We find that Tianshou's implementation performs better than the original paper by Christodoulou \citep{christodoulou2019sacd}, thus we use the default hyperparameters in Tianshou on all 20 games.
% We use the setting of fixed $\alpha$ ($\alpha=0.05$) and 3-steps learning, both of which have been proven to be effective tricks. 

We start the game with up to 30 no-op actions, similar to \citep{mnih2013playing}, to provide the agent with a random starting position. To obtain summary statistics across games, following Hasselt \citep{van2016ddqn}, we normalize the score for each game as follows: $\text { Score }_{\text {normalized }}=\frac{\text { Score }_{\text {agent }}-\text { Score }_{\text {random }}}{\text { Score }_{\text {human }}-\text { Score }_{\text {random }}}.$
% \begin{equation}
%     \text { Score }_{\text {normalized }}=\frac{\text { Score }_{\text {agent }}-\text { Score }_{\text {random }}}{\text { Score }_{\text {human }}-\text { Score }_{\text {random }}}.
% \end{equation}

\subsection{Overall Performance}

% \begin{table}
% \begin{center}
% \scriptsize
% \caption{Mean and median scores across all 21 Atari games, measured in percentages of human performance.}
% \begin{tabular}{| c | c | c | c |}
% \hline
%   & Rainbow(10M) & Discrete SAC(10M) & Ours(10M) \\ 
% \hline
% Mean   &  187.4 \%  &  151.4 \% & \textbf{220.0} \%\\
% \hline
% Median &  79.2 \%  &  90.8 \% & \textbf{114.1} \% \\
% \hline 
% \end{tabular}
% \end{center}
% \label{table-score-humannorm-rainbow}
% \end{table}

% \begin{table}[t]
% \centering
% \caption{Mean and median normalized scores of our method, discrete SAC and TES-SAC across all 20 Atari games at $1$M steps}
% \label{table-score-humannorm-1m}
% \begin{tabular}{c|c|c|c}
% \toprule
%         & Discrete SAC(1M)  & TES-SAC(1M)    & Ours(1M)                \\  \hline
% Mean    & 0.5\%           & 3.0\%        & \textbf{38.5\%}   \\  
% Median  & 0.4\%           & 2.1\%        & \textbf{11.1\%}   \\
% \bottomrule
% \end{tabular}
% \end{table}

% \begin{table}[!t]
% \centering
% % \scriptsize
% \caption{Mean and median normalized scores of our method and discrete SAC across all 20 Atari games at $10$M steps}
% \label{table-score-humannorm-10m}
% \begin{tabular}{c|c|c}
% \toprule
%           & Discrete SAC(10M)        & Ours(10M)            \\ \hline
% Mean      & 151.4\%                & \textbf{220.0\%}   \\
% Median    & 90.8\%                 & \textbf{114.1\%}   \\
% \bottomrule
% \end{tabular}
% \end{table}

\begin{table}[!t]
\centering
\caption{Mean and median normalized scores of discrete SAC, TES-SAC, Rainbow, Soft-DQN and our method across all 20 Atari games at $1$M and $10$M steps.}
\label{table-score-humannorm-1m}
\resizebox{\linewidth}{!}{
\begin{tabular}{c|cccc|cccc}
\toprule
        & Discrete SAC(1M)  & TES-SAC(1M) & Soft-DQN(1M)   & Ours(1M)   & Rainbow(10M) &Discrete SAC(10M)       & Soft-DQN(10M) & Ours(10M)             \\  \hline
Mean    & 0.5\%           & 3.0\%        &  
\textbf{41.7}\% & 38.5\%   & 187.4 \% & 151.4\% & 199.2\% & \textbf{220.0\%} \\  
Median  & 0.4\%           & 2.1\%   & \textbf{20.0\%}     & 11.1\%   & 79.2 \% &90.8\% & 107.7\%  & \textbf{114.1\%}  \\
\bottomrule
\end{tabular}
}
\end{table}
\begin{table}[!t]
\centering
% \scriptsize
\caption{Raw scores across all 20 Atari games. For methods discrete SAC (1M) and TES-SAC(1M), the scores come from the corresponding paper, and the NE means the score does not exist in the original paper.}
\resizebox{\linewidth}{!}{
\begin{tabular}{ c | c  c  c c |c  c c c}
\hline
 Game &  Discrete SAC ($1$M) & TES-SAC($1$M)& Soft-DQN($1$M) & Ours($1$M) &Rainbow($10$M)&Discrete SAC ($10$M) &Soft-DQN($10$M) &  Ours ($10$M) \\ 
\hline
Alien           &   216.90    &     685.93      &  726.33 & \textbf{981.67}     &   1798.33  &  \textbf{2717.67} & 2018.00  &2158.33 \\
Amidar          &   7.9       &     42.07       &  130.03 &\textbf{132.97}     &   394.23  &  354.77  & \textbf{438.80}& 407.20      \\ 
Assault         &   350.0     &     337.03      & 881.97  & \textbf{1664.77}    &   1802.53  &            7189.97          &   \textbf{7258.87}     &   6785.60         \\
Asterix         &   272.0     &     378.5       & 676.67 & \textbf{733.33}     &   5853.33  &            2860.00            &   3761.67  &    \textbf{5993.33}         \\
BattleZone      &   4386.7    &     5790        &  \textbf{7933.33} & 6266.67    &   24266.67 &            16850.00           &    \textbf{24733.33}    & 9466.67         \\
BeamRider       &   432.1     &     NE          &  3321.60 &\textbf{3468.60}     &   3310.40  &            7169.60            &   7048.20  &    \textbf{10506.60}        \\
Breakout        &   0.7       &     2.65        &  \textbf{46.17} & 11.47 &    \textbf{492.93}  &             29.03             &   155.83   &    60.43          \\
CrazyClimber    &   3668.7    &     4.0         &  \textbf{25390.00} & 20753.33   &  30286.67 &           126320.00          &   95156.67    &   \textbf{140726.67}        \\
Enduro          &   0.8       &     NE          &  \textbf{54.23} & 0.93       &   1517.70  &            1326.77            &   1144.07  &    \textbf{2246.40}         \\
Freeway         &   4.4       &     13.57       &  17.70 &\textbf{20.17}       &   20.13   &             15.73             &    \textbf{32.30}   &   20.17          \\
Frostbite       &   59.4      &     81.03       & 294.33 &  \textbf{347.00}     &   4163.67  &   646.33                   &     2959.00    &  \textbf{4806.00}         \\
Jamesbond       &   68.3      &     31.33       &  273.33 & \textbf{368.33}     &    656.67  &            1386.67            &   965.00    &   \textbf{2085.00}         \\
Kangaroo        &   29.3      &     \textbf{307.33} & 160.00 & 120.00           &    3716.67  &            2426.67            &   2703.33  &    \textbf{5556.67}         \\
MsPacman        &   690.9     &     1408        &   1528.00 & \textbf{1639.00}     &   2738.67  &            \textbf{3221.33}       & 2386.33     &         3175.67         \\
Pong            &   -20.98    &     -20.84      &   \textbf{20.00} & \textbf{15.53}       &    \textbf{20.93}   &             20.37             &   20.73  &      20.37          \\
Qbert           &   280.5     &     74.93       &  \textbf{1400.83}  & 986.67     &    15299.17 &            12946.67           &   14293.33   &    \textbf{15325.83}        \\
RoadRunner      &   305.3     &     NE          & 5510.00 &  \textbf{12793.33}   &  \textbf{45173.33} &34043.33    & 33370.00         &   43203.33\\
SpaceInvaders   &   160.8     &     NE          & \textbf{488.83} & 383.50     &  \textbf{1330.50}  &             458.83            &   816.00  &     586.50         \\
Seaquest        &   211.6     &     116.73      & 681.33  & \textbf{744.00}     &    2105.33  &            1853.33            &    \textbf{3438.67}  &    2764.00         \\
UpNDown         &   250.7     &     207.6       & \textbf{8727.33} &  8114.67     &  9110.00  &            17803.33           &   \textbf{79313.00}  &   63441.33        \\
\hline
\end{tabular}
}
\label{table-raw_score}
\end{table}
\begin{figure} [!t]
\centering
	\subfigure{
		\includegraphics[width=0.31\textwidth]{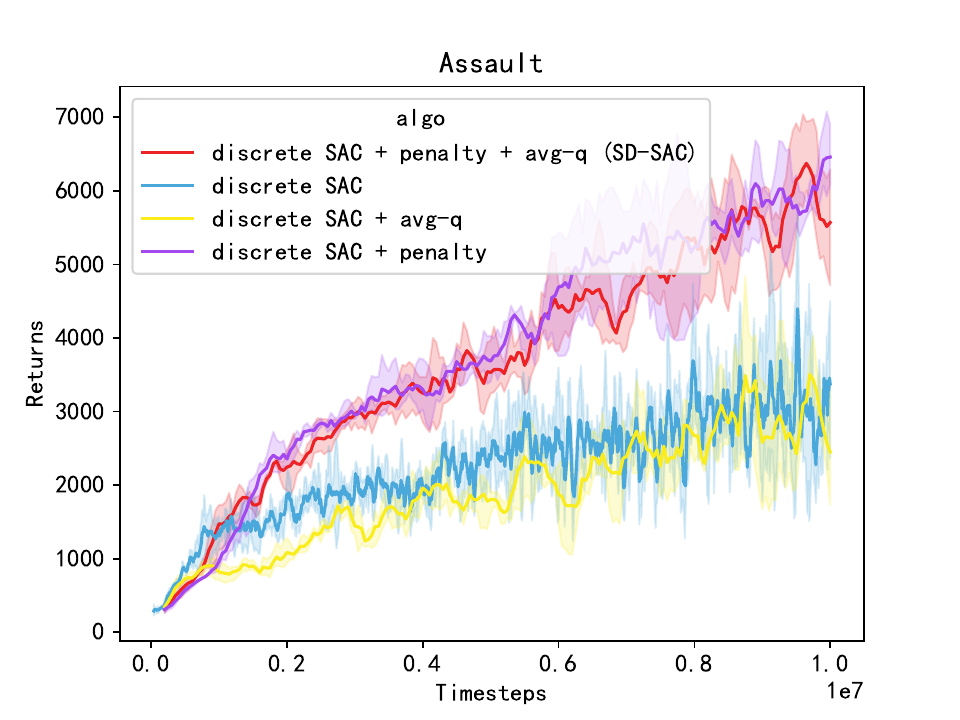}
		}
	\subfigure{
		\includegraphics[width=0.31\textwidth]{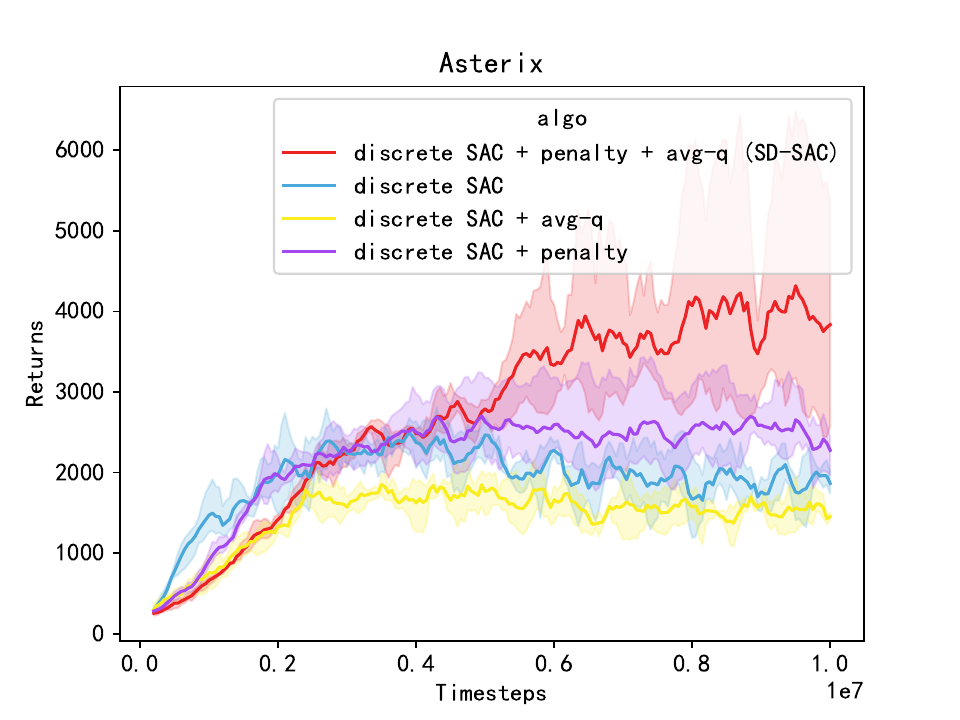}
		}
	\subfigure{
	    \includegraphics[width=0.31\textwidth]{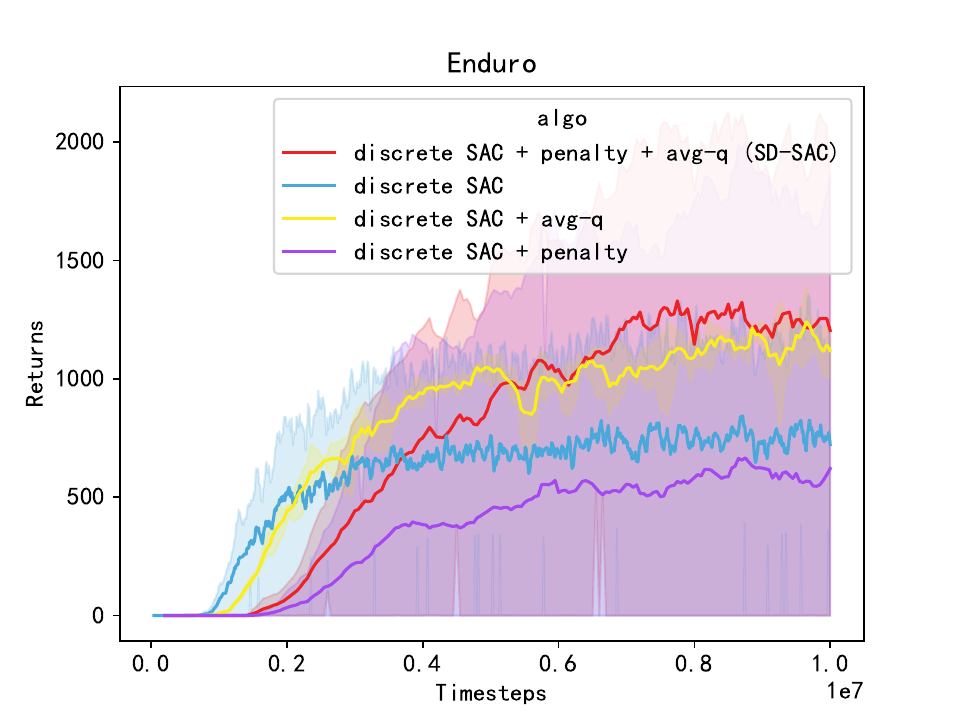}
		}
	\subfigure{
		\includegraphics[width=0.31\textwidth]{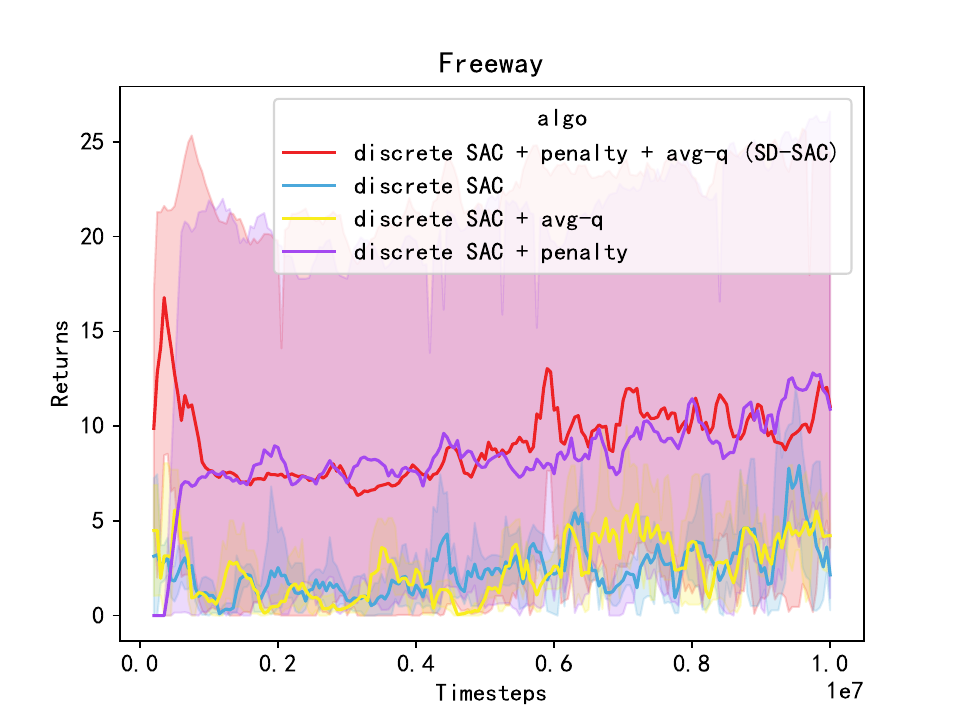}
		}
	\subfigure{
		\includegraphics[width=0.31\textwidth]{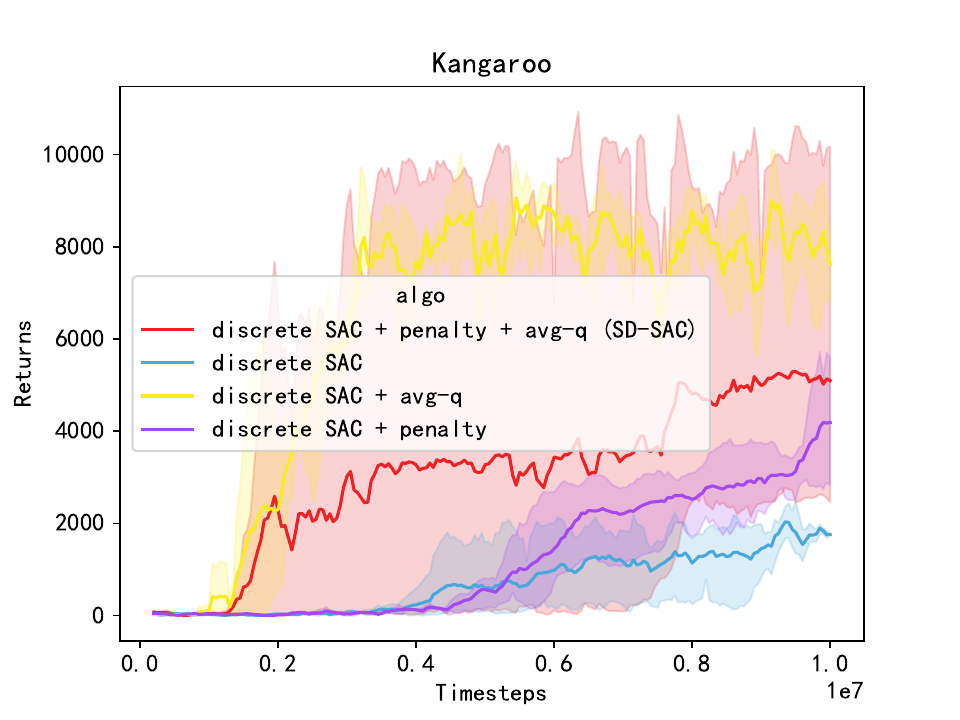}
		}
	\subfigure{
		\includegraphics[width=0.31\textwidth]{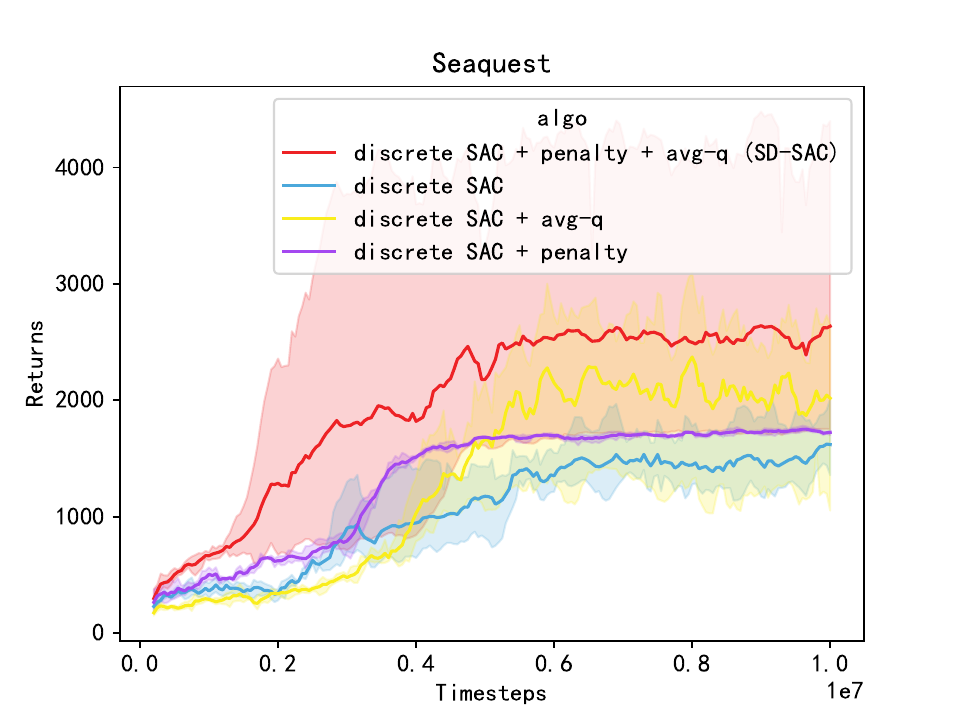}
		}
	\caption{Scores of variant discrete SAC, which includes discrete SAC, discrete SAC with entropy-penalty, discrete SAC with double average Q learning with Q-clip,for Atari games Assault, Asterix, Enduro, Freeway, Kangaroo and Seaquest.}
	\label{fig_overall_performance} 
\end{figure}

% We use precisely the same benchmarks in \citep{christodoulou2019sacd,xu2021targetannealing}. 
Table \ref{table-score-humannorm-1m} provides an overview of results and detailed results are presented in the Table \ref{table-raw_score} and Appendix \ref{appendix_detailed_experiment}. 
Since TES-SAC is not open-sourced and our re-implement algorithm following the paper underperforms the reported results, we adopt the normalized scores of discrete SAC and TES-SAC reported in the corresponding publication \citep{xu2021targetannealing}. 
When comparing our method to the discrete SAC and TES-SAC, mean normalized scores increase by 38\%  and 35.5\%, respectively. 
And our method improves the median normalized scores by 10.7\% and 9.0\% while compared with discrete SAC and TES-SAC. 
%The significant improvement is attributed to our way of Q estimation on discrete action spaces.

To verify the effect of a longer training process, table \ref{table-score-humannorm-1m} also compares discrete SAC, Rainbow, Soft-DQN, and our method performance on 10 million steps. 
Compared with discrete SAC, our method has improved the normalized scores by 68.6\% and 23.3\% on mean and median, respectively. Additionally, our proposed method outperformed Rainbow by 32.6\% on the mean and by 34.9\% on the median. Better Q-estimation and steady policy updates are responsible for the performance increase in average scores. The experimental results demonstrate that benefiting from the deterministic greedy policy and entropy regularization in the evaluation step, Soft-DQN's performance improves rapidly in the early stages and achieves the best results at 1 million steps. However, due to the early convergence of the deterministic greedy policy, Soft-DQN's performance stagnates after 4 million steps, as seen in Fig.~\ref{fig_total_curves_20_atari}. Our method outperforms Soft-DQN in the final 10 million steps by 20.8\% on average and 6.4\% on median, due to the training stability brought by entropy penalty and the optimistic exploration altered by the double avg-Q with Q-clip.

\subsection{Ablation Study}

Fig.~\ref{fig_overall_performance} shows the learning curves for 6 environments. Entropy-penalty (purple curve) increases performance compared to the discrete SAC in each of the six environments and even increases 2x scores in Assault. This shows that discrete SAC can perform excellently after removing unstable training. 
Except for Asterix, the alternative choice of clipped double Q-learning, which is double average Q learning with Q-clip (yellow curve), also shows some improvement compared to the discrete SAC in 5 environments. 
Additional improvements can be derived when the combination of both alternative design choices is used simultaneously. 

To evaluate the influence of hyperparameter tuning, we also conducted a comprehensive hyperparameter analysis. By experimenting with different $\alpha$ and learning rate in the discrete SAC algorithm, we identify the performance upper bound of discrete SAC. The results show that, under various $\beta$ values, SD-SAC consistently outperforms this upper bound, demonstrating that entropy penalty serves as a better and more balanced constraint. This further confirms that SD-SAC can significantly achieve a more stable training process. We present the experiment details and results in Appendix \ref{appendix_hyperparameter_analysis}.

% \subsection{Hyperparameter Analysis}

% Our alternate design method incorporates two hyperparameters, i.e., entropy-penalty coefficient $\beta$ and Q-clip range $c$. Fig.~\ref{fig_coefficient} compares various entropy-penalty coefficient $\beta$ and Q-clip range $c$ values. 
% The constraint proportion of policy change is determined by the entropy-penalty coefficient $\beta$. Intuitively, an excessive penalty term will lead to policy under-optimization. 
% We experiment with different $\beta$ in \{0.1, 0.2, 0.5, 1\}. 
% We find that $\beta=0.5$ can effectively limit entropy randomness while improving performance.
% The Q-clip constrains different ranges of Q value range $c$, and experiments with different ranges $c$ in \{0.5, 1, 2, 5\} show that 0.5 is a reasonable constraint value. 
% \begin{figure}[!t]

% \includegraphics[width=0.23\textwidth]{images/experiments/comparative_coefficient/Seaquest_penalty.jpeg}
% \includegraphics[width=0.23\textwidth]{images/experiments/comparative_coefficient/Seaquest_q_clip.jpeg}

% \caption{Average return on Seaquest: a) variants entropy-penalty coefficient $\beta$ with 0.1, 0.2, 0.5 and 1. b) variants q-clip epsilon $c$ with 0.5, 1, 2 and 5.
% }
% \label{fig_coefficient}
% \end{figure}

\subsection{Qualitative Analysis}

% \begin{figure} [!t]
% 	\subfloat[baseline discrete sac]{
% 		\includegraphics[width=0.23\textwidth]{images/experiments/loss_surfaces_white/base_loss_surfaces_final.png}
% 		}
% 	\subfloat[our method]{
% 		\includegraphics[width=0.23\textwidth]{images/experiments/loss_surfaces_white/total_loss_surface_final.png}
% 		}

% 	\caption{The loss surfaces of discrete sac and our method on Atari game Seaquest with trained weights after 10m steps}
% 	\label{fig_loss_surface} 
% \end{figure}

\begin{figure} [ht]
    \centering
     \subfigure[Loss Surfaces]{
        \begin{minipage}[b]{0.35\textwidth}
        \includegraphics[width=1\textwidth]{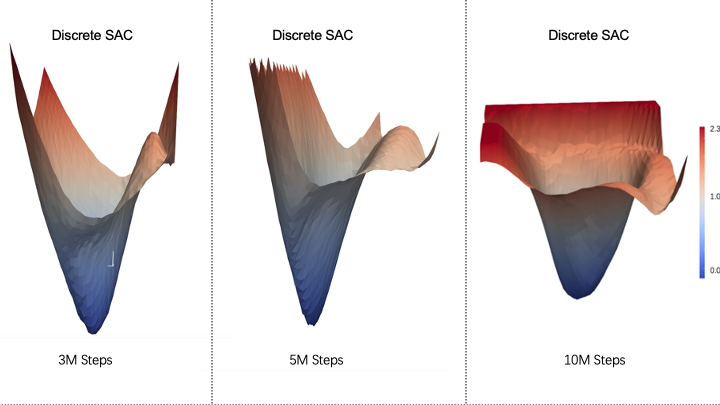}\\
	\includegraphics[width=1\textwidth]{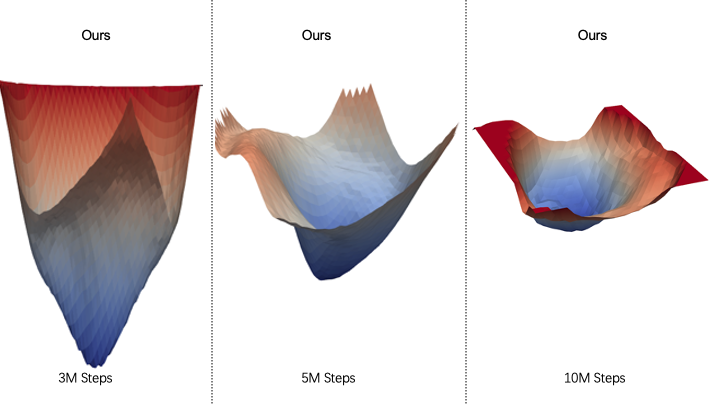}
        \end{minipage}
    }
    \subfigure[Scores]{
		\includegraphics[width=0.45\textwidth]{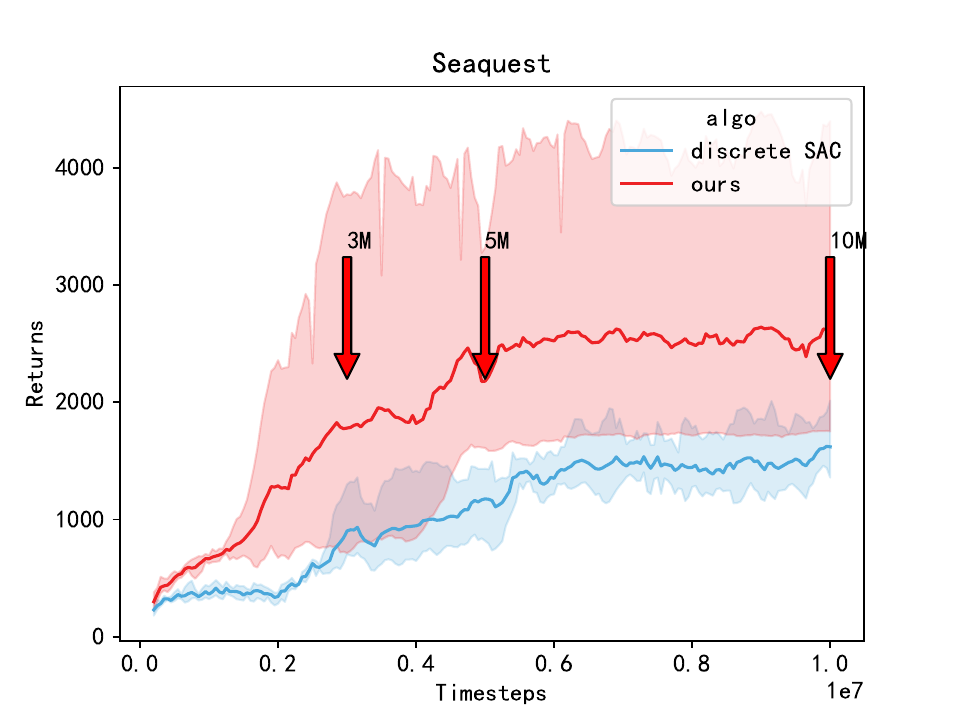}
    }
	\caption{The loss surfaces of discrete SAC and our method on Atari game Seaquest with trained weights at 3 million, 5 million and 10 million steps.}
	\label{fig_loss_surface} 
\end{figure}

Fig.~\ref{fig_loss_surface} shows loss surfaces of the discrete SAC and our method by using the visualization method proposed in \citep{li2018visualizing,ota2021training} with the loss of TD error of Q functions. According to the sharpness/flatness in these two sub-figures, our method has a nearly convex surface, while discrete SAC has a more complex loss surface. 
The surface of our method has fewer saddle points than the discrete SAC, which further shows that it can be more smoothly optimized during the training process.

% \begin{figure}[h]
% \includegraphics[width=0.23\textwidth]{images/experiments/loss_surfaces/base_loss_surface_show.png}
% \includegraphics[width=0.23\textwidth]{images/experiments/loss_surfaces/total_loss_surface_show.png}

% \caption{loss surfaces
% }
% \label{fig_loss_surface}
% \end{figure}

% \begin{figure*} [t!]
% 	\subfloat[baseline discrete sac]{
% 		\includegraphics[width=0.23\textwidth]{images/experiments/loss_surfaces/base_loss_surface_show.png}
% 		}
% 	\subfloat[our method]{
% 		\includegraphics[width=0.23\textwidth]{images/experiments/loss_surfaces/total_loss_surface_show.png}
% 		}

% 	\caption{1dfkjsdjfsdfks}
% 	\label{fig_loss_surface} 
% \end{figure*}
\begin{figure} [!ht]
    \centering
	\subfigure[Honor of Kings]{
		\includegraphics[width=0.48\textwidth,height=4.5cm]{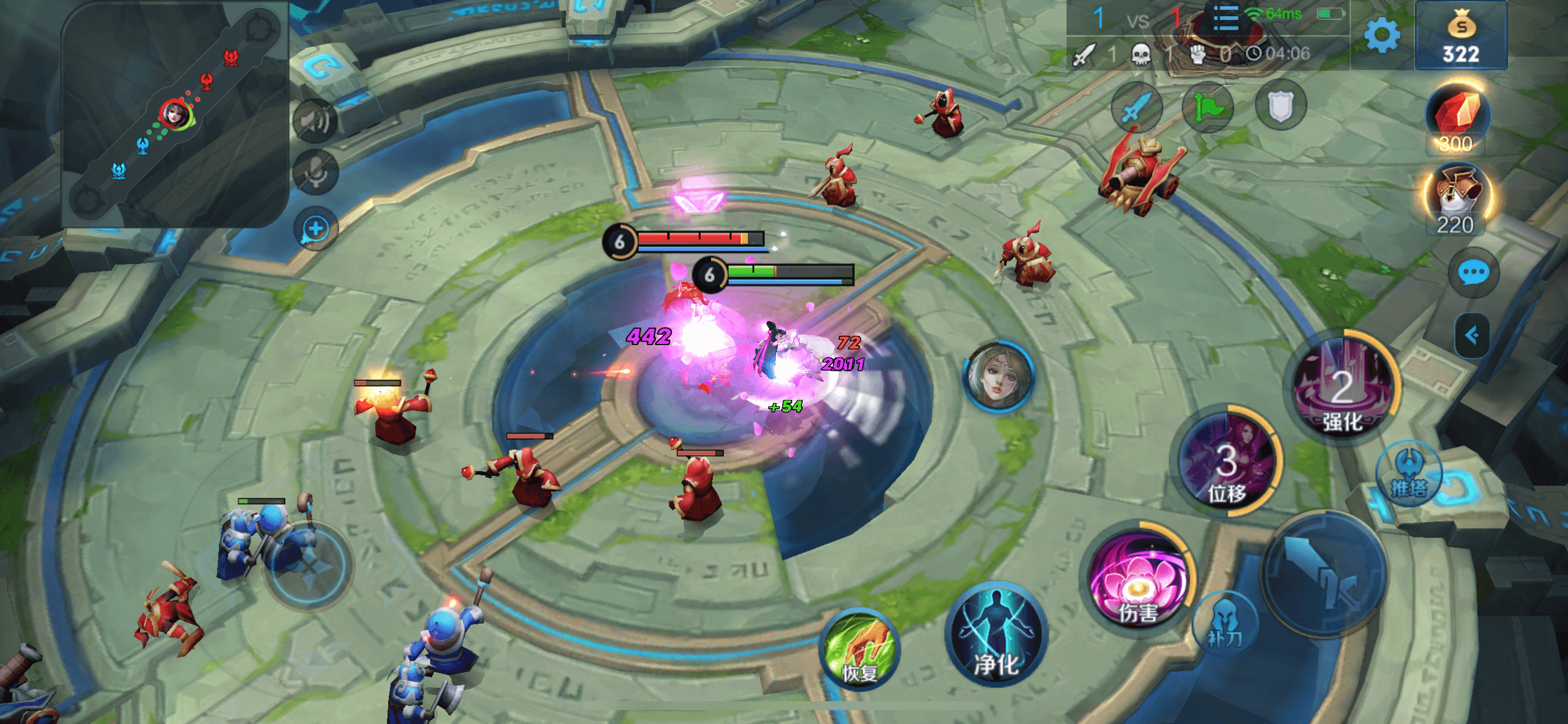}
		}
	\subfigure[ELO Scores]{
		\includegraphics[width=0.4\textwidth]{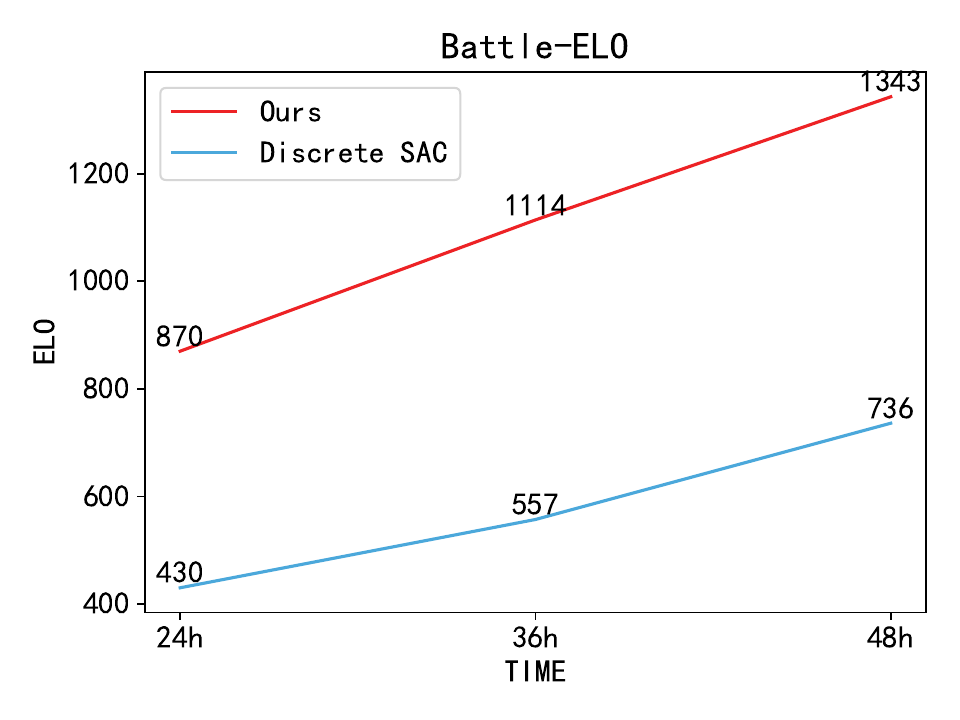}
		}

	\caption{a) A screenshot of the Honor of Kings 1v1 game. b) The ELO scores, compared with discrete SAC and our method, were tested for three snapshots of 24, 36, and 48 hours during training.}
	\label{fig_hok_result} 
\end{figure}

\subsection {Case Study using Honor of Kings}

We further deploy our method into Honor of Kings 1v1, a commercial game in the industry, to investigate the scale-up ability of our proposed SD-SAC algorithm.

%\subsubsection{Overview of Honor of Kings 1v1}
%https://en.wikipedia.org/wiki/Honor_of_Kings
Honor of Kings \footnote{\url{https://github.com/tencent-ailab/hok_env}} is a popular MOBA (Multiplayer Online Battle Arena) game and a good testbed for RL research \citep{ye2020supervised,ye2020mastering,ye2020towards,chen2021heroes,wei2022hokenv}.
The game descriptions are in \citep{ye2020mastering, ye2020towards}. 
%the most popular mobile MOBA game in China. A comprehensive overview of the game can be found on wikipedia\footnote{https://en.wikipedia.org/wiki/Honor\_of\_Kings}. 
In our experiments, we use the one-versus-one mode (1v1 solo), with both sides being the same hero: Diao Chan. 
The state of the game is represented by feature vectors, as reported in \citep{ye2020mastering, wei2022hokenv}. 
%The four parts of these state features are the following: public hero features, which describe the hero's status; private hero features, which describe the specific kill information for all the heroes in the game; soldier features describe the status of soldiers in the troops and the final organ feature which describes the status of turrets and crystals. 
The action space is discrete, i.e., we discretize the direction of movement and skill, same to \citep{ye2020mastering, ye2020towards}. 
%The action of Honor of Kings 1v1 contains 6 heads of discrete outputs, which represent action button, the movement offset over the x-axis and y-axis, the skill offset over the x-axis and y-axis, and the target game unit. Additionally, the reward design includes attacks on enemy turrets, killing enemy soldiers, and dealing with damage to enemy heroes. 
The goal of the game is to destroy the opponent's turrets and base crystals while protecting its own. % The ELO rating system, calculated from the win rate, is used to measure the ability of two agents. 
We use the ELO rating system~\citep{elo1978rating}, which calculate scores from the win rate, to measure the ability of two agents. A detailed introduction of the ELO system is presented in Appendix \ref{appendix_elo}.

We selected three snapshots of 24, 36, and 48 hours during the training process, resulting in 6 agents (SD-SAC-24h, SD-SAC-36h, SD-SAC-48h, DSAC-24h, DSAC-36h, DSAC-48h). We conducted 48 one-on-one matches for each agent, resulting in a total of 720 matches and thus serving as the basis of ELO calculation. 

The results are shown in Fig.~\ref{fig_hok_result}. Throughout the entire training period, our method outperforms discrete SAC by a significant margin, which indicates our method's efficiency in large-scale cases.  Specifically, SD-SAC-48h achieved 35 wins, 7 draws, and 6 losses, with a win rate of 72.92\%. The agent also exhibits higher skill hit rate, higher Kill/Death ratio and better turret-dashing ability.  

\section{Conclusions and Future Work} \label{section_discussion}
Many algorithmic design choices in reinforcement learning are limited to the regime of the chosen benchmark tasks. 
%Many algorithmic designs in reinforcement learning are constrained by the scope of the selected benchmark tasks. % Our study highlights, for example, soft actor-critic (SAC), that widely accepted design choices in continuous action space do not necessarily generalize to new discrete environments. We conduct failure mode analyses on Atari benchmarks to understand and diagnose the implications of default design choices.
We highlight that soft actor-critic (SAC), that widely accepted design choices in continuous action space do not necessarily generalize to new discrete environments. We conduct failure mode analysis and obtain two main insights: 1) due to the deceptive reward, the unstable coupling update of policy and Q function will further disturb training; 2) the underestimation bias caused by double Q-learning results in the agent's pessimistic exploration and inefficient sample usage.
We thereby propose two alternative design choices for SAC: entropy-penalty and double-average Q-learning with Q-clip, resulting in a new algorithm, called SD-SAC. Experiments show that our alternative design choices increase the training stability and Q-value estimation accuracy, which ultimately improves overall performance. In addition, we also apply our method to the large-scale MOBA game Honor of Kings 1v1 to show the scalability of our optimizations. 

Finally, the success obscures certain flaws, one of which is that our improved discrete SAC still performs poorly in instances involving long-term decision-making. One possible reason is that SAC can not accurately estimate the future only by rewarding the current frame. In order to accomplish long-term choices with SAC, our next study will concentrate on improving the usage of the incentive signal across the whole episode.

\bibliography{body/references.bib}

\appendix
\clearpage

\section{Additional Details and Experiment Results}

\subsection{Detailed Experiment Results on 20 Atari Game Environments}
\label{appendix_detailed_experiment}
In Figure ~\ref{fig_total_curves_20_atari}, we present the learning curves of all 20 experiments.
\begin{figure} [!ht]
        \centering
	\subfigure{
		\includegraphics[width=0.31\textwidth]{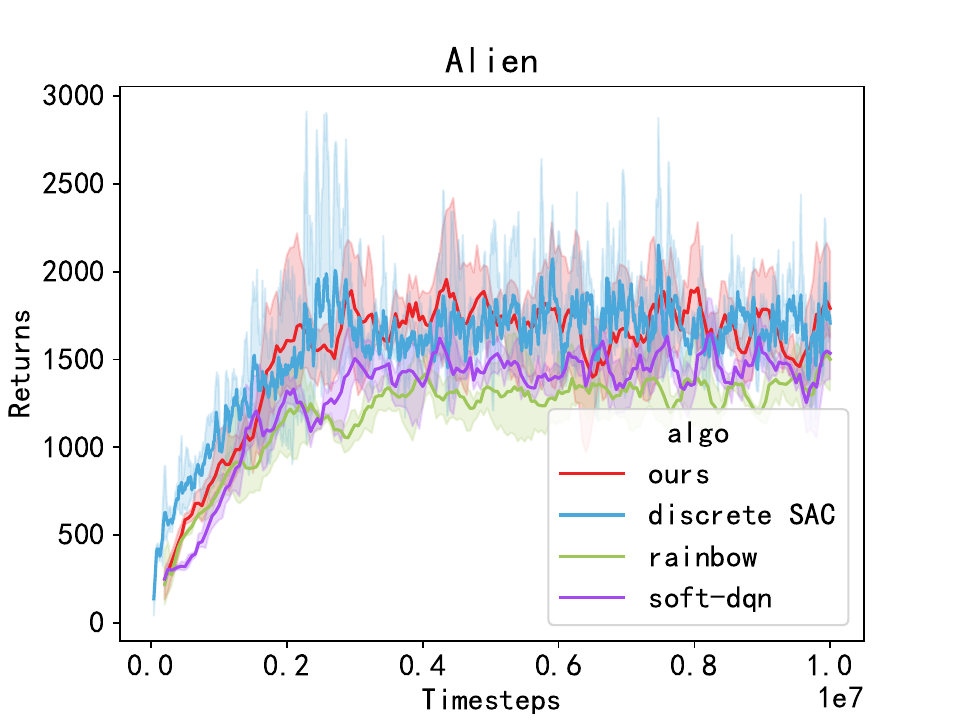}
		}
	\subfigure{
		\includegraphics[width=0.31\textwidth]{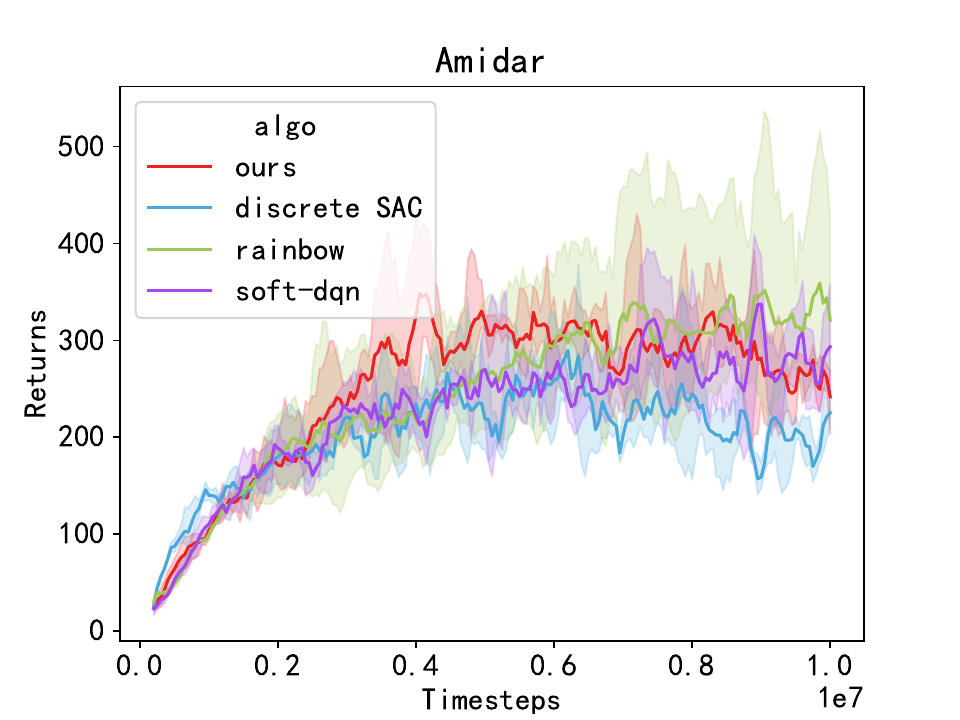}
		}
	\subfigure{
		\includegraphics[width=0.31\textwidth]{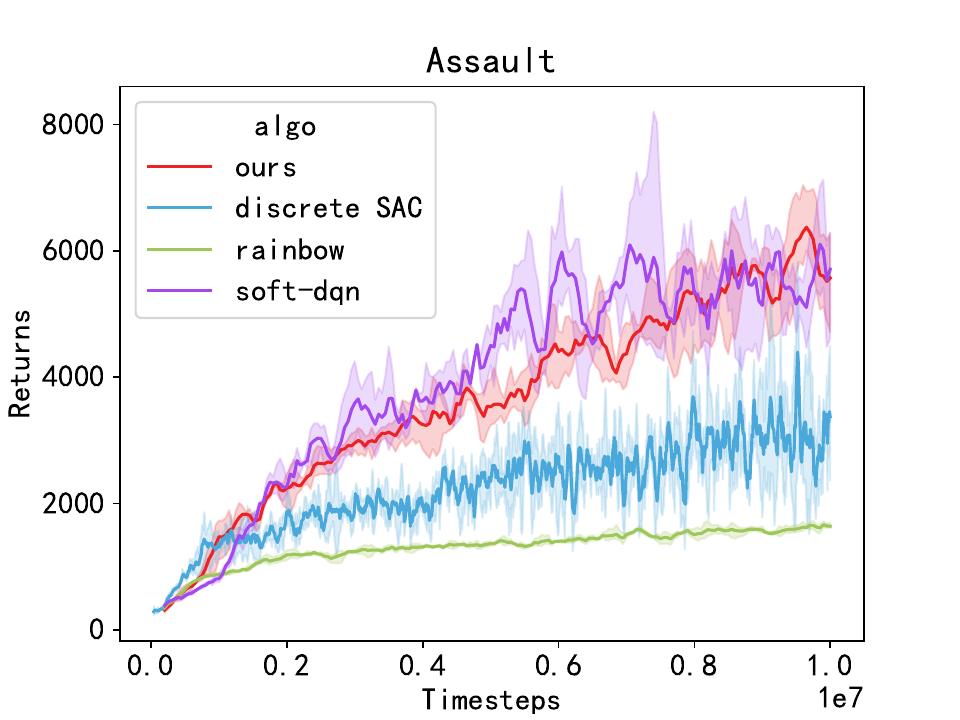}
		}
	\subfigure{
		\includegraphics[width=0.31\textwidth]{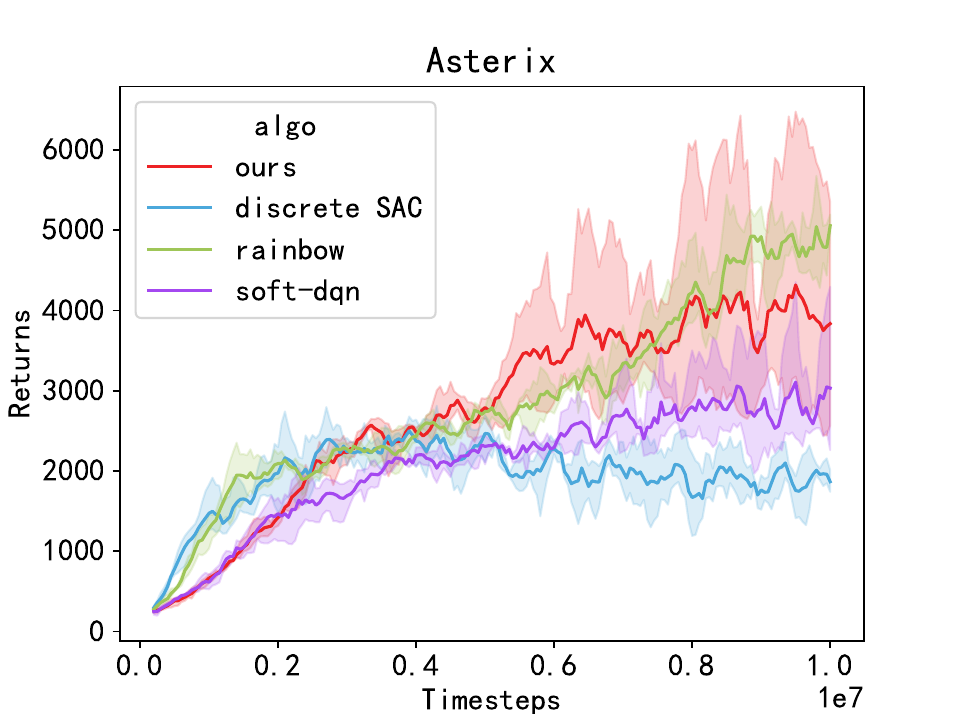}
		}
	\subfigure{
		\includegraphics[width=0.31\textwidth]{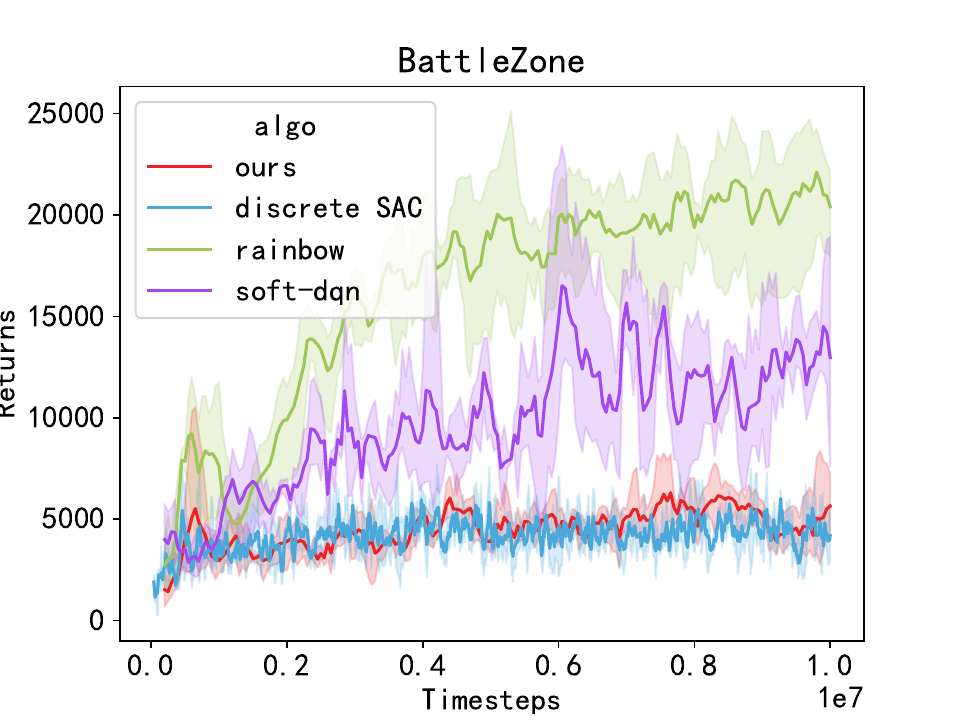}
		}
	\subfigure{
		\includegraphics[width=0.31\textwidth]{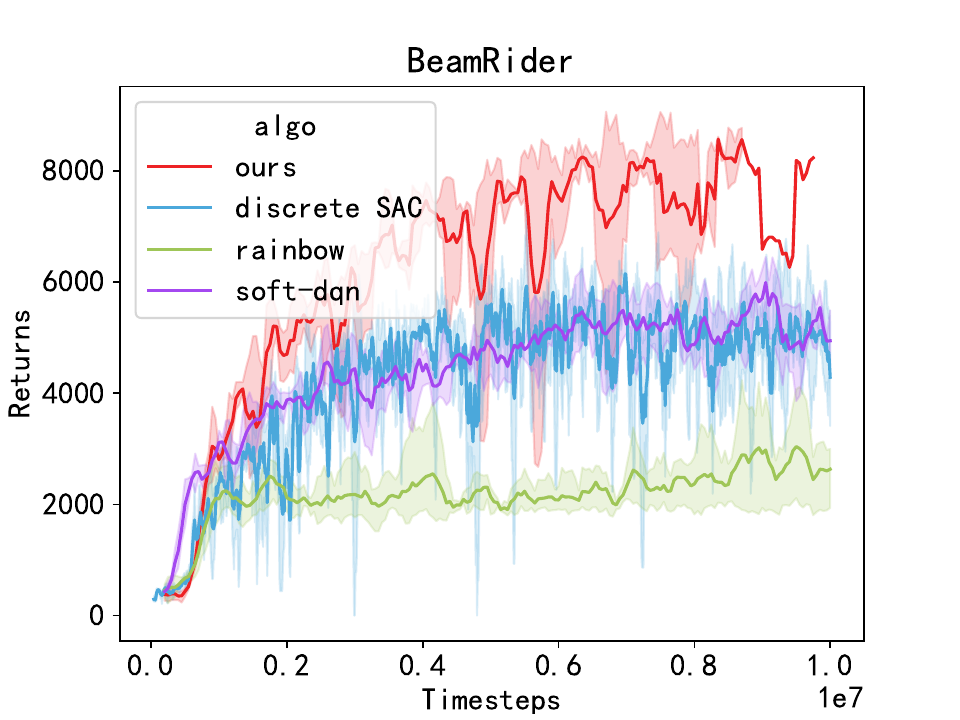}
		}
	\subfigure{
		\includegraphics[width=0.31\textwidth]{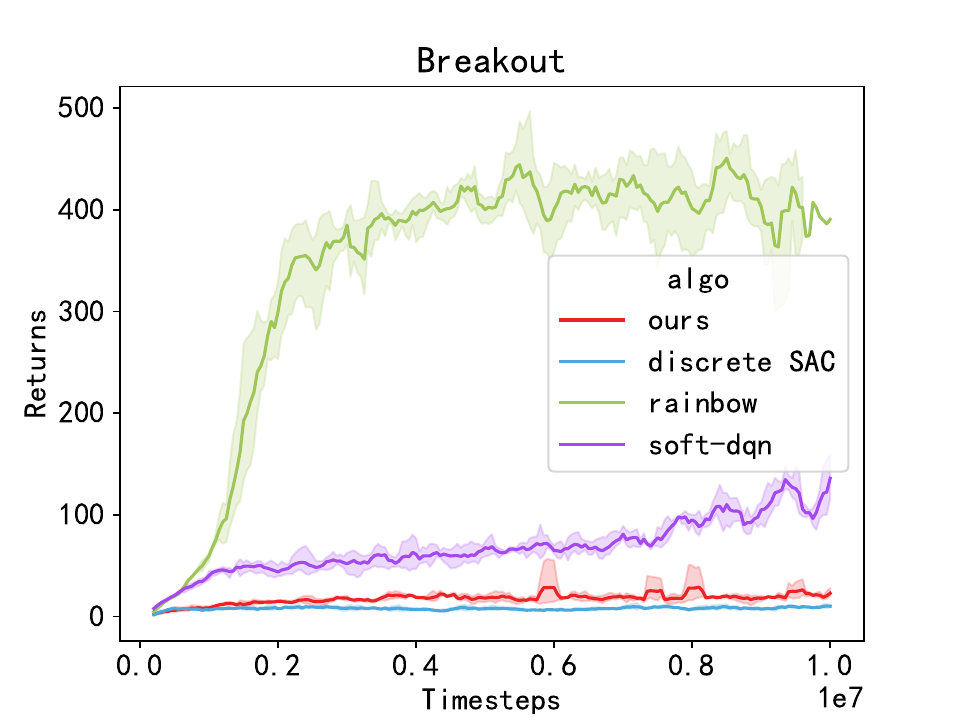}
		}
	\subfigure{
		\includegraphics[width=0.31\textwidth]{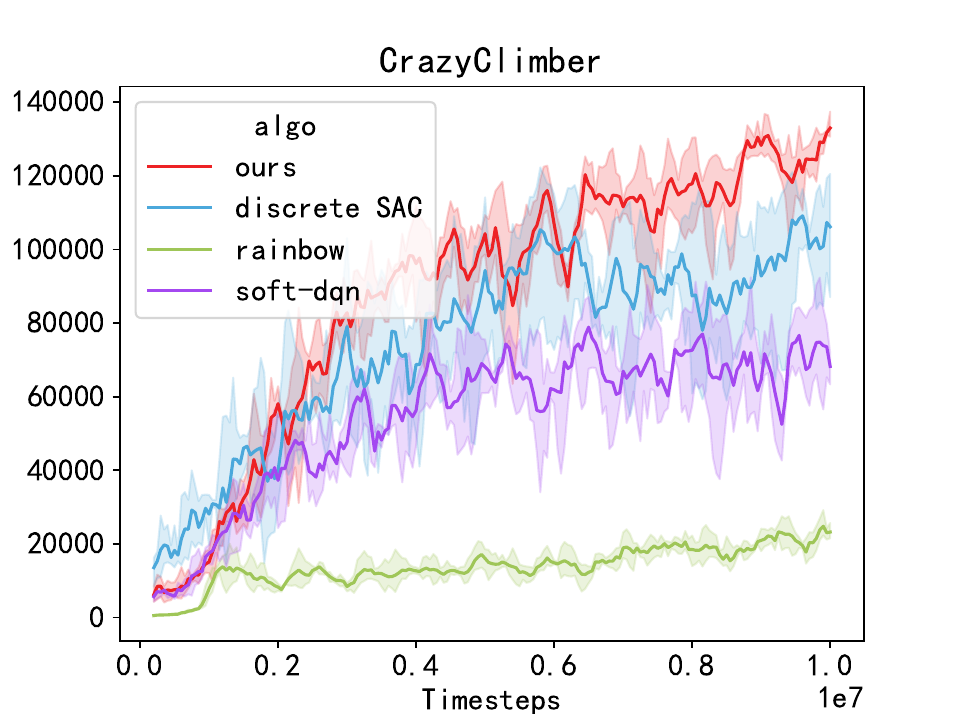}
		}
	\subfigure{
		\includegraphics[width=0.31\textwidth]{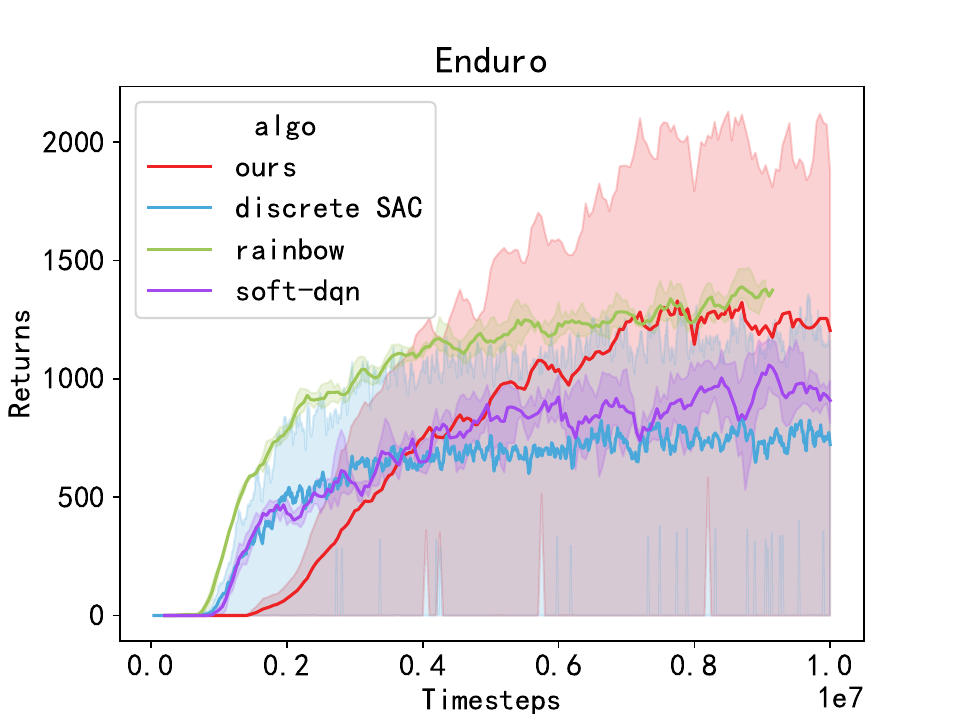}
		}
	\subfigure{
		\includegraphics[width=0.31\textwidth]{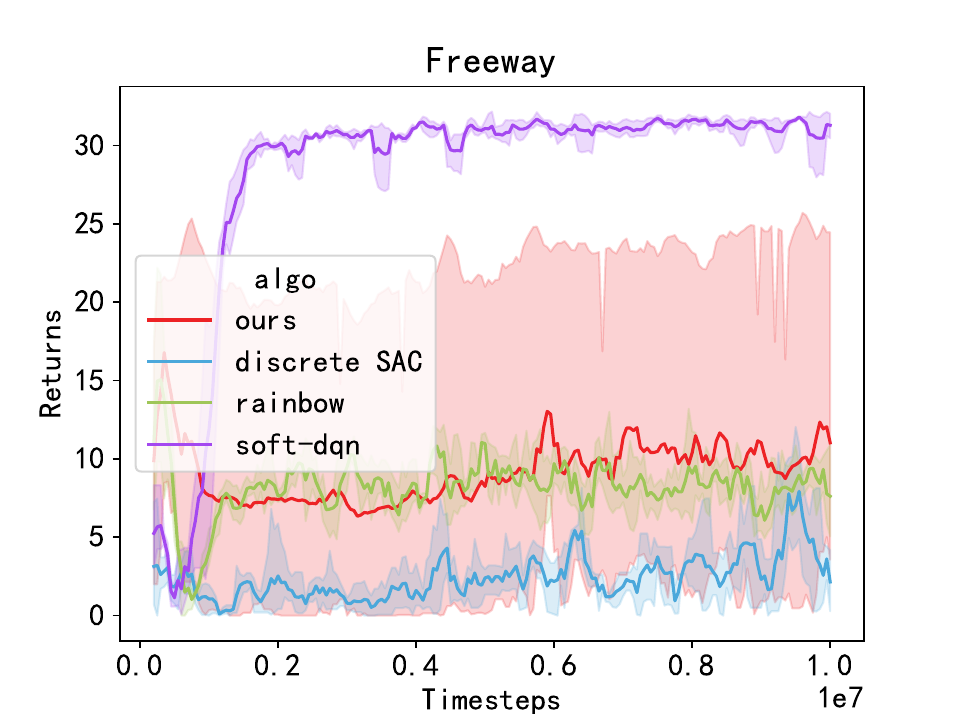}
		}
	\subfigure{
		\includegraphics[width=0.31\textwidth]{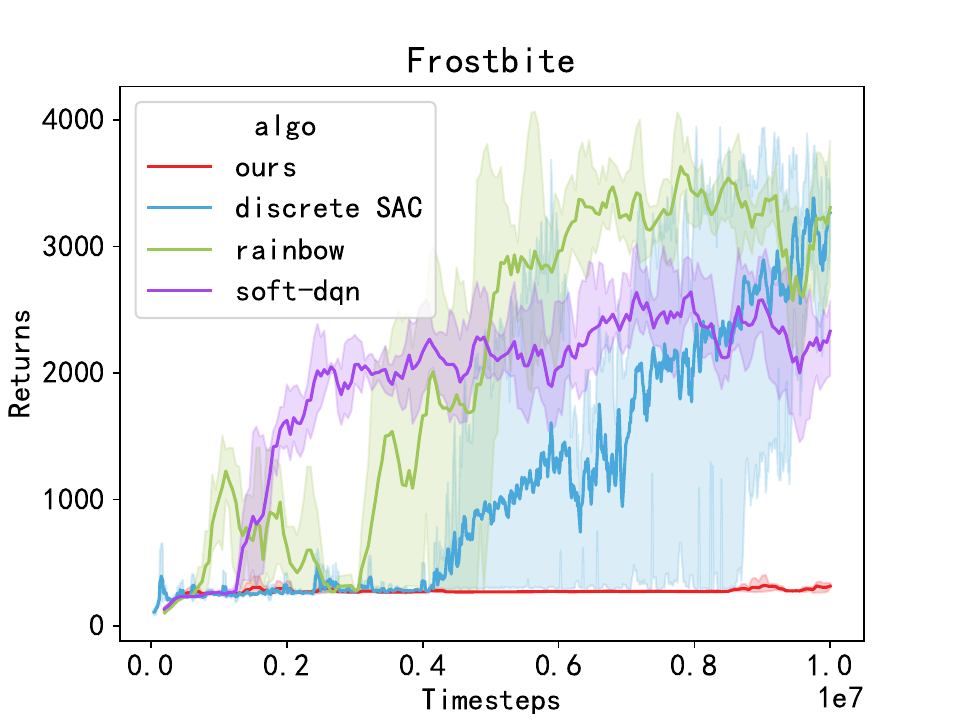}
		}
	\subfigure{
		\includegraphics[width=0.31\textwidth]{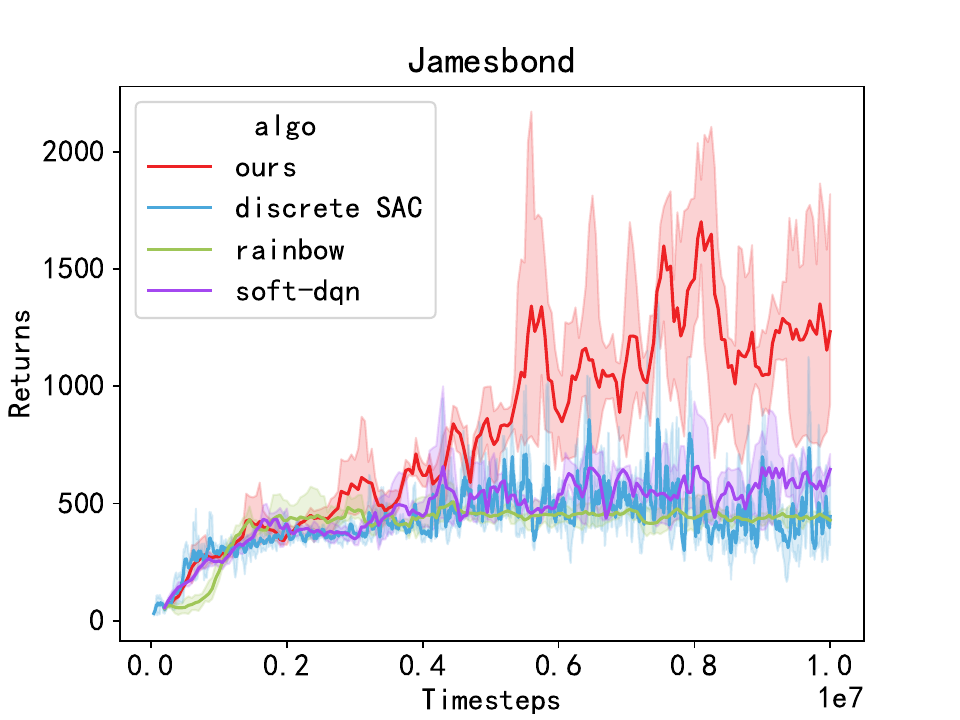}
		}
    \end{figure}
\clearpage   
    \begin{figure}[!ht]\ContinuedFloat
    \centering
	\subfigure{
		\includegraphics[width=0.31\textwidth]{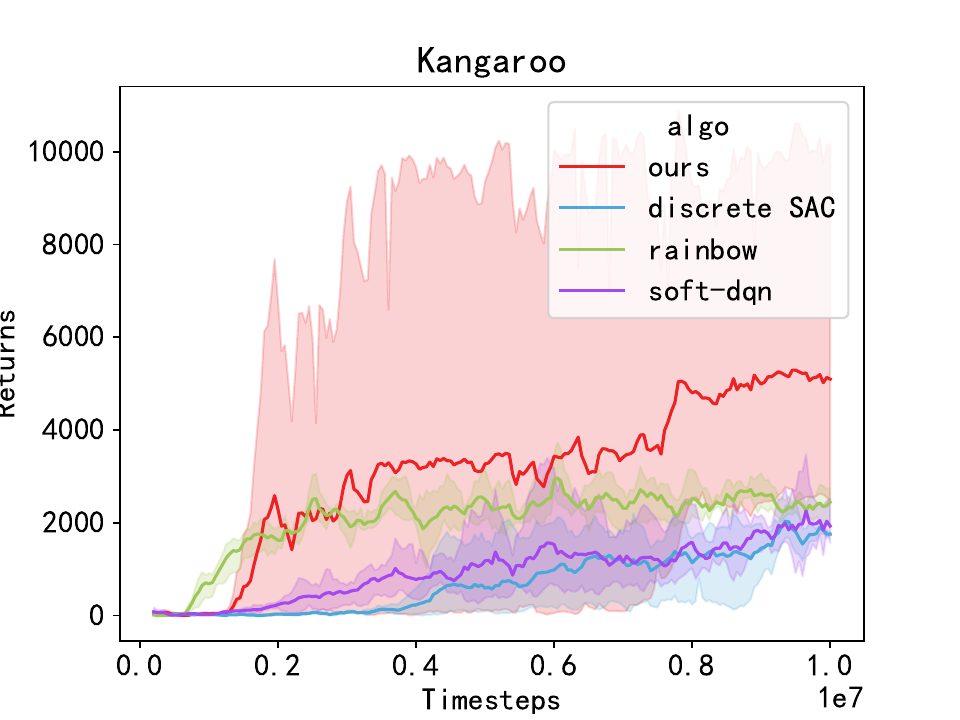}
		}
	\subfigure{
		\includegraphics[width=0.31\textwidth]{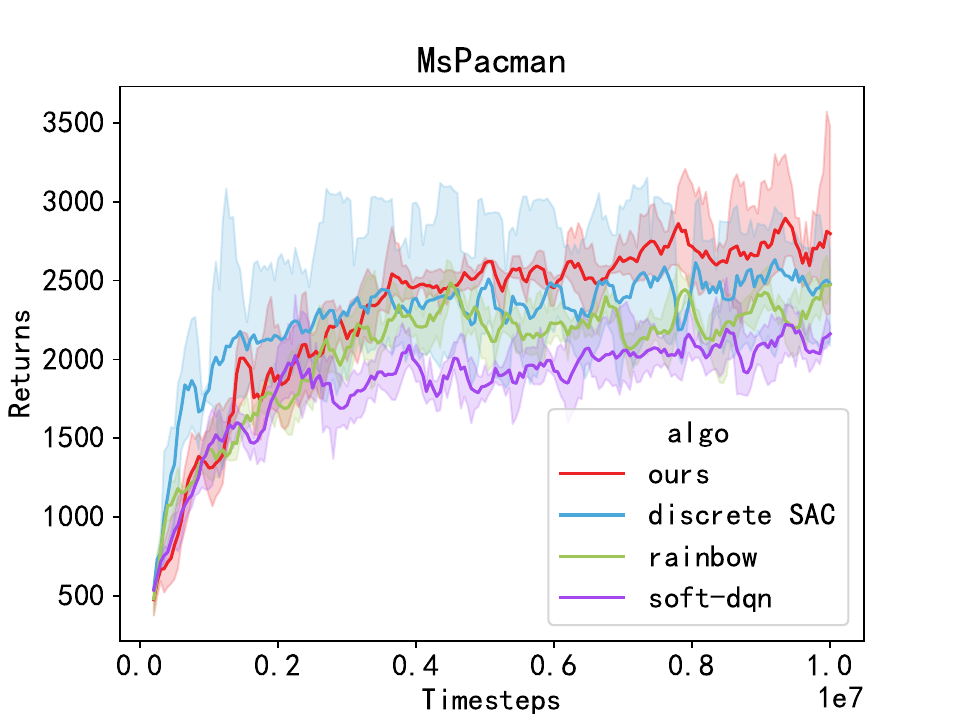}
		}
	\subfigure{
		\includegraphics[width=0.31\textwidth]{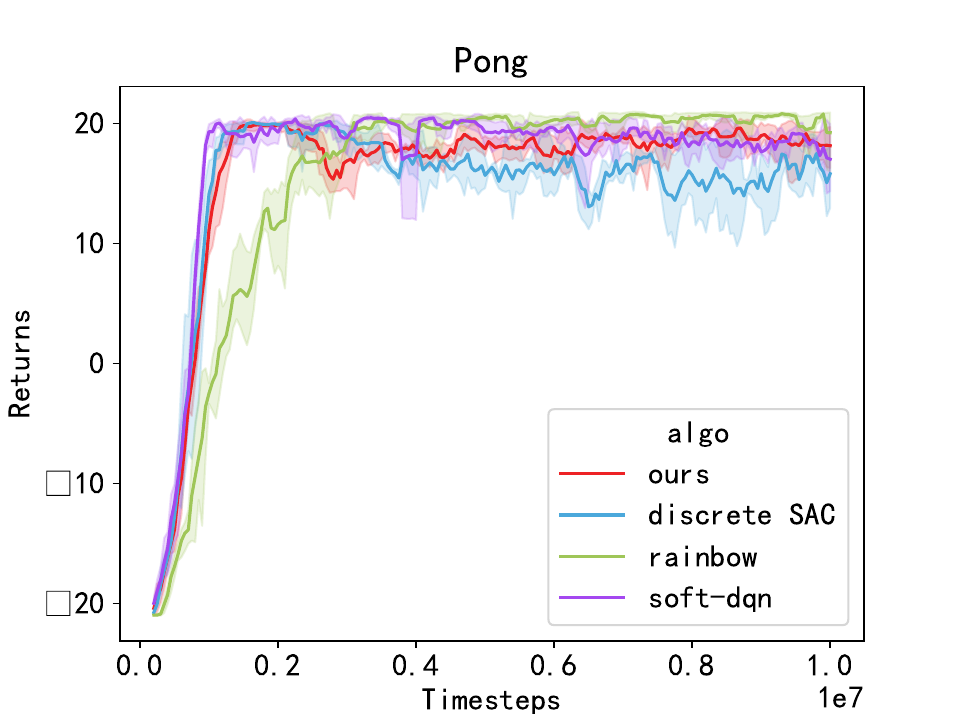}
		}
	\subfigure{
		\includegraphics[width=0.31\textwidth]{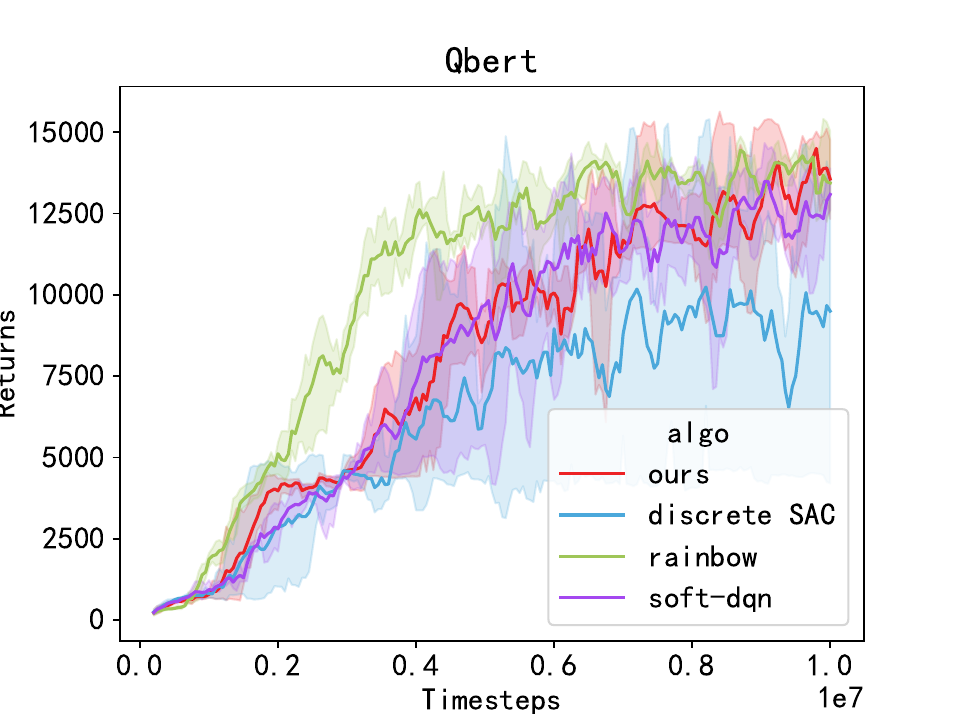}
		}
	\subfigure{
		\includegraphics[width=0.31\textwidth]{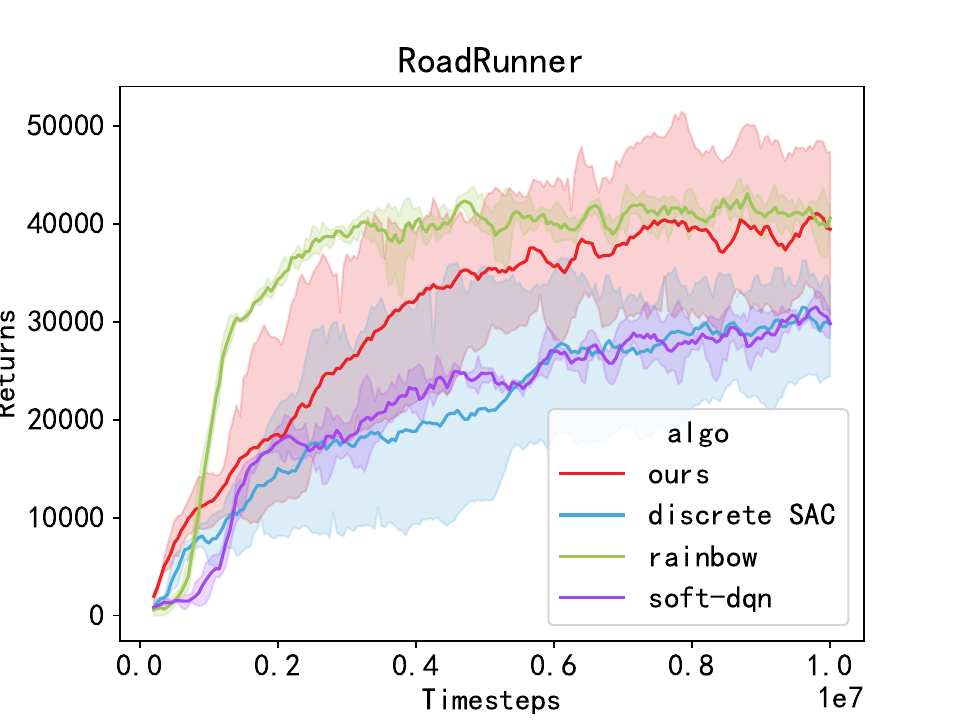}
		}
	\subfigure{
		\includegraphics[width=0.31\textwidth]{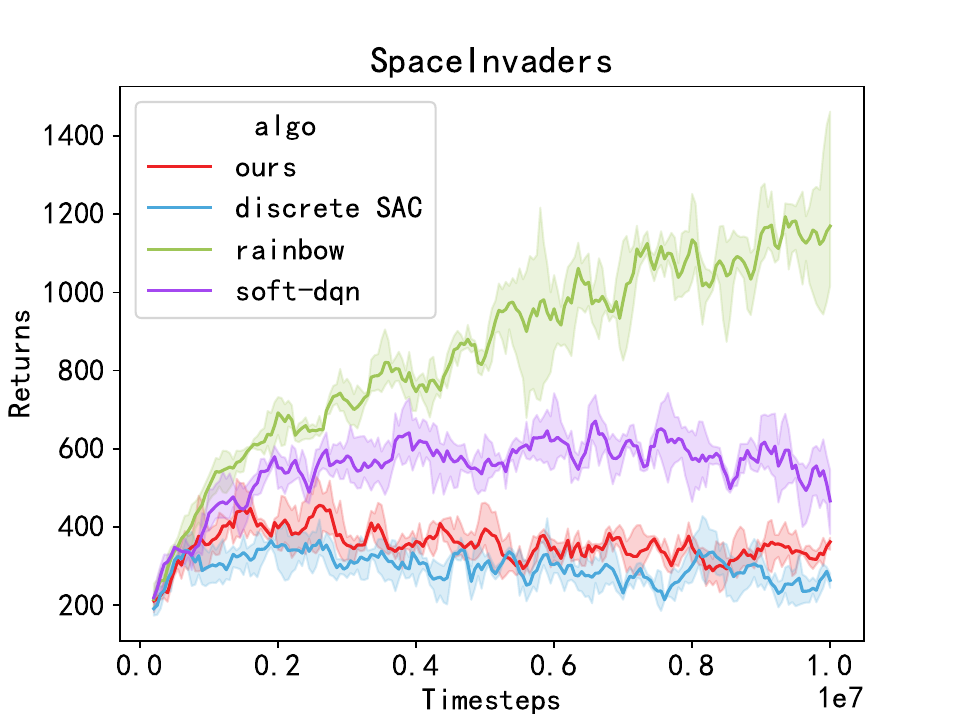}
		}
	\subfigure{
		\includegraphics[width=0.31\textwidth]{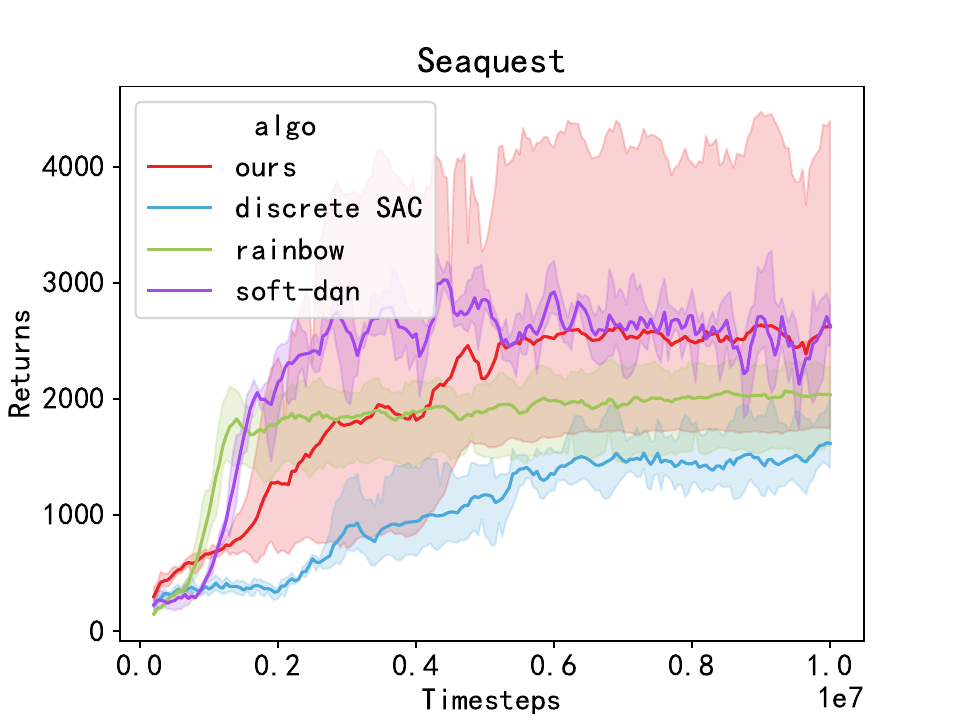}
		}
	\subfigure{
		\includegraphics[width=0.31\textwidth]{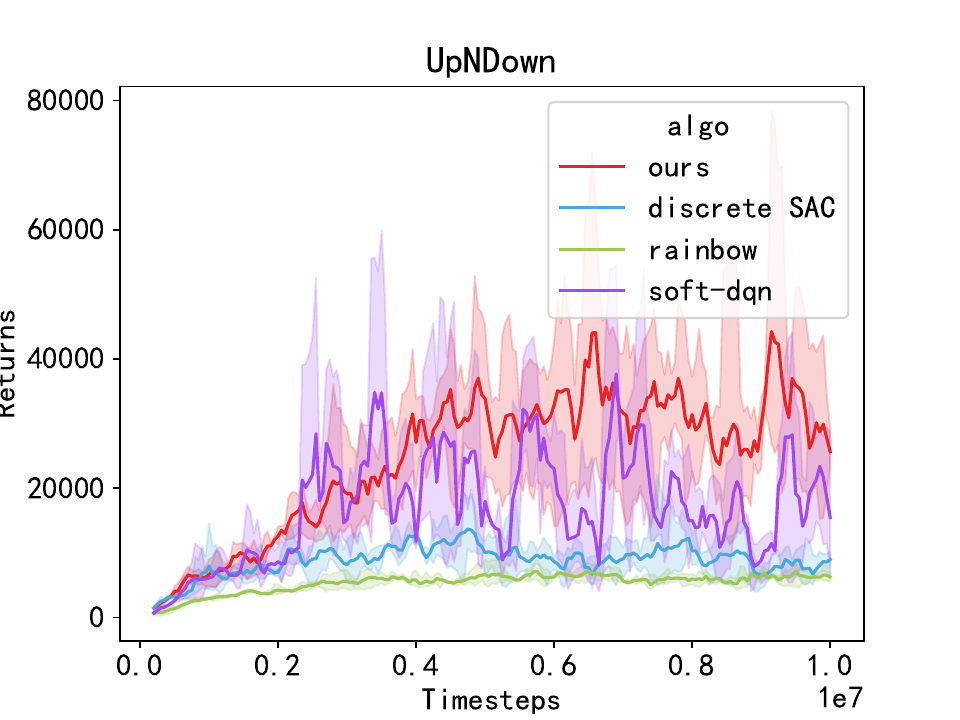}
		}
	\caption{Learning curves for discrete SAC, Rainbow, Soft-DQN, and ours, for each game. Every curve is smoothed with a moving average of 10 to improve readability.}
	\label{fig_total_curves_20_atari} 
\end{figure}

\subsection{Further Comparison of SD-SAC and DSAC}
\begin{figure} [htbp]
    \centering
	\subfigure[Episode Length]{
		\includegraphics[width=0.48\textwidth]{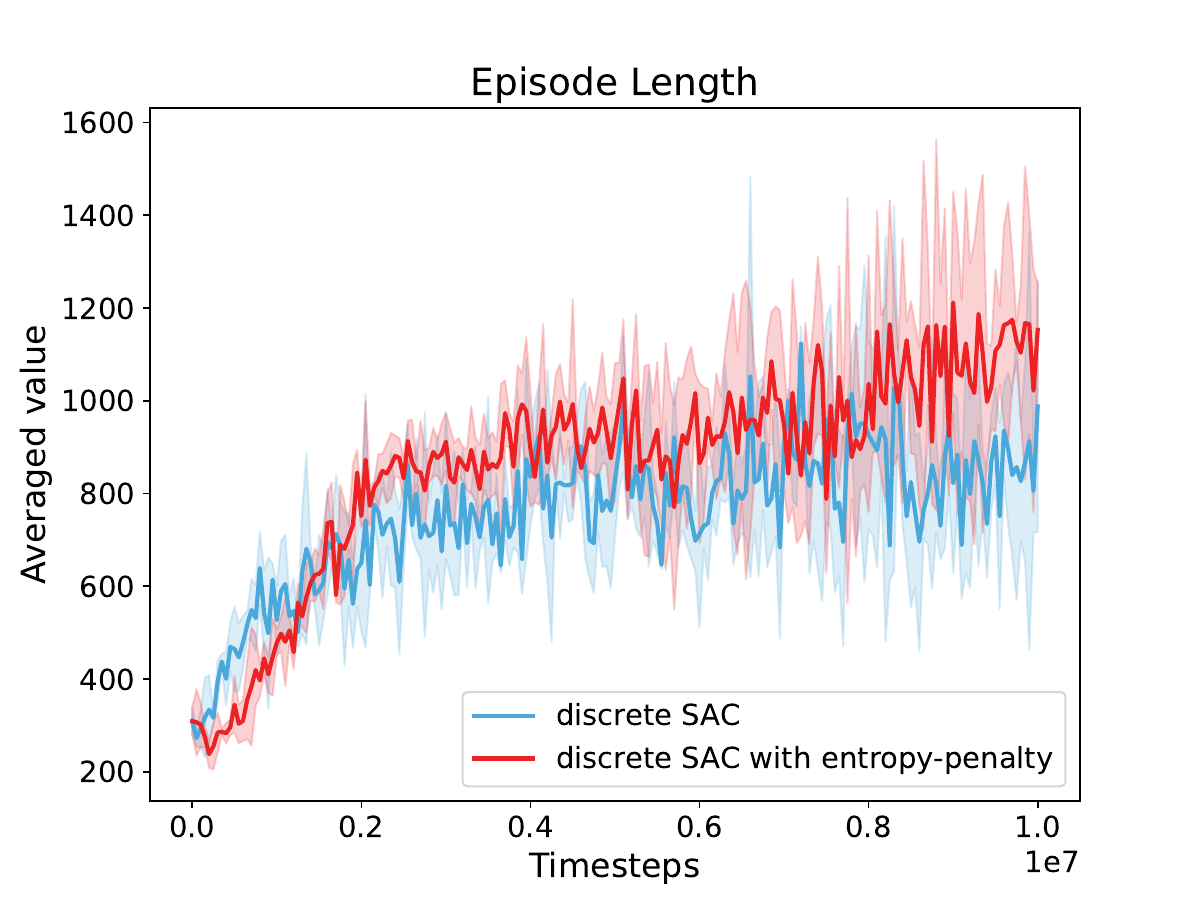}
		}
	\subfigure[Number of Steps with Rewards]{
		\includegraphics[width=0.48\textwidth]{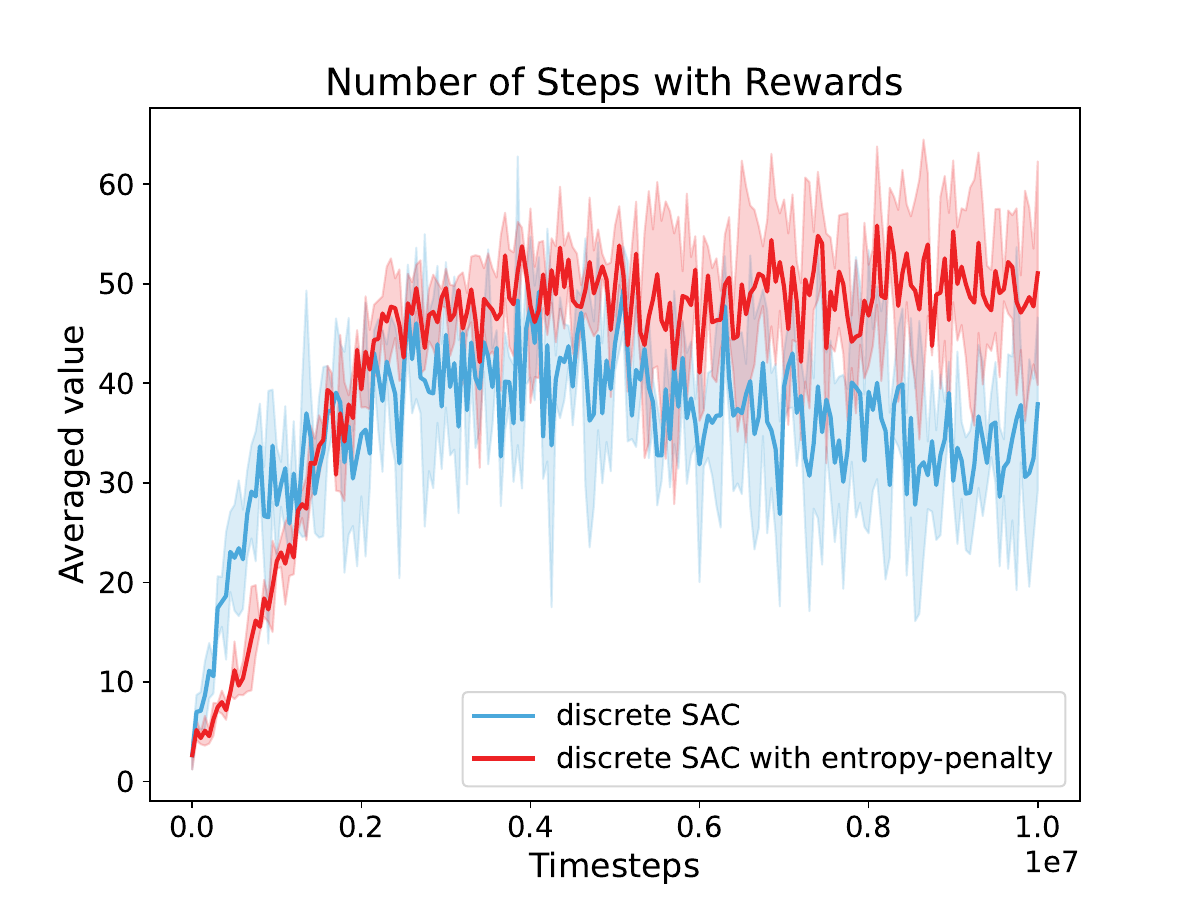}
		}
	\caption{Comparison of SD-SAC and DSAC in terms of episode length and number of steps with rewards in Asterix.}
	\label{fig_length_comparison} 
\end{figure}

Beyond the comparison of Figure \ref{fig_entropy_improvements_analysis}, we further measure SD-SAC and DSAC in terms of episode length and number of steps with rewards in Figure \ref{fig_length_comparison}. After $2e6$ steps, the algorithm with entropy penalty demonstrates significant longer episode lengths and more reward steps compared to discrete SAC. This indicates that the entropy penalty helps the agent learn both scoring and avoidance skills, leading to continued performance improvement.

\subsection{Result Curves of Individual Runs}
For clearer demonstration, we show the results of individual run curves for figures during our major analysis in Section 4 and 5.

\subsubsection{Comparison Between True Q Values and Estimate Q Values}
\begin{figure} [!t]
    \centering
    \subfigure[Discrete SAC\label{fig_underestimation_a_indiv}]{
        \begin{minipage}[b]{0.31\textwidth}
        \includegraphics[width=0.98\textwidth]{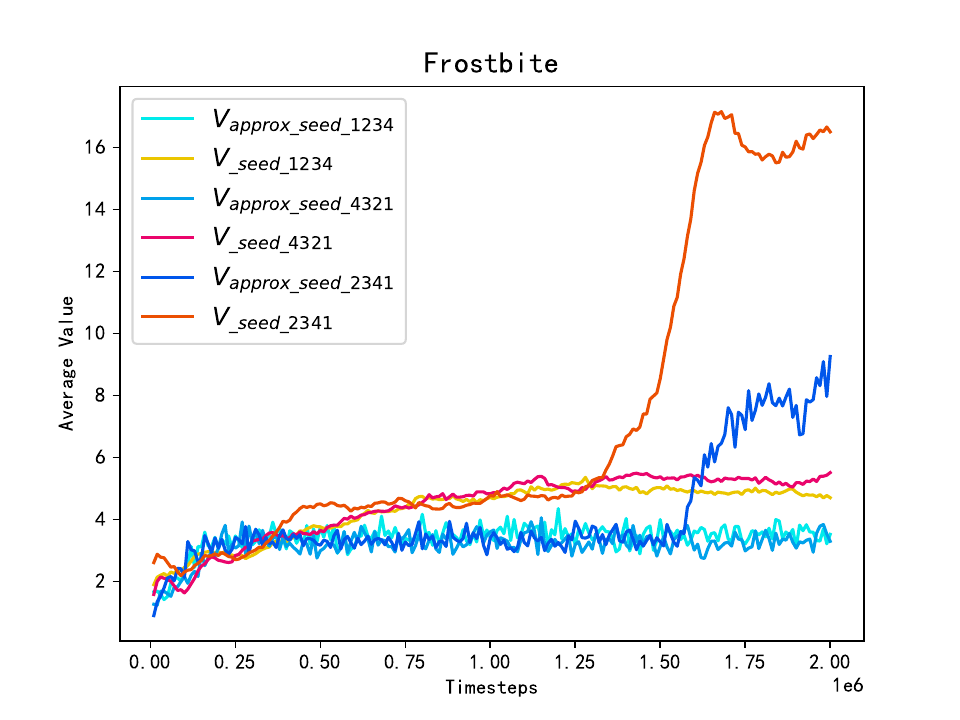}\\
	\includegraphics[width=0.98\textwidth]{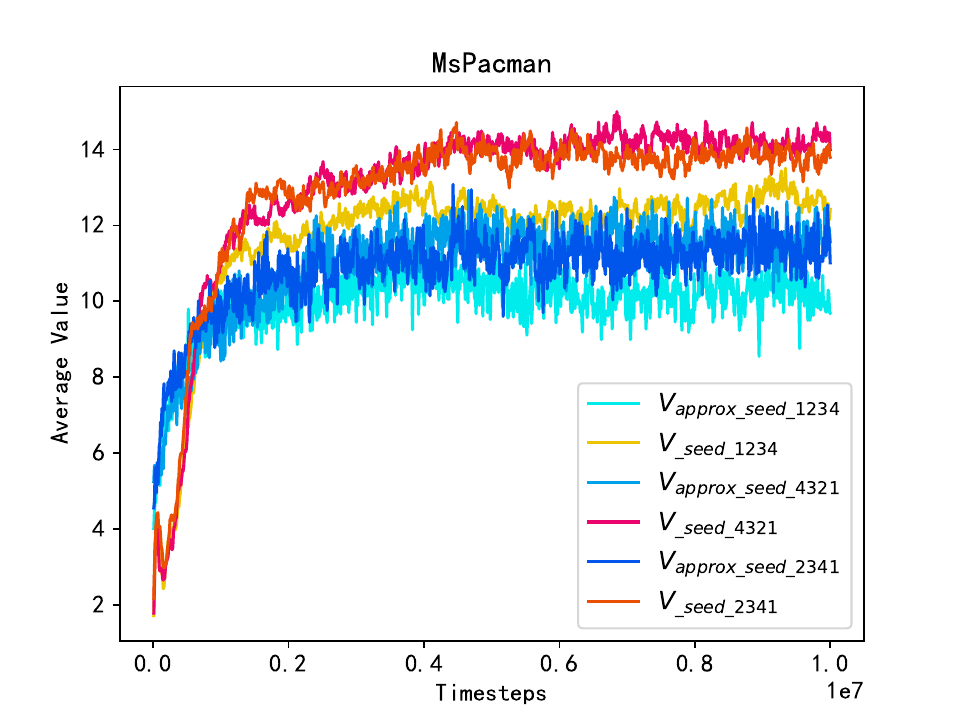}
        \end{minipage}
    }
    \subfigure[Single Q\label{fig_underestimation_b_indiv}]{
        \begin{minipage}[b]{0.31\textwidth}
        \includegraphics[width=0.98\textwidth]{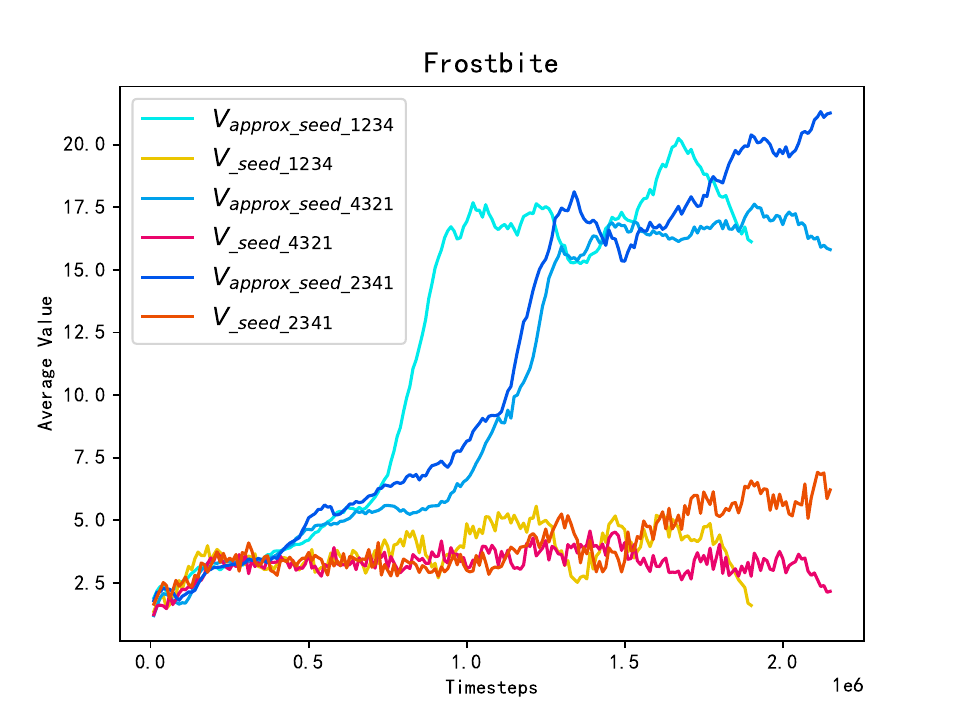}\\
	\includegraphics[width=0.98\textwidth]{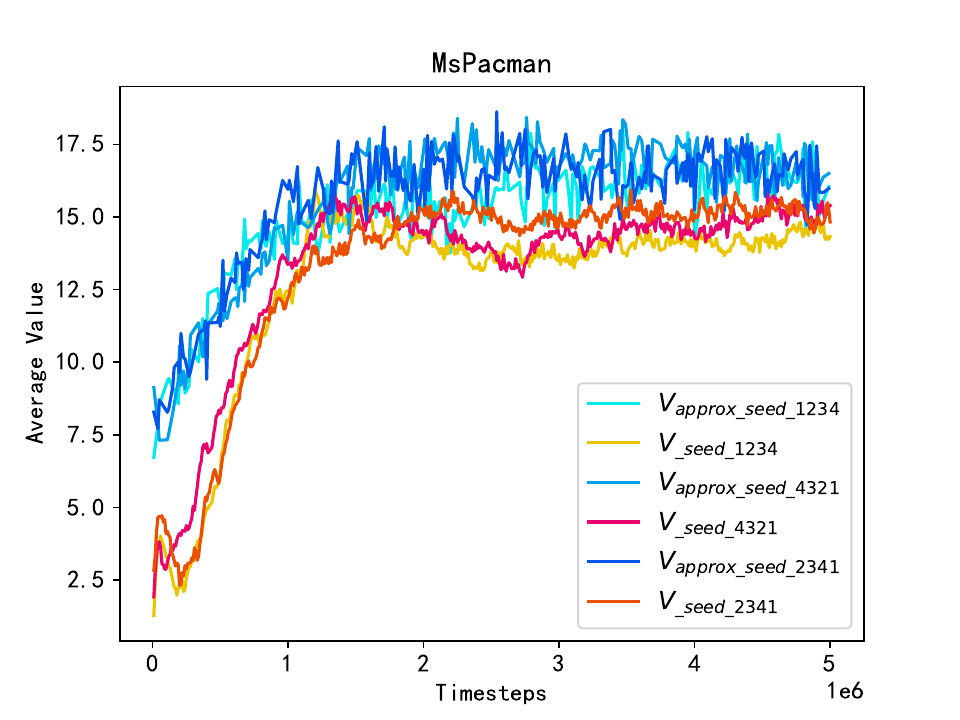}
        \end{minipage}
    }
    \subfigure[Score\label{fig_underestimation_c_indiv}]{
        \begin{minipage}[b]{0.31\textwidth}
        \includegraphics[width=0.98\textwidth]{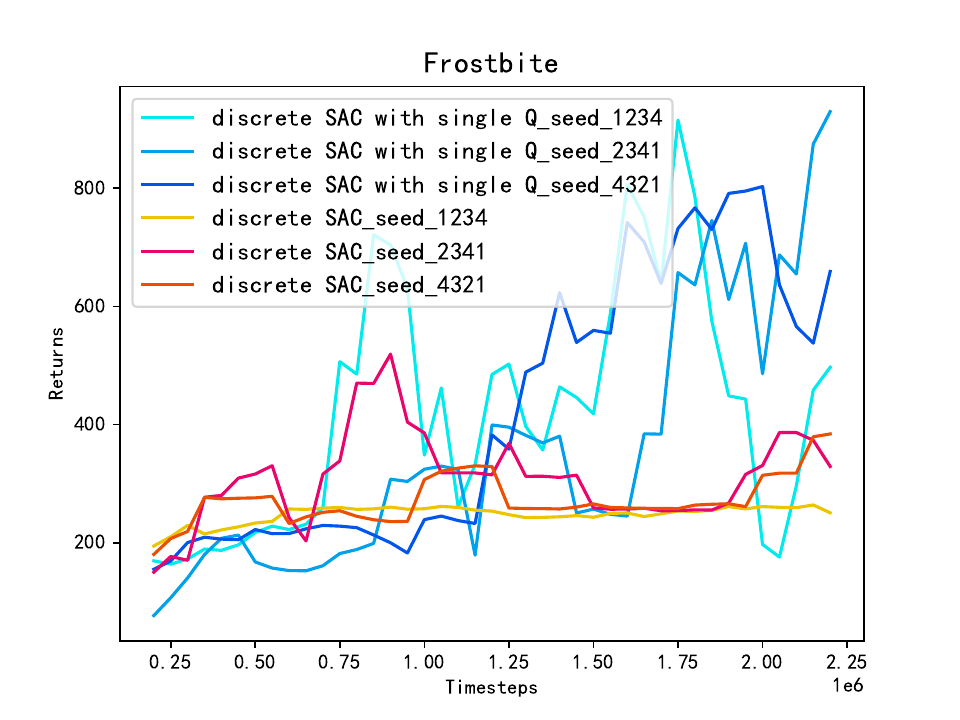}\\
	\includegraphics[width=0.98\textwidth]{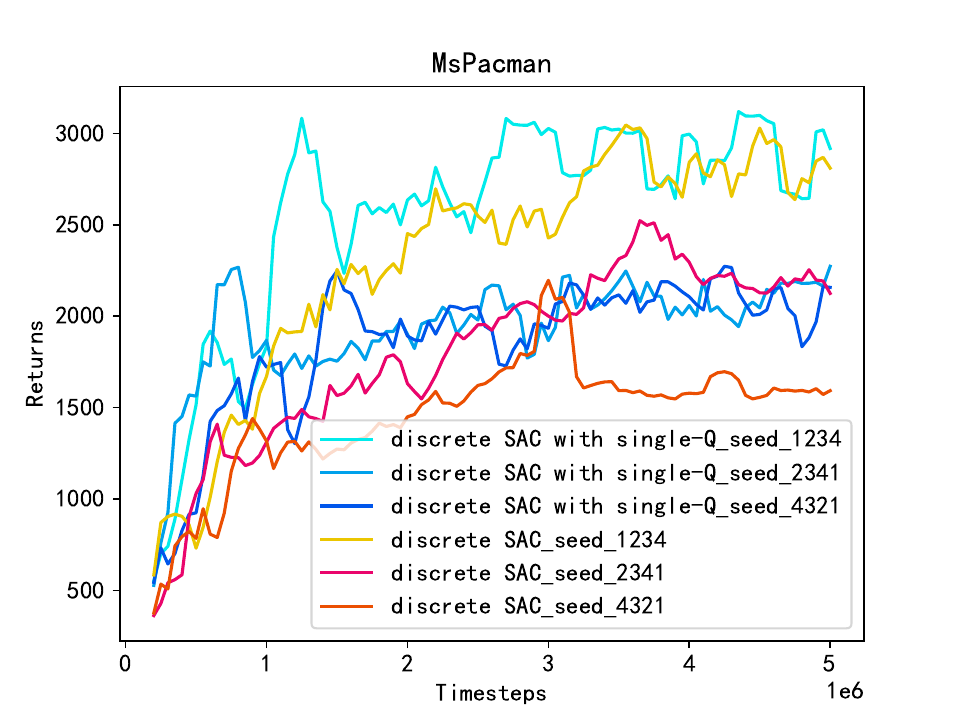}
        \end{minipage}
    }
\caption{The results of Atari game Frostbite/MsPacman environment over 2/5 million time steps: a) Measuring Q-value estimates of discrete SAC; b) Measuring Q-value estimates of discrete SAC with single Q; c) Score comparison between discrete SAC and discrete SAC with single Q.}
\label{fig_underestimation_indiv}
\end{figure}

We plot Figure \ref{fig_underestimation} by individual runs in Figure \ref{fig_underestimation_indiv}. The results show that for each individual seed, discrete SAC consistently suffers from an underestimation problem, while using a single Q leads to an overestimation issue.

\subsubsection{Comparison Between Different Policy Constraints}
\begin{figure} [!t]
    \centering
        \subfigure[Q Function Variance]{
		\includegraphics[width=0.40\textwidth]{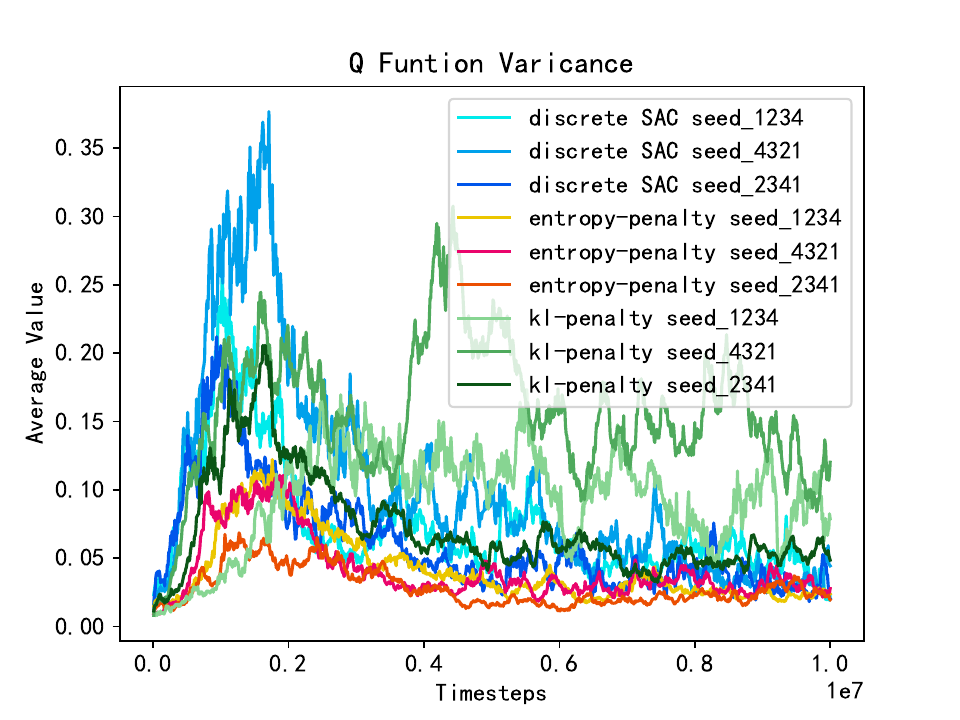}
		\label{subfig_improve_qvar_seed}
		}
	\subfigure[Entropy]{
		\includegraphics[width=0.40\textwidth]{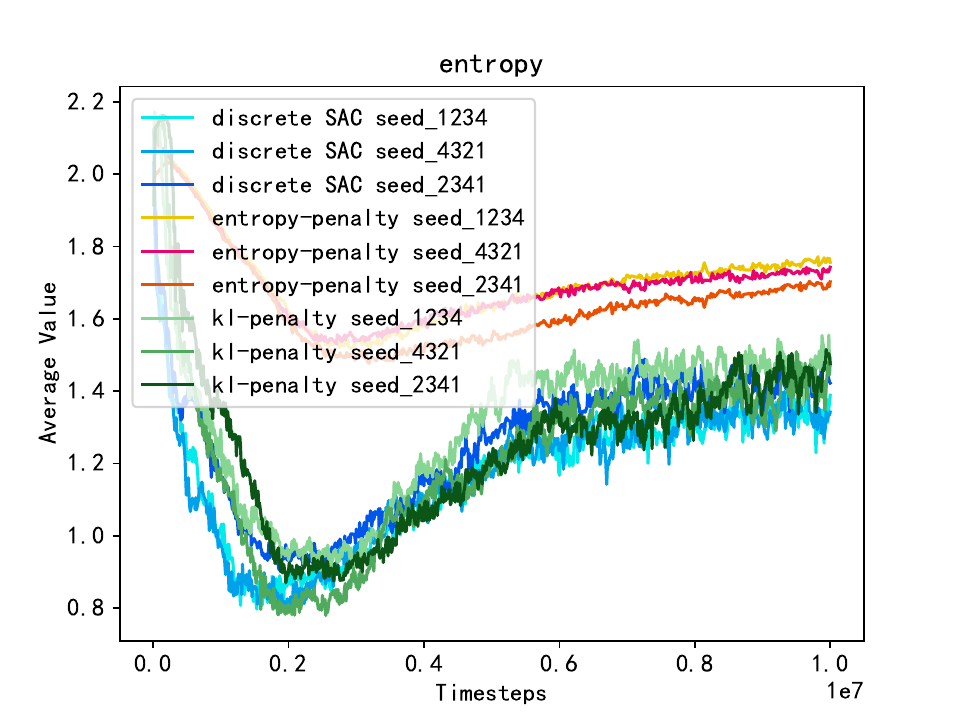}
		\label{subfig_improve_entropy_seed}
		}
	\subfigure[Q-value]{
		\includegraphics[width=0.40\textwidth]{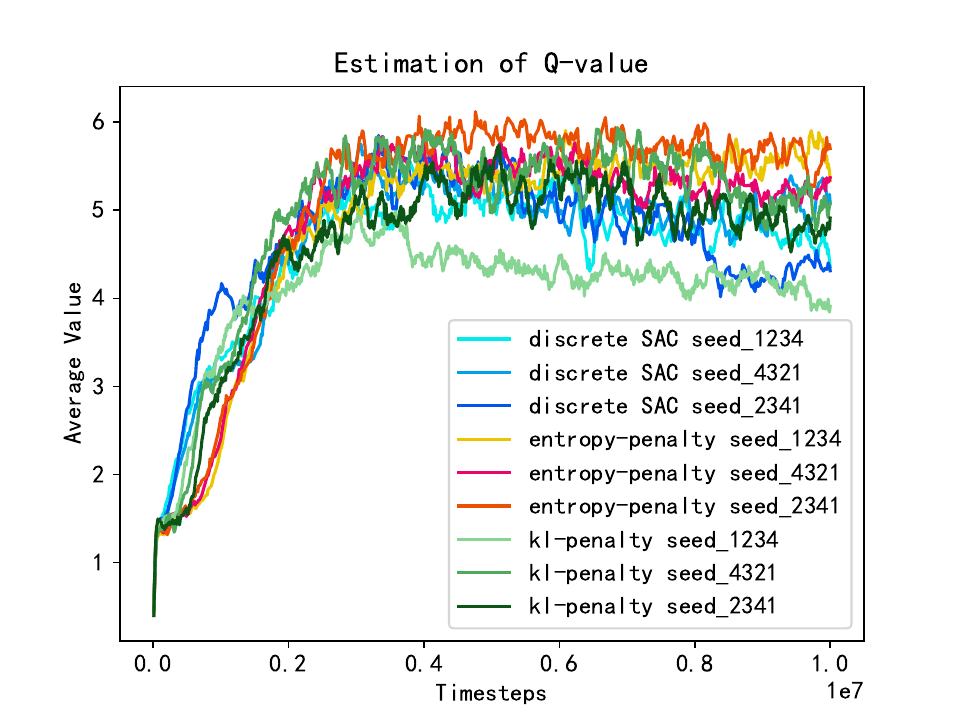}
		\label{subfig_improve_qvalue_seed}
		}
	\subfigure[Score]{
		\includegraphics[width=0.40\textwidth]{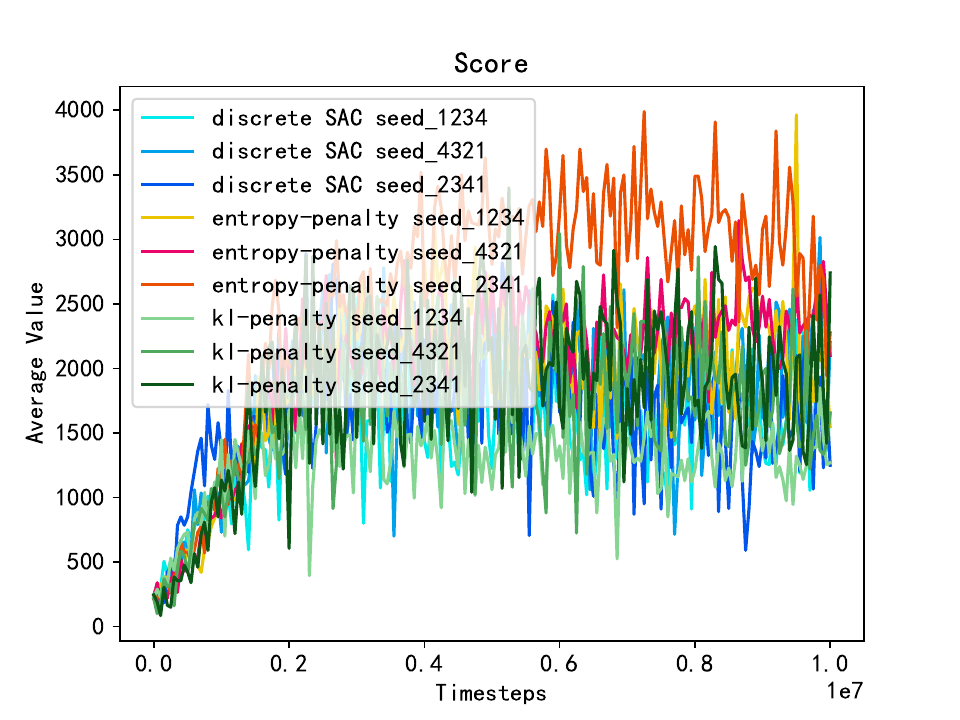}
	   \label{subfig_improve_return_seed}	
        }
	\caption{Measuring Q function variance, policy action entropy, estimation of Q-value, and score on Atari game Asterix compared between discrete SAC, discrete SAC with KL-penalty and discrete SAC with entropy-penalty over 10 million time steps.}
	\label{fig_entropy_improvements_analysis_indiv} 
\end{figure}

We provide Figure \ref{fig_entropy_improvements_analysis} by individual runs in Figure \ref{fig_entropy_improvements_analysis_indiv}. The results show that entropy penalty enables a stable training and better performance across all seeds.

\subsubsection{Comparison Between Different Q-value Estimation Methods}
\begin{figure} [htbp]
    \centering
    \subfigure[Discrete SAC]{
        \begin{minipage}[b]{0.38\textwidth}
        \includegraphics[width=1\textwidth]{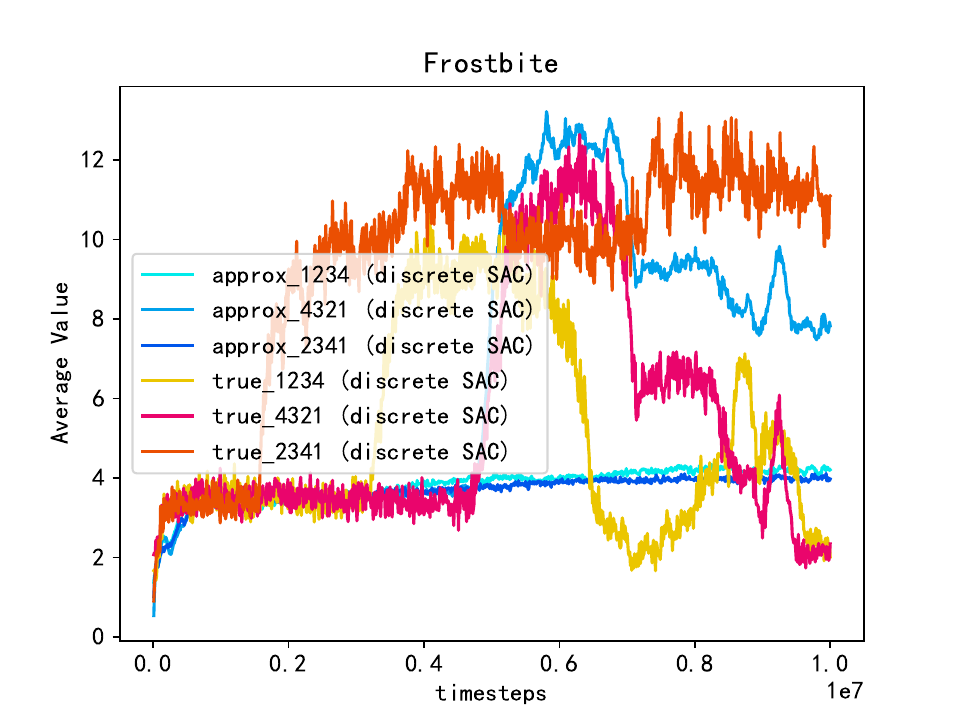}\\
	\includegraphics[width=1\textwidth]{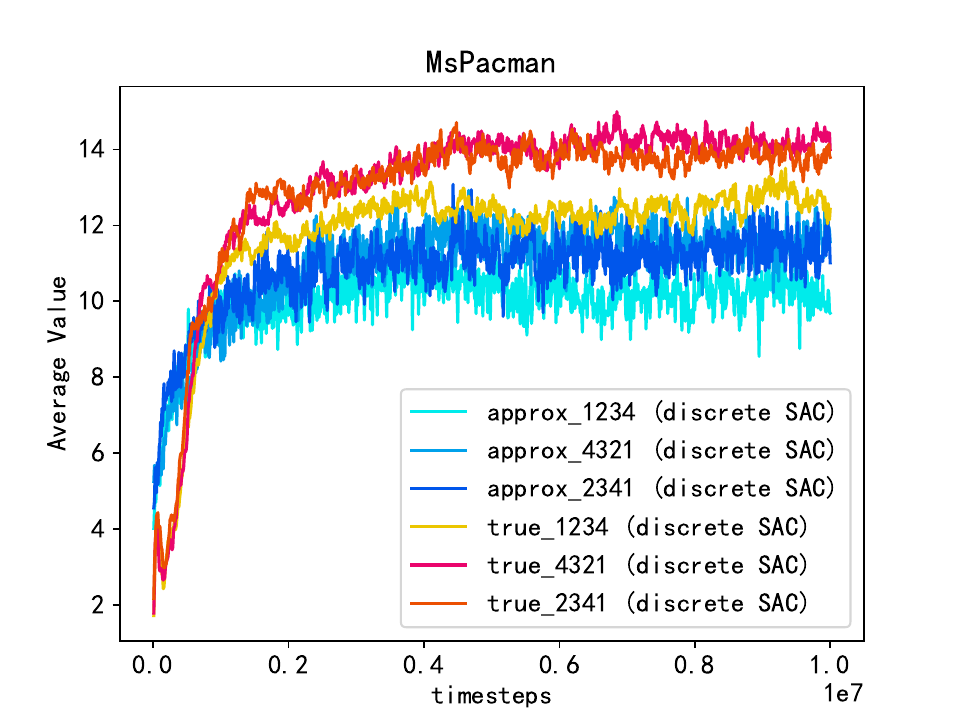}
        \end{minipage}
    }
    \subfigure[REDQ]{
        \begin{minipage}[b]{0.38\textwidth}
        \includegraphics[width=1\textwidth]{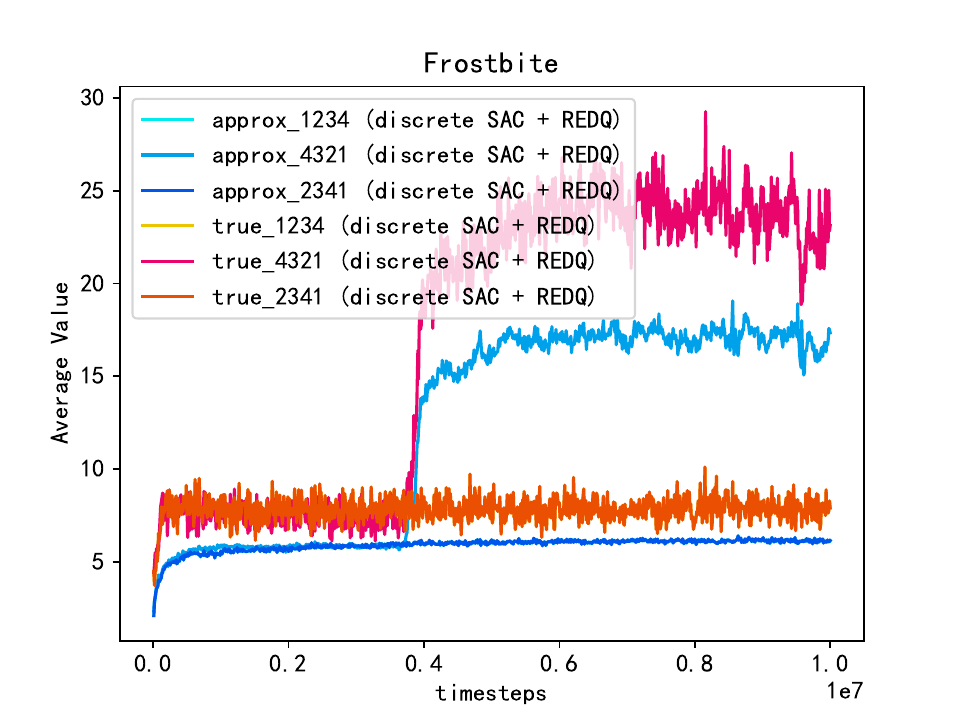}\\
	\includegraphics[width=1\textwidth]{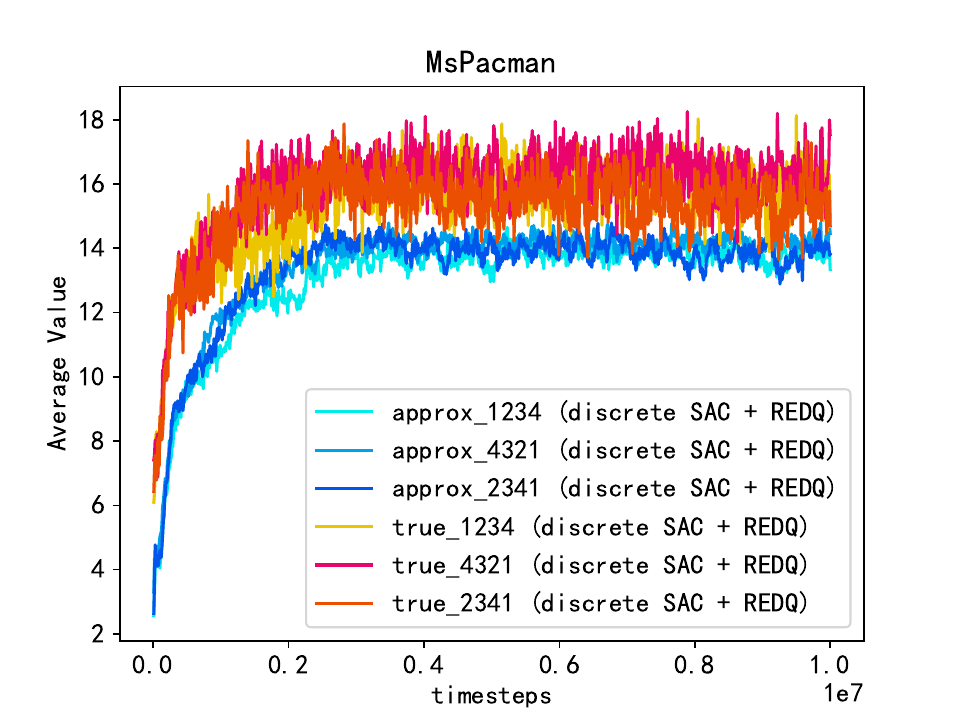}
        \end{minipage}
    }
    \subfigure[REM]{
        \begin{minipage}[b]{0.38\textwidth}
       \includegraphics[width=1\textwidth]{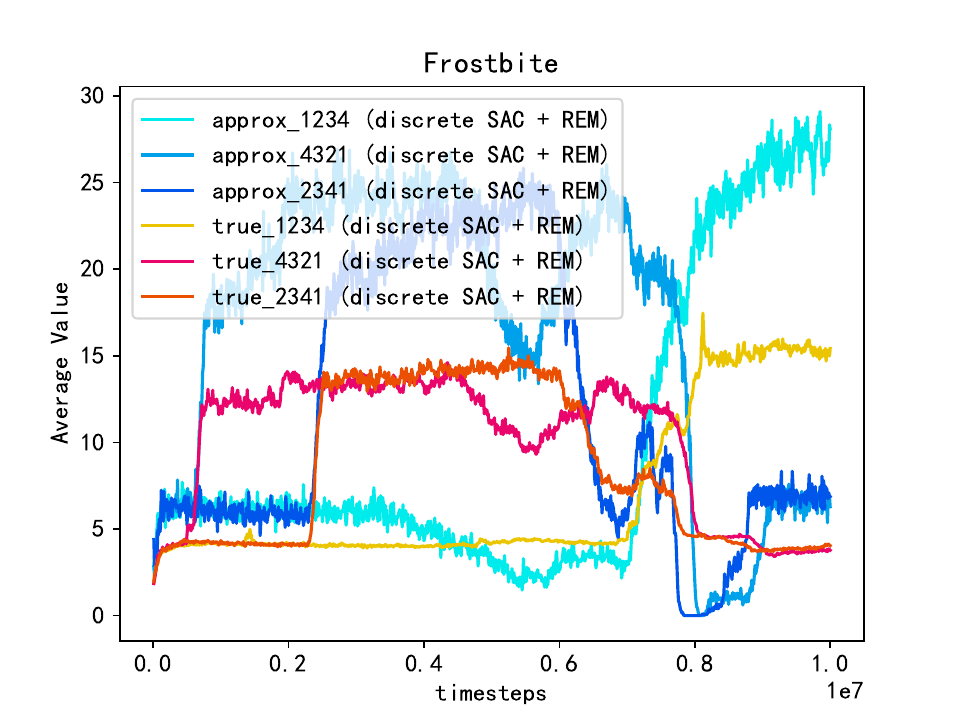}\\
	\includegraphics[width=1\textwidth]{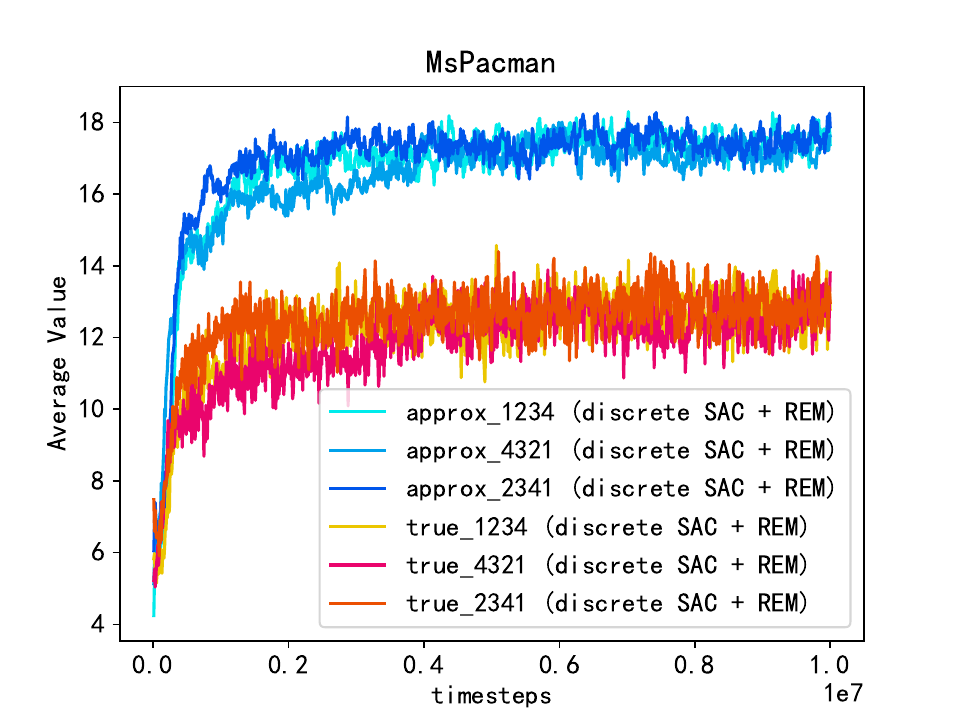}
        \end{minipage}
    }
    \subfigure[Ours]{
        \begin{minipage}[b]{0.38\textwidth}
        \includegraphics[width=1\textwidth]{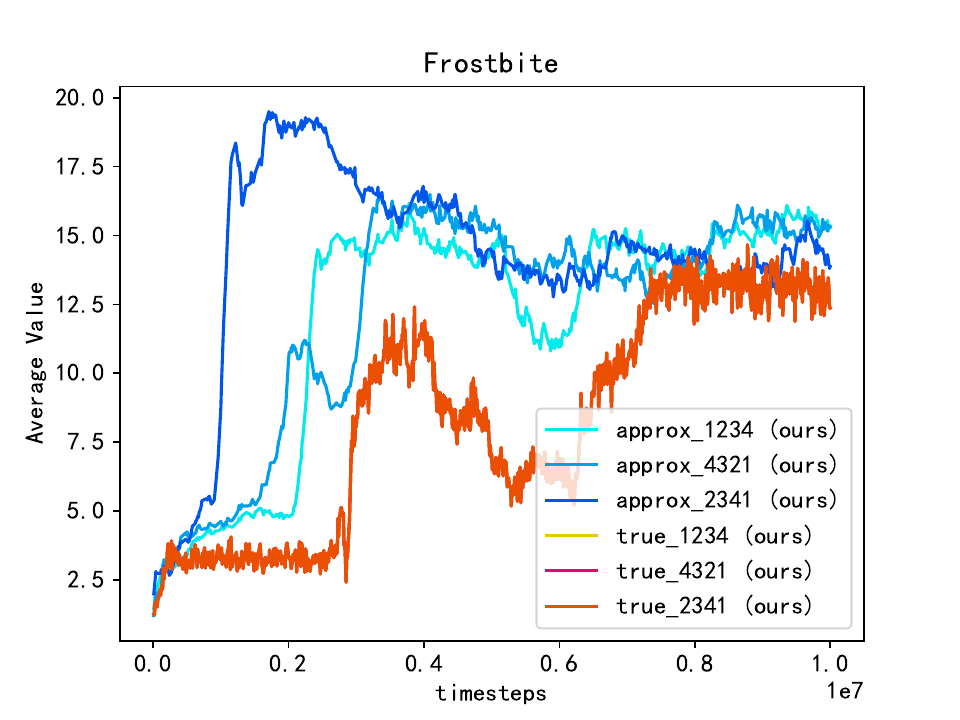}\\
	\includegraphics[width=1\textwidth]{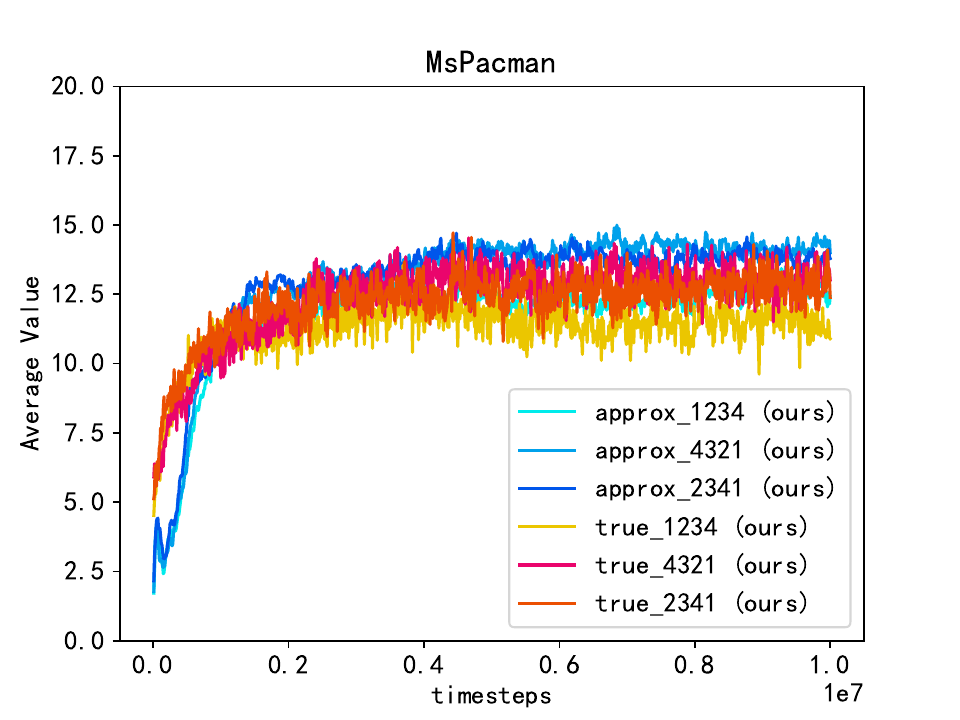}
        \end{minipage}
    }
	\caption{Measuring estimation of Q-value on Atari Game Frostbite/MsPacman environment compared between discrete SAC, discrete SAC with REDQ, discrete SAC with REM, and SD-SAC (discrete SAC with double average Q-learning with Q-clip) over 10 million steps.}
	\label{fig_doubelavgq_improvement_indiv} 
\end{figure}

We provide Figure \ref{fig_doubelavgq_improvement} by individual runs in Figure \ref{fig_doubelavgq_improvement_indiv}. Our approach demonstrates effectiveness in alleviating underestimation and reduce bias in all individual runs.

\subsection{Hyperparameter Used in SD-SAC}
Please refer to Table \ref{table-Hyperparameter}.

\begin{table}[!t]
\centering
\caption{Hyperparameter for Discrete SAC and SD-SAC}
\label{table-Hyperparameter}

\begin{tabular}{c|cc}
\toprule
     Hyperparameter & Discrete SAC & SD-SAC \\ \hline
     learning rate & $  10^{-5}$ & $  10^{-5}$ \\
     optimizer & Adam & Adam \\
     mini-batch size & 64 & 64 \\
     discount ($\gamma$) & 0.99 & 0.99\\
     buffer size & $10^5$ & $10^5$ \\
     hidden layers & 2& 2 \\
     hidden units per layer & 512 & 512  \\
     target smoothing coefficient ($\tau$) &  0.005 & 0.005 \\
    Learning iterations per round & 1 & 1 \\
    alpha & 0.05 & 0.05 \\
    n-step & 3 & 3 \\
    $\beta$ &  False & 0.5 \\
    $c$ &  False & 0.5 \\

    % Auto-alpha & NE & NE \\

% Mean    & 0.5\%           & 3.0\%        &  
% \textbf{41.7}\% & 38.5\%   & 187.4 \% & 151.4\% & 199.2\% & \textbf{220.0\%} \\  
% Median  & 0.4\%           & 2.1\%   & \textbf{20.0\%}     & 11.1\%   & 79.2 \% &90.8\% & 107.7\%  & \textbf{114.1\%}  \\
\bottomrule
\end{tabular}

\end{table}

\subsection{Introduction of the ELO System}
\label{appendix_elo}
The ELO rating system~\citep{elo1978rating} is a widely-used mechanism for assessing the relative skill levels of players or agents, commonly applied in chess, competitive games, and other adversarial environments. In our study, the ELO ratings of agents are calculated through the following process:

\begin{enumerate}
    \item Each agent is assigned an initial ELO rating $R_{base}$.
    \item Before agent A competes against agent B, the expected score for each agent is calculated based on their current ELO ratings: $E_A=\frac{1}{1+10^{\left(R_B-R_A\right) / R_{base}}}$
    \item The ELO ratings are updated based on the outcome of the match between A and B, where $S_A$ is the actual score and $K$ is a constant: $R_A^{\prime}=R_A+K \cdot\left(S_A-E_A\right)$
    \item Through multiple matches among various agents, the ELO ratings are adjusted according to the results of these matches. The final ELO ratings reflect each agent's relative strength compared to the others.
\end{enumerate}

\subsection{Other Atari Environments for Unstable Coupling Training of Discrete SAC}
\label{appendix_environments}
We conduct cross-validation in other Atari environments, as presented in Fig. \ref{fig_example_assault}-\ref{fig_example_mspacman} . The result show that in other environments with deceptive rewards, the rapid decrease in policy entropy due to larget Q variance similarly affects training.

\subsubsection{Games with Deceptive Rewards}
\begin{figure} [htbp]
    \centering
	\subfigure[Assault]{
		\includegraphics[width=0.25\textwidth]{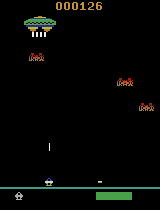}
		\label{subfig_deceptive_games_example_assault}
		}
	\subfigure[Jamesbond]{
		\includegraphics[width=0.25\textwidth]{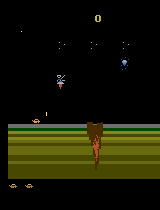}
		\label{subfig_deceptive_games_example_jamesbond}
		}
	\subfigure[MsPacman]{
		\includegraphics[width=0.25\textwidth]{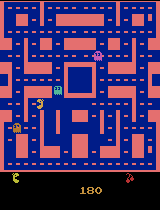}
		\label{subfig_deceptive_games_example_mspacman}
		}
	\caption{Three examples of Atari game environments with deceptive rewards.}
	\label{fig_deceptive_games_example} 
\end{figure}

We take the Atari games Assault, Jamesbond and MsPacman as examples to further illustrate the manifestation and impact of deceptive rewards in game environments. (Fig. \ref{fig_deceptive_games_example})

In the game Assault, a mothership releases different kinds of aliens. They move along the screen, with the bottom-most alien firing various types of weapons. The player controls a cannon that can shoot bullets horizontally or vertically to attack the aliens and fireballs they shot. Hitting an alien scores points, while being hit or cannon overheating results in a loss of life.

In Jamesbond, the player controls a craft that needs to complete various mission to achieve final victory. In the first mission, the player must navigate through a desert with craters, acoid overhead satellite scans and helicopter bombings, and score points by hitting diamonds through fixed-angle shooting.

As for MsPacman, the player controls a Pacman, who scores points by eating dots in a maze while avoiding floating ghosts. When Pacman eats an energy pill, she can attack the ghosts to gain higher scores.

In all three environments, the agent can quickly gain deceptive rewards through short-term payoffs. For Assault and Jamesbond, all points come from shooting actions that hit specific targets, while avoiding obstacles can prevent the loss of life but does not bring clear rewards. Thus, agents often excel at shooting but struggle with dodging. In MsPacman, the numerous dots in the maze provide many rewards for the agent's movement. As a result, the agent finds it difficult to learn advances strategies such as avoiding ghosts and picking up energy pills to attack ghosts. The presence of deceptive rewards leads to the training process stuck in local optima, making it challenging to explore better, long-term strategies.

\subsubsection{Plots of Training Process}
\begin{figure} [htbp]
    \centering
	\subfigure[Q Function Variance]{
		\includegraphics[width=0.3\textwidth]{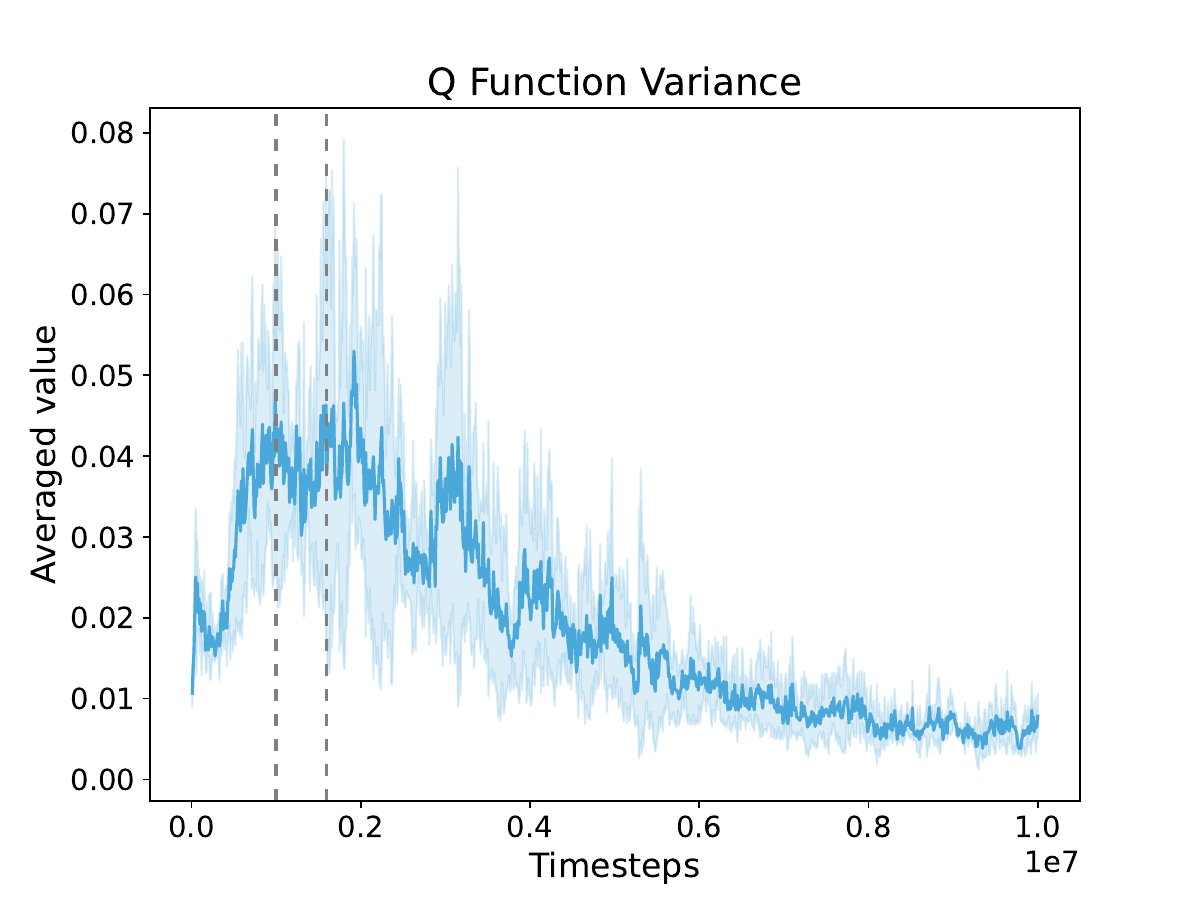}
		}
	\subfigure[Q-value]{
		\includegraphics[width=0.3\textwidth]{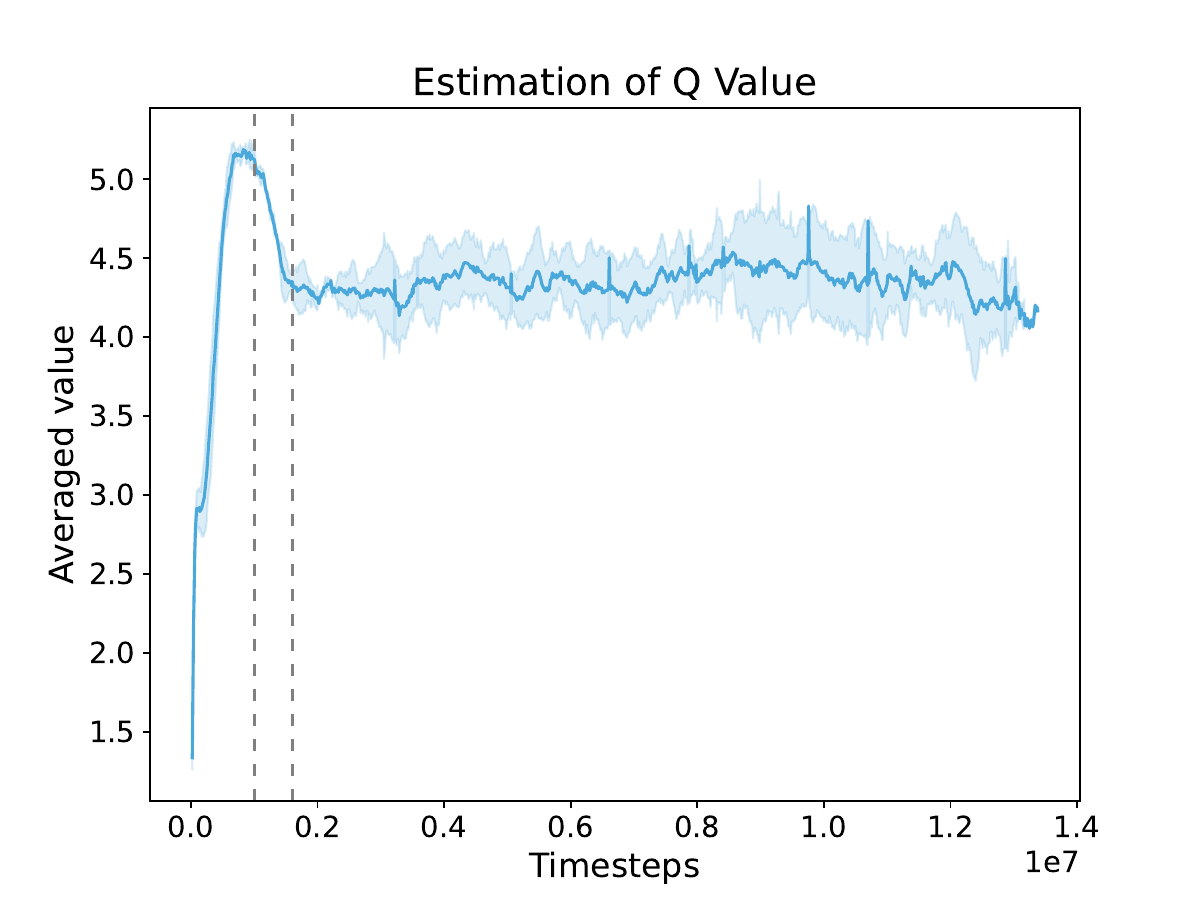}
		}
	\subfigure[Entropy]{
		\includegraphics[width=0.3\textwidth]{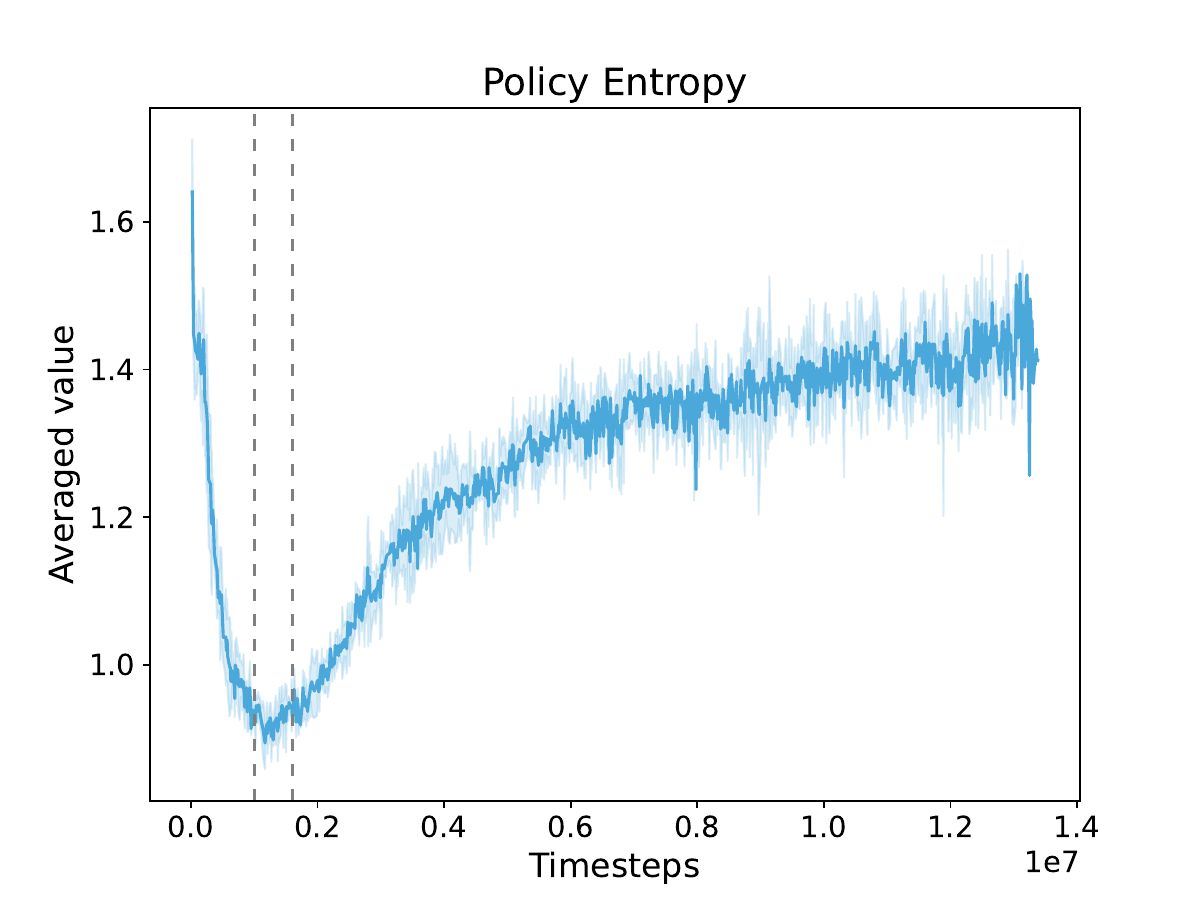}
		}
        \\
  	\subfigure[Episode Length]{
		\includegraphics[width=0.3\textwidth]{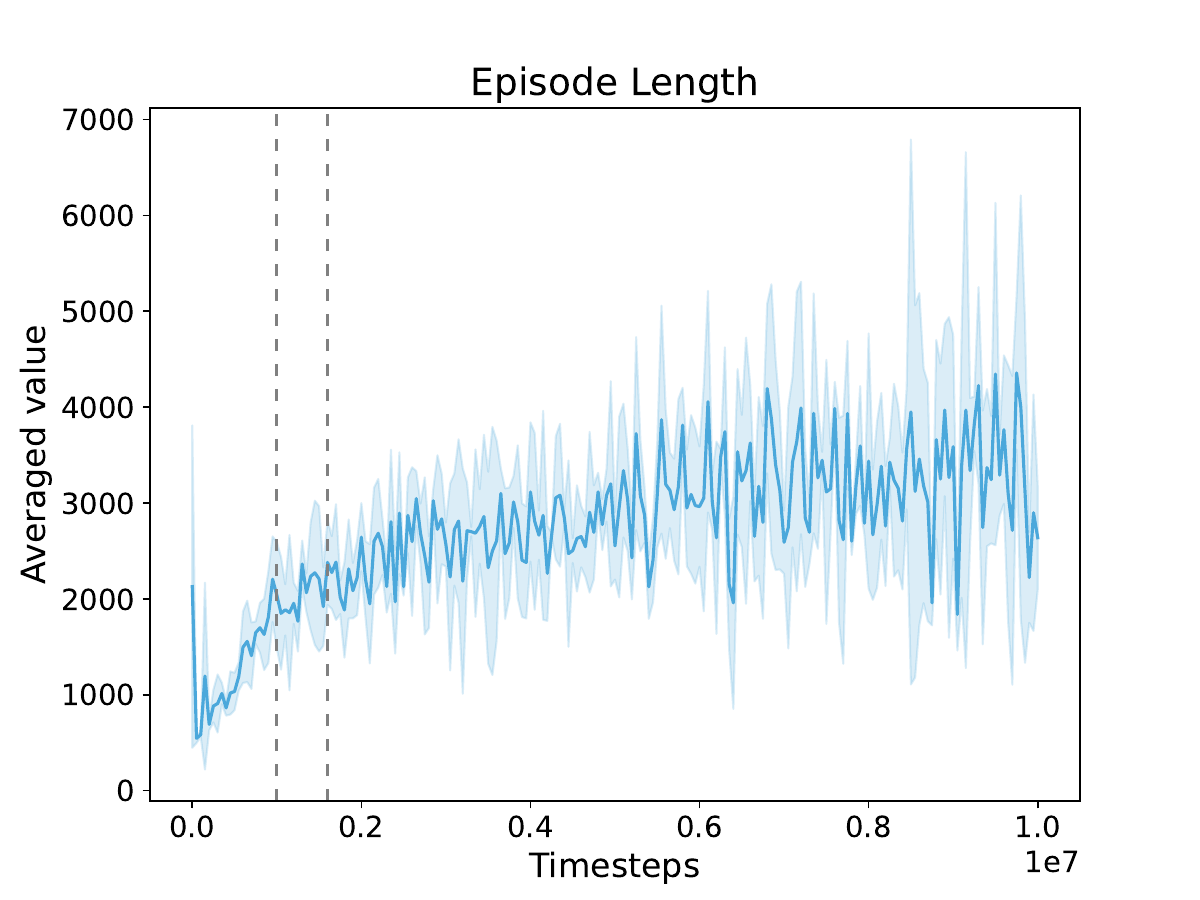}
		}
        \subfigure[Steps with Rewards]{
		\includegraphics[width=0.3\textwidth]{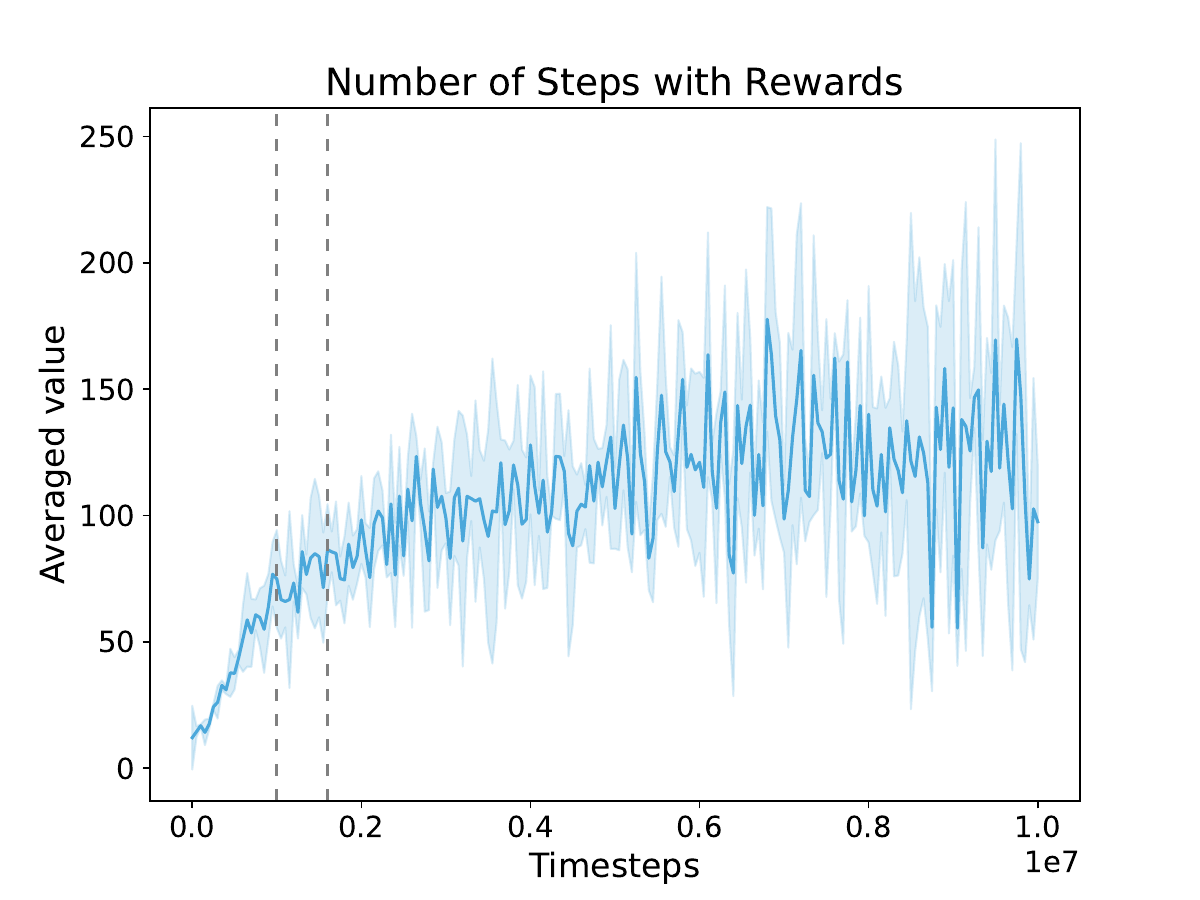}
		}
        \subfigure[Score]{
		\includegraphics[width=0.3\textwidth]{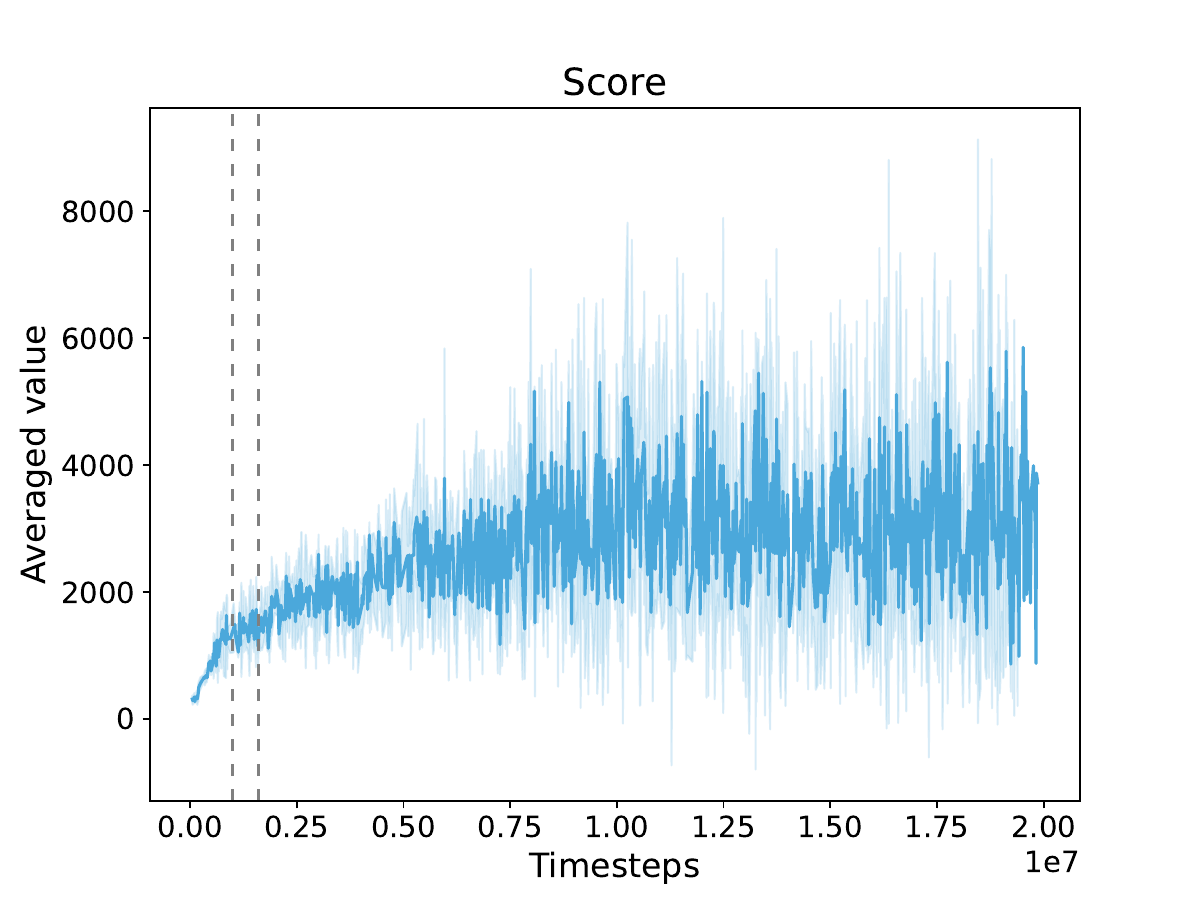}
		}
	\caption{Plots of Q function variance, estimation of Q-value, policy action entropy, episode length, number of steps with rewards and score on Atari Game Assault environment with discrete SAC over 10 million time steps.}
	\label{fig_example_assault} 
\end{figure}

\begin{figure} [htbp]
    \centering
	\subfigure[Q Function Variance]{
		\includegraphics[width=0.3\textwidth]{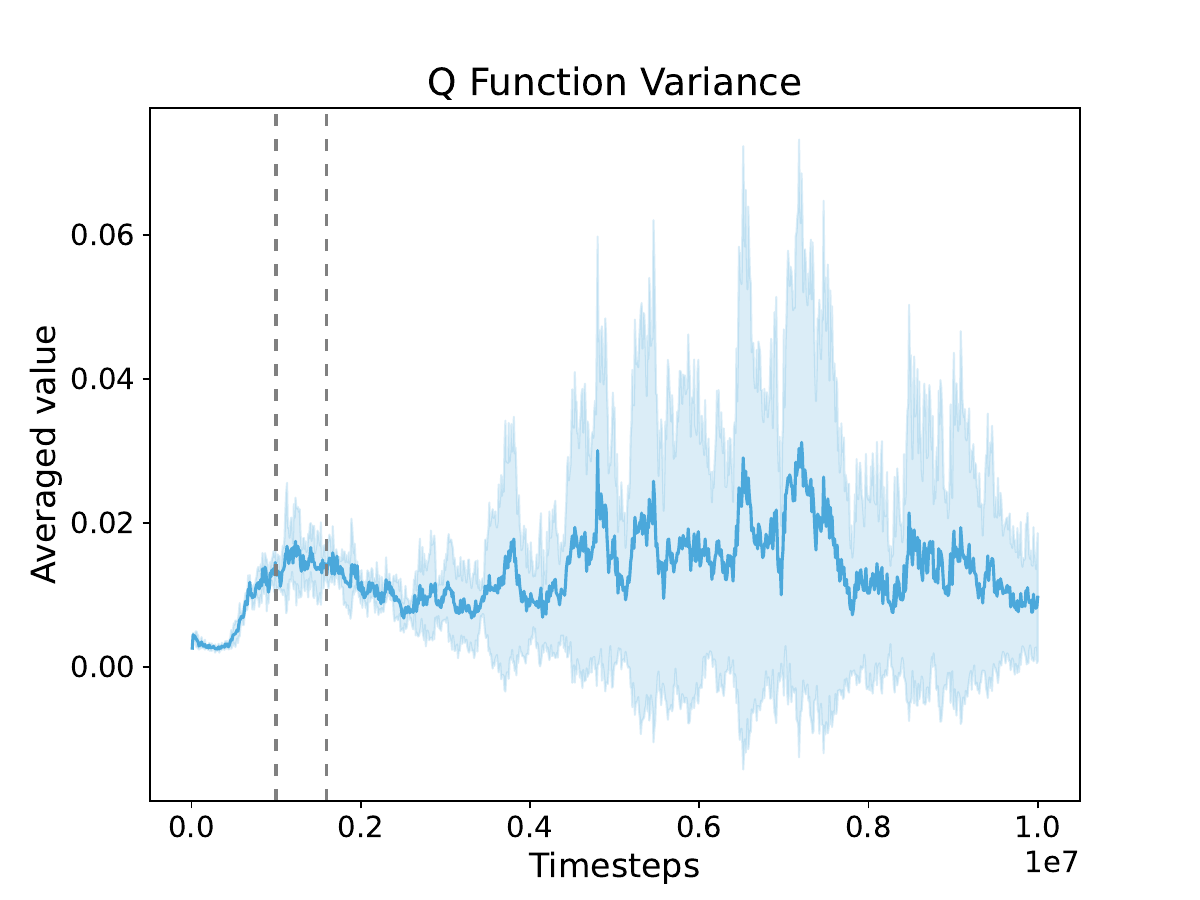}
		}
	\subfigure[Q-value]{
		\includegraphics[width=0.3\textwidth]{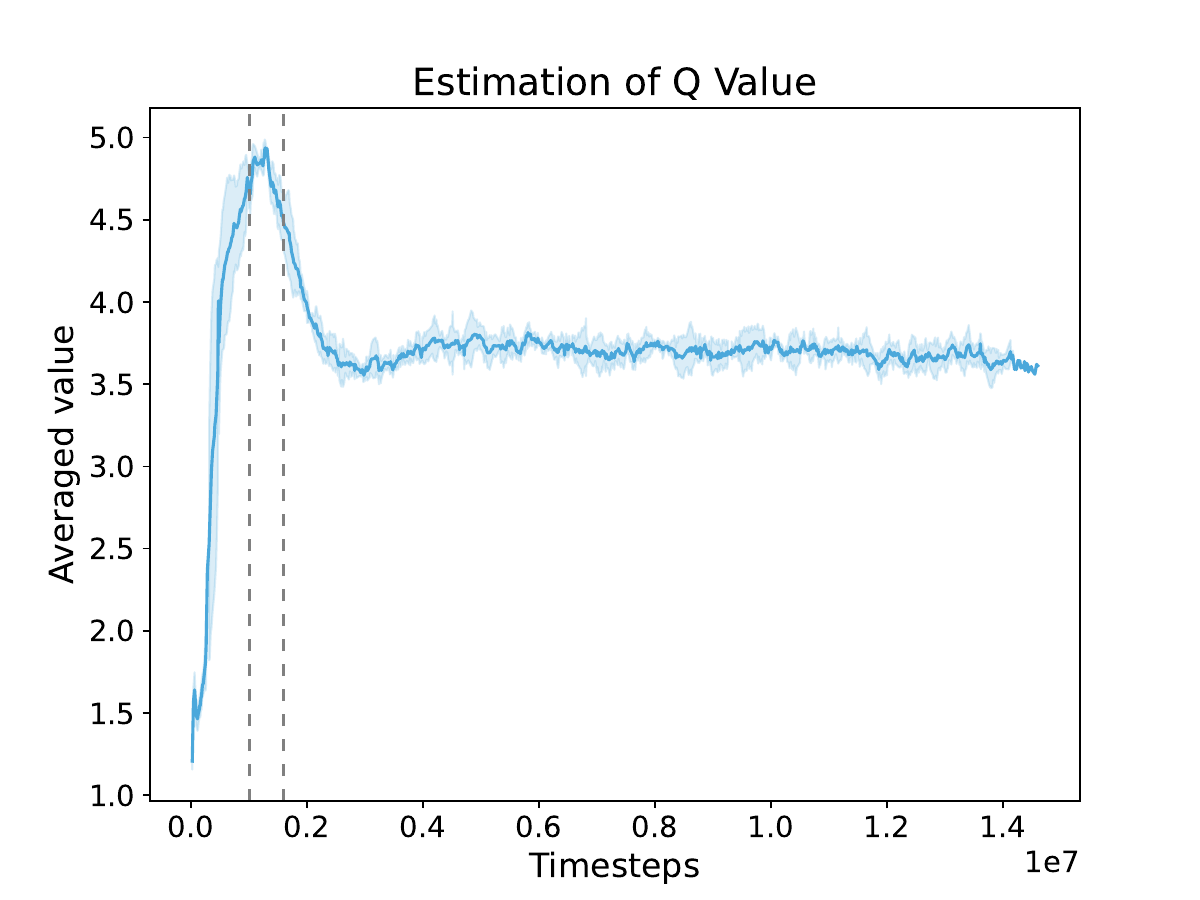}
		}
	\subfigure[Entropy]{
		\includegraphics[width=0.3\textwidth]{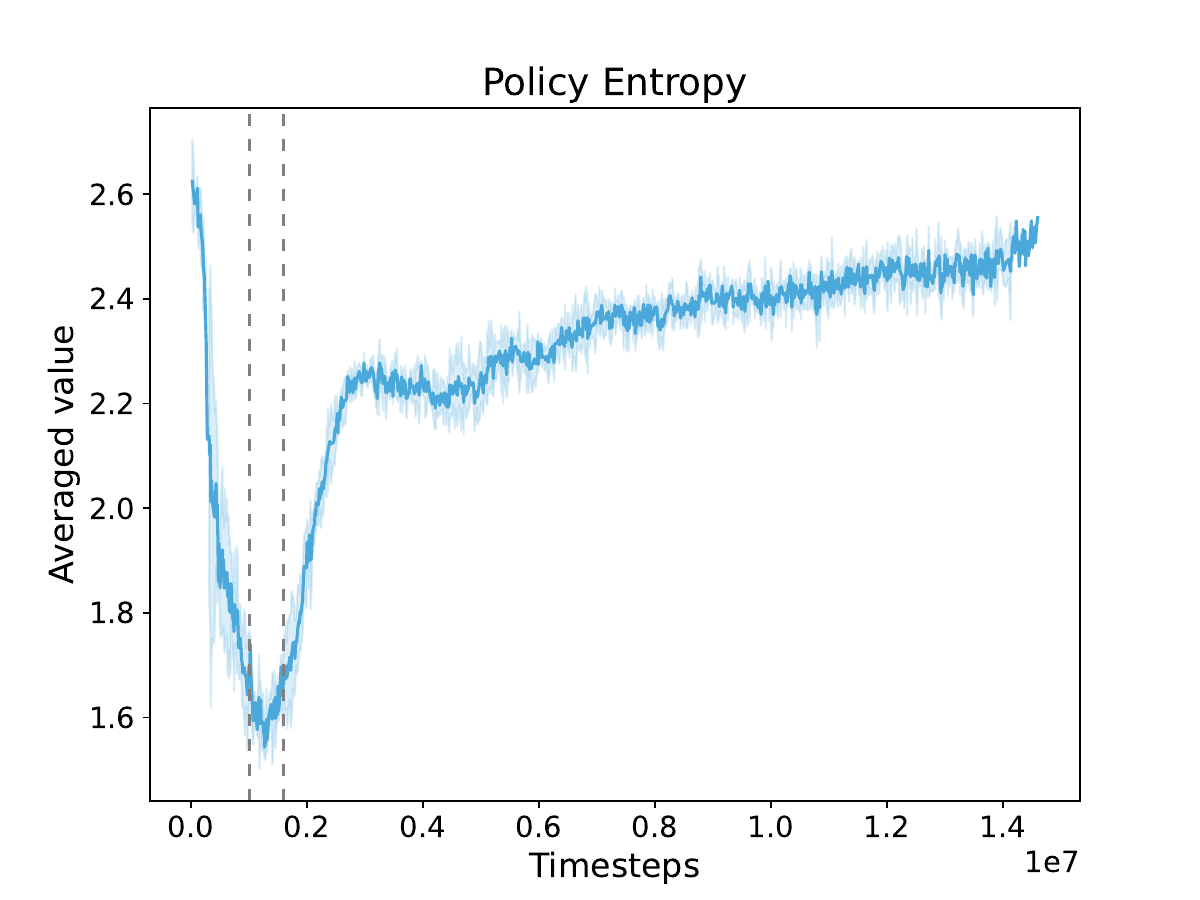}
		}
        \\
  	\subfigure[Episode Length]{
		\includegraphics[width=0.3\textwidth]{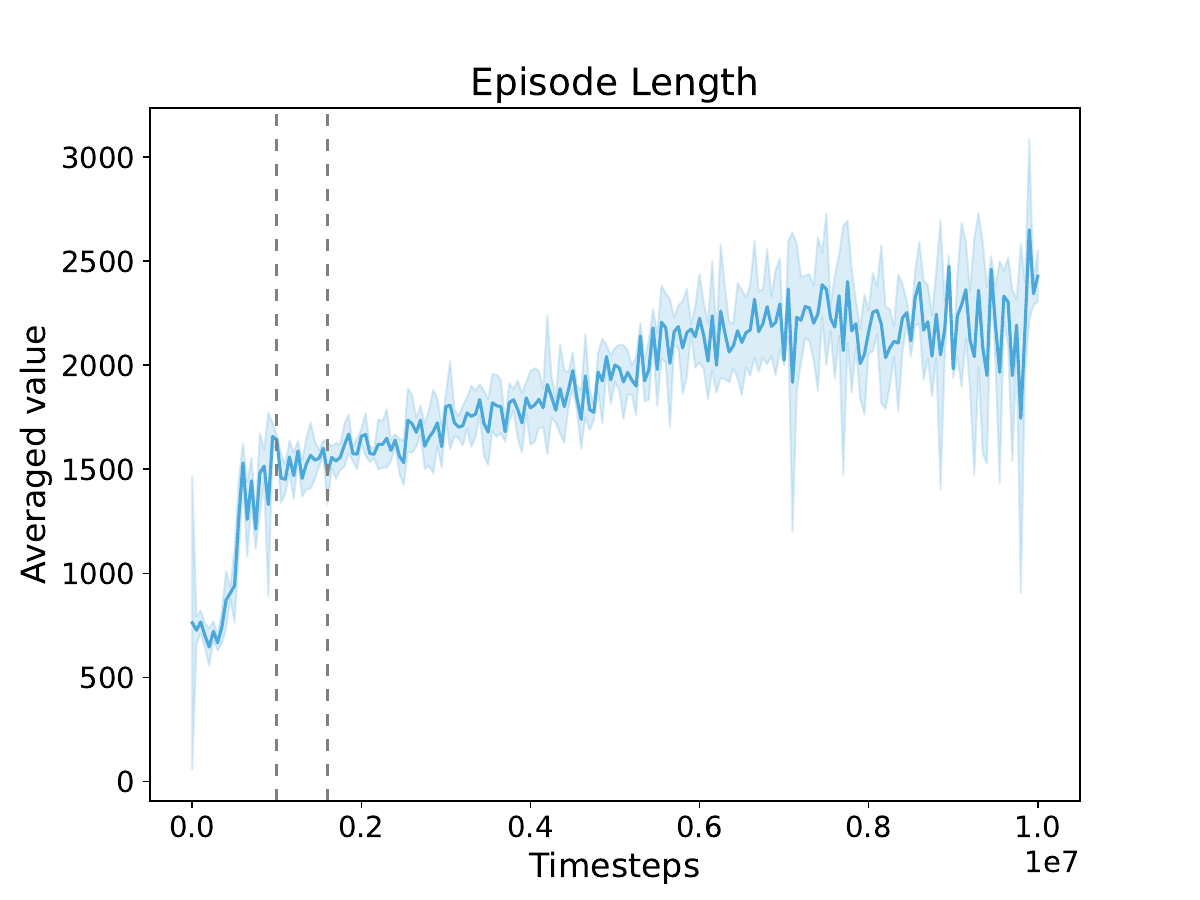}
		}
        \subfigure[Steps with Rewards]{
		\includegraphics[width=0.3\textwidth]{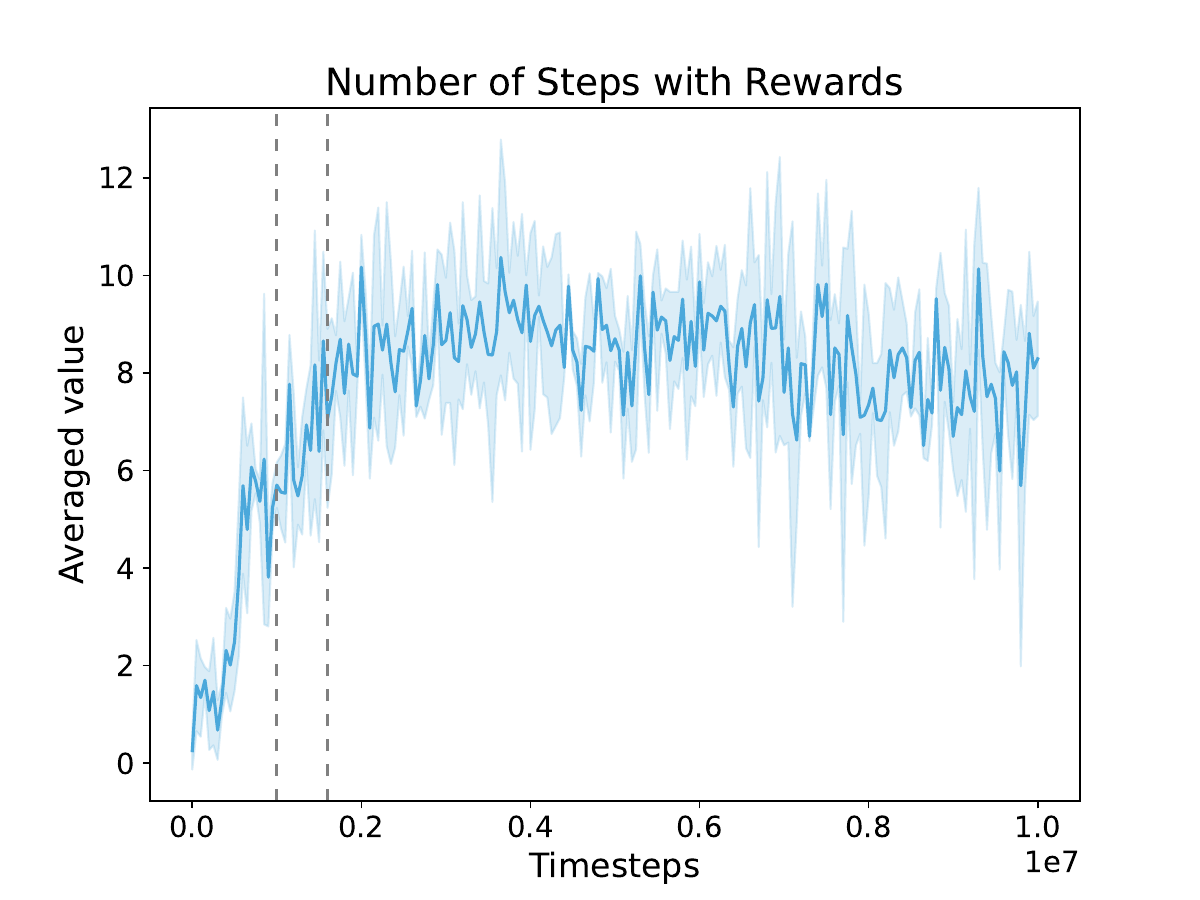}
		}
        \subfigure[Score]{
		\includegraphics[width=0.3\textwidth]{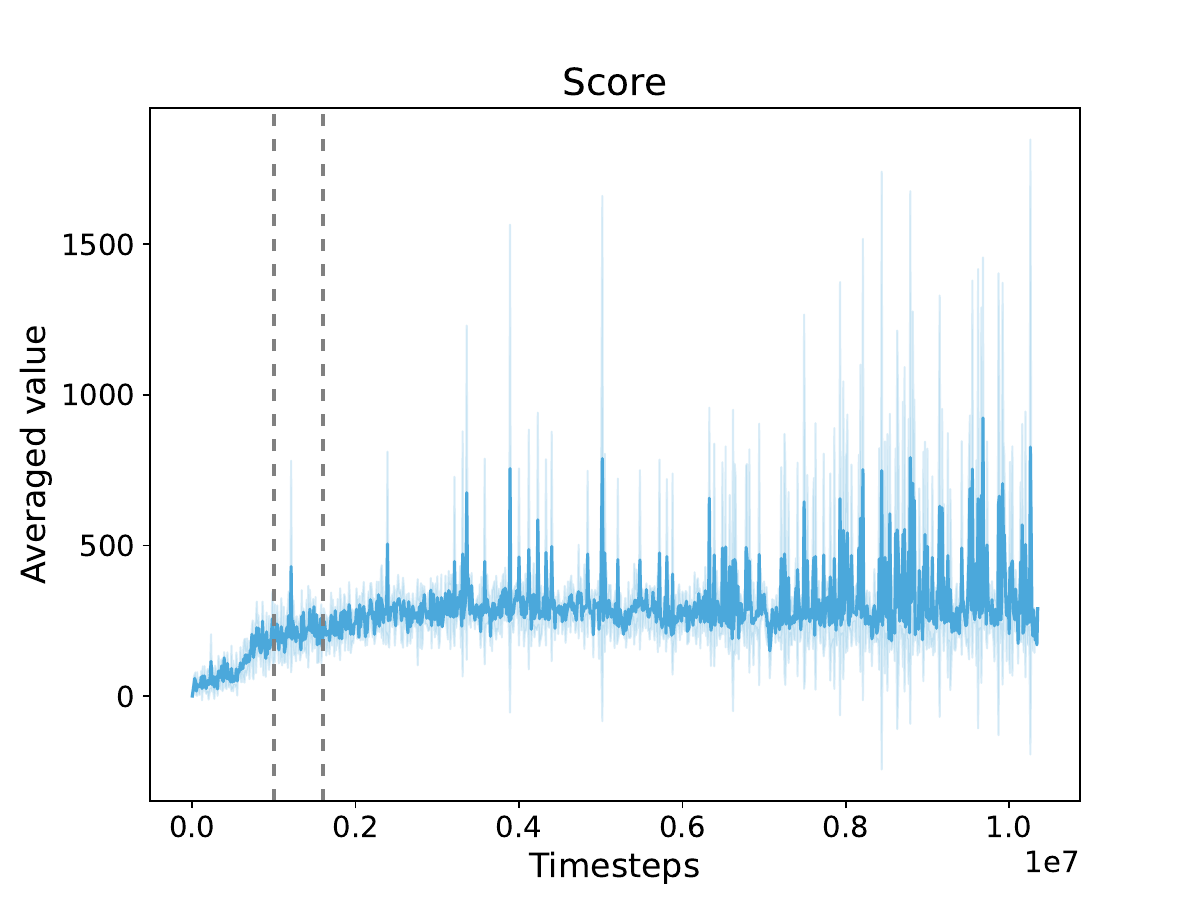}
		}
	\caption{Plots of Q function variance, estimation of Q-value, policy action entropy, episode length, number of steps with rewards and score on Atari Game Jamesbond environment with discrete SAC over 10 million time steps.}
	\label{fig_example_jamesbond} 
\end{figure}

\begin{figure} [htbp]
    \centering
	\subfigure[Q Function Variance]{
		\includegraphics[width=0.3\textwidth]{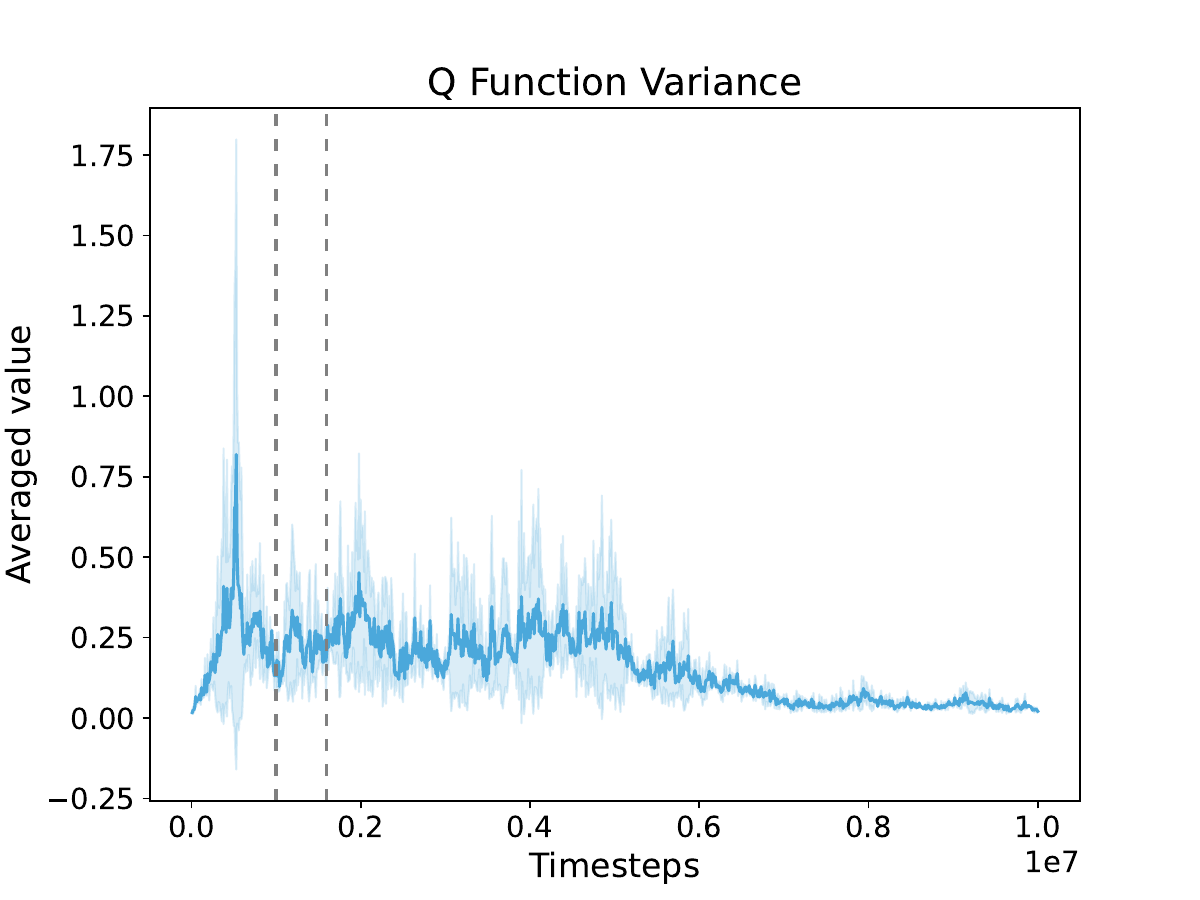}
		}
	\subfigure[Q-value]{
		\includegraphics[width=0.3\textwidth]{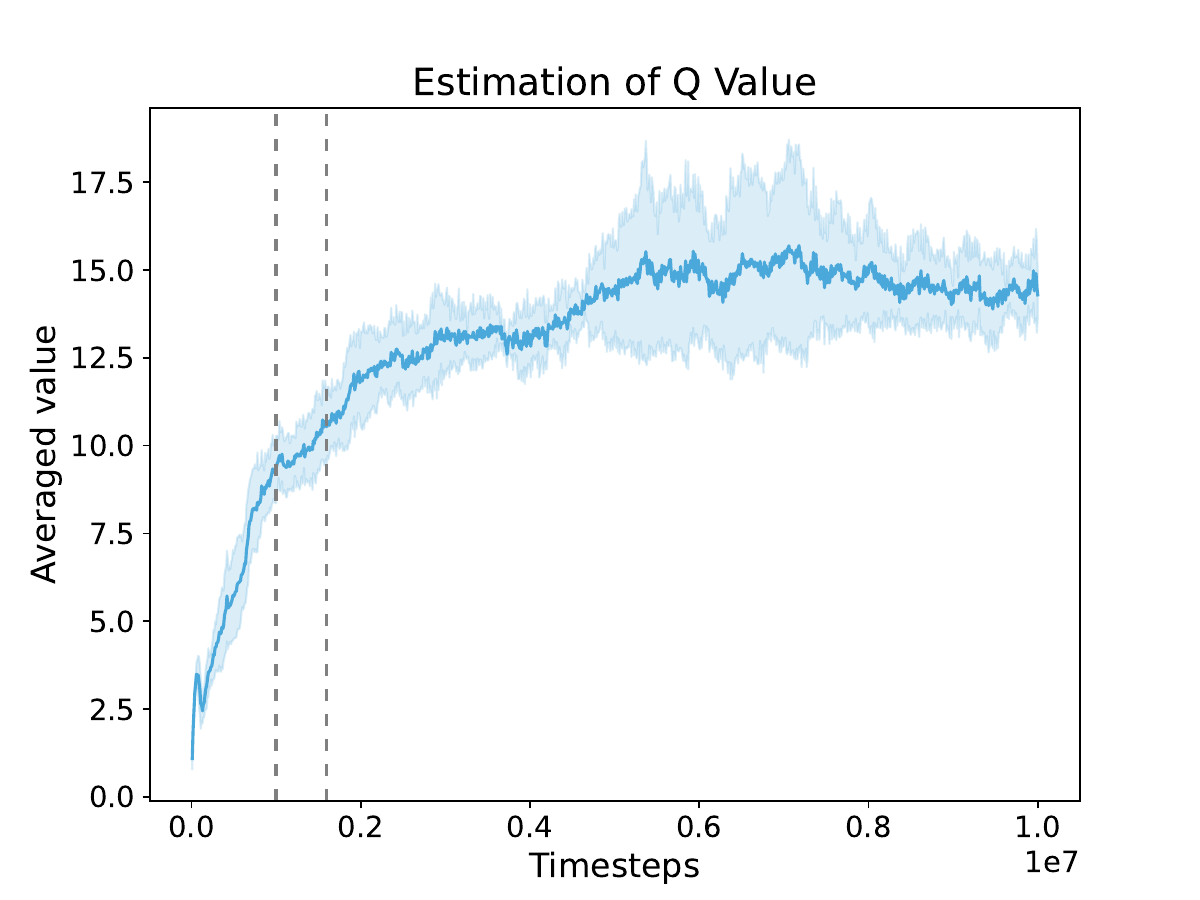}
		}
	\subfigure[Entropy]{
		\includegraphics[width=0.3\textwidth]{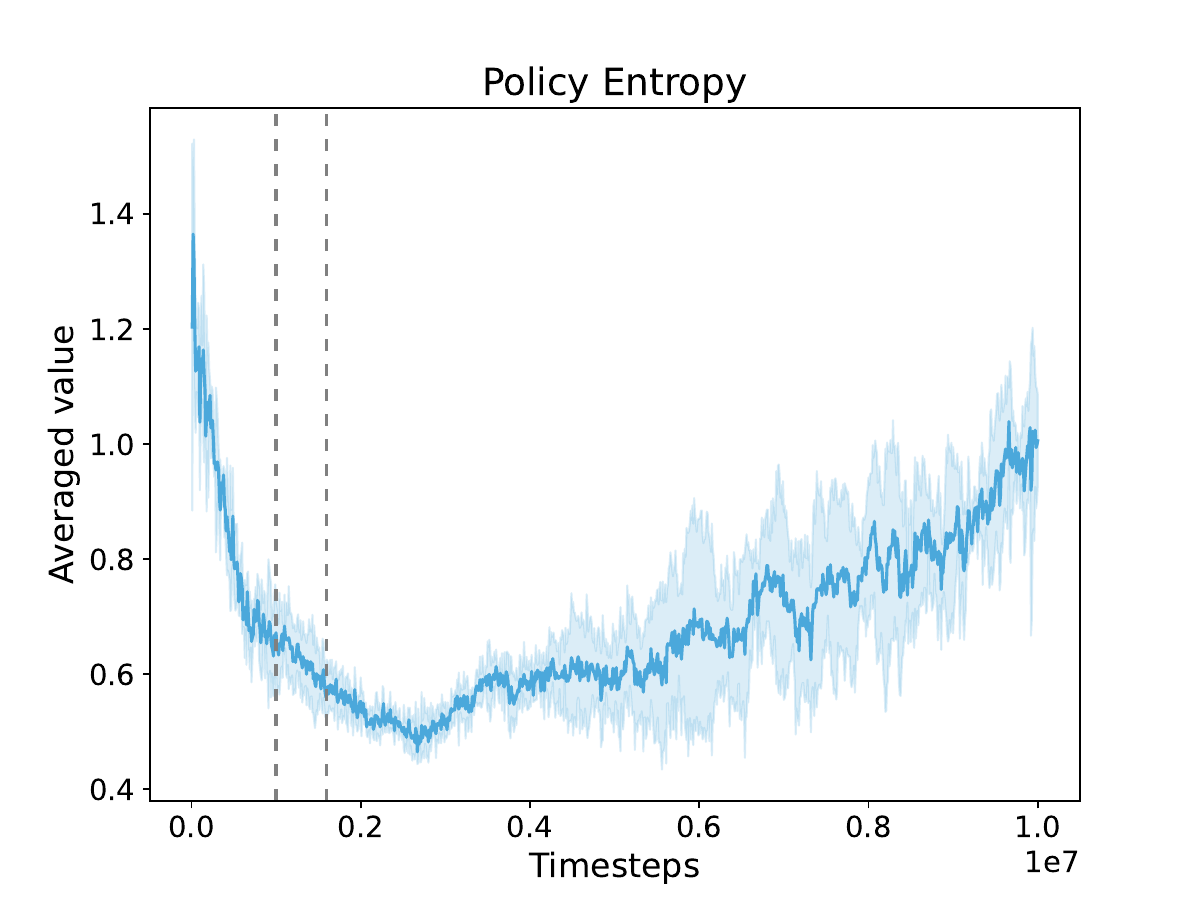}
		}
        \\
  	\subfigure[Episode Length]{
		\includegraphics[width=0.3\textwidth]{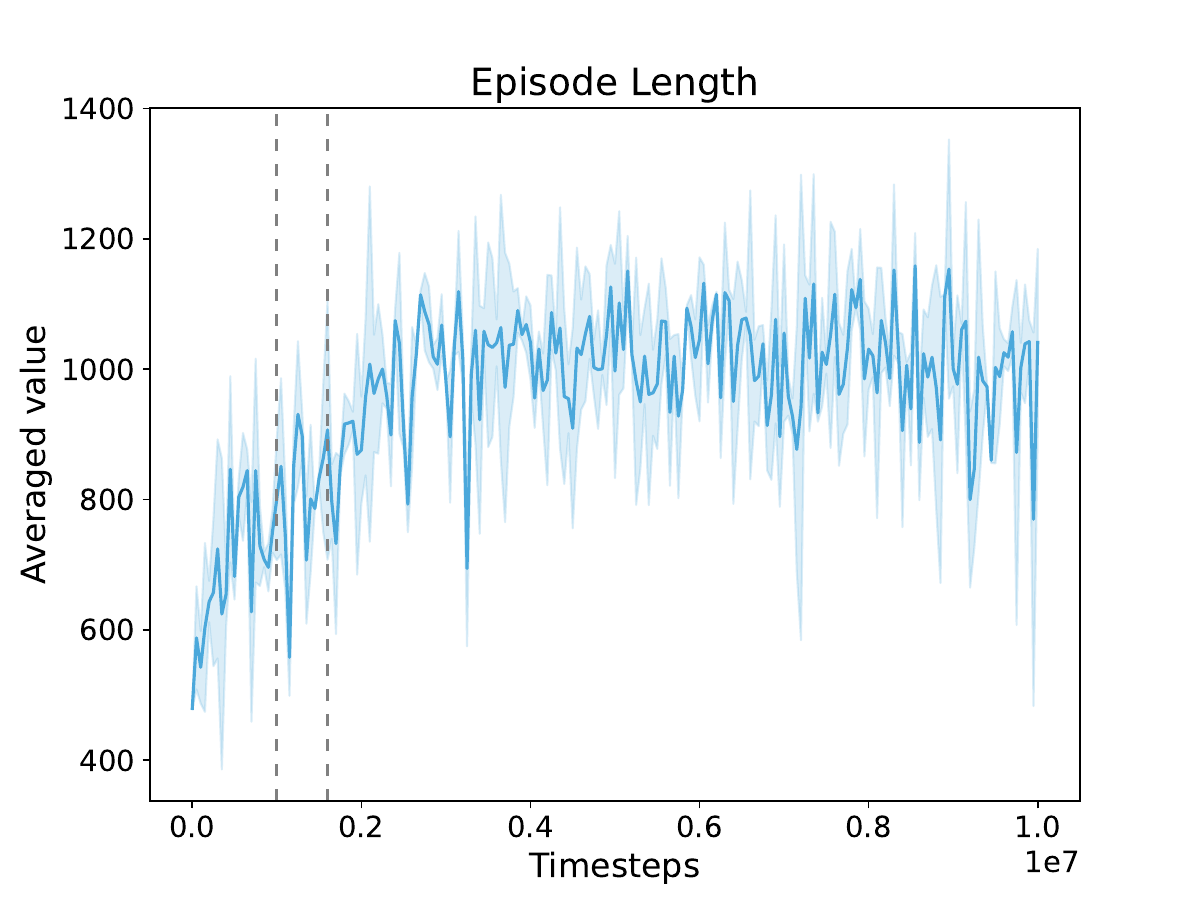}
		}
        \subfigure[Steps with Rewards]{
		\includegraphics[width=0.3\textwidth]{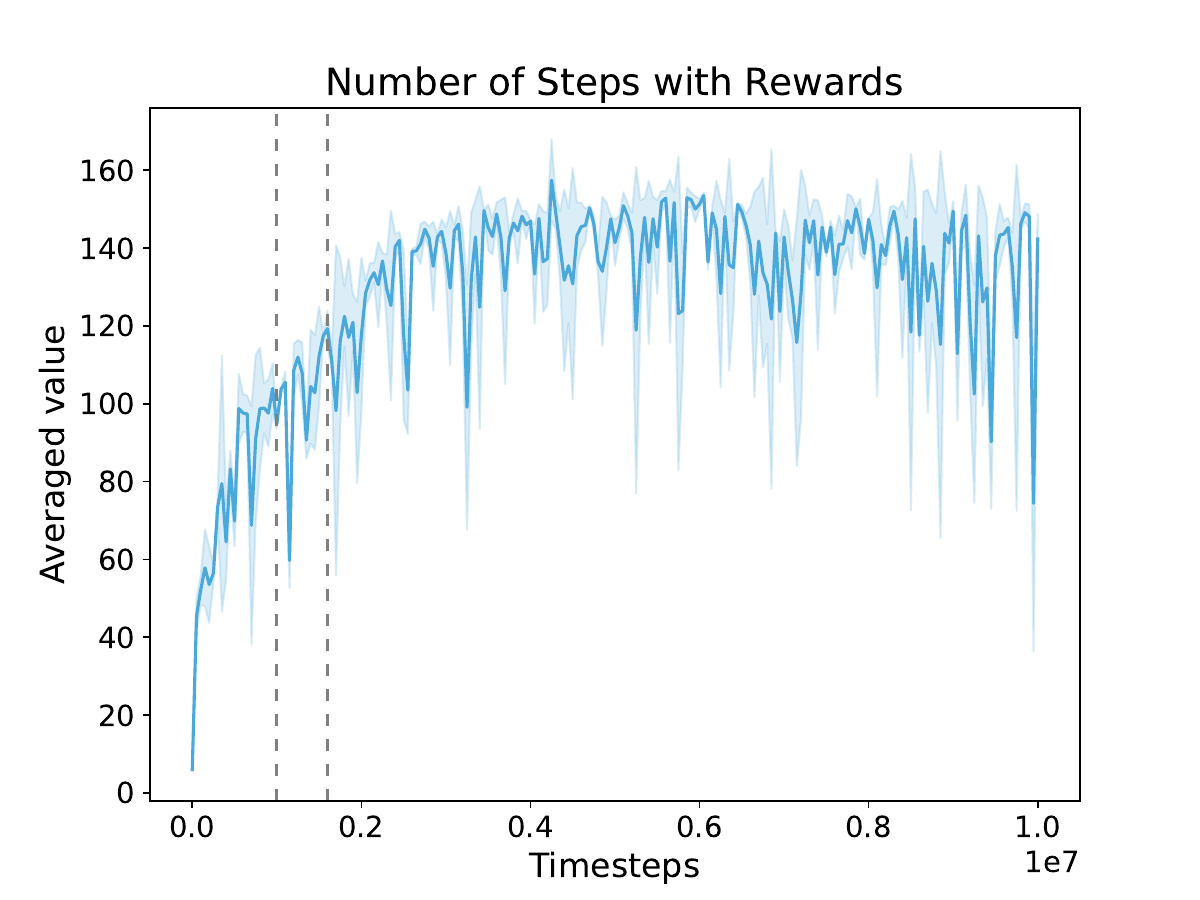}
		}
        \subfigure[Score]{
		\includegraphics[width=0.3\textwidth]{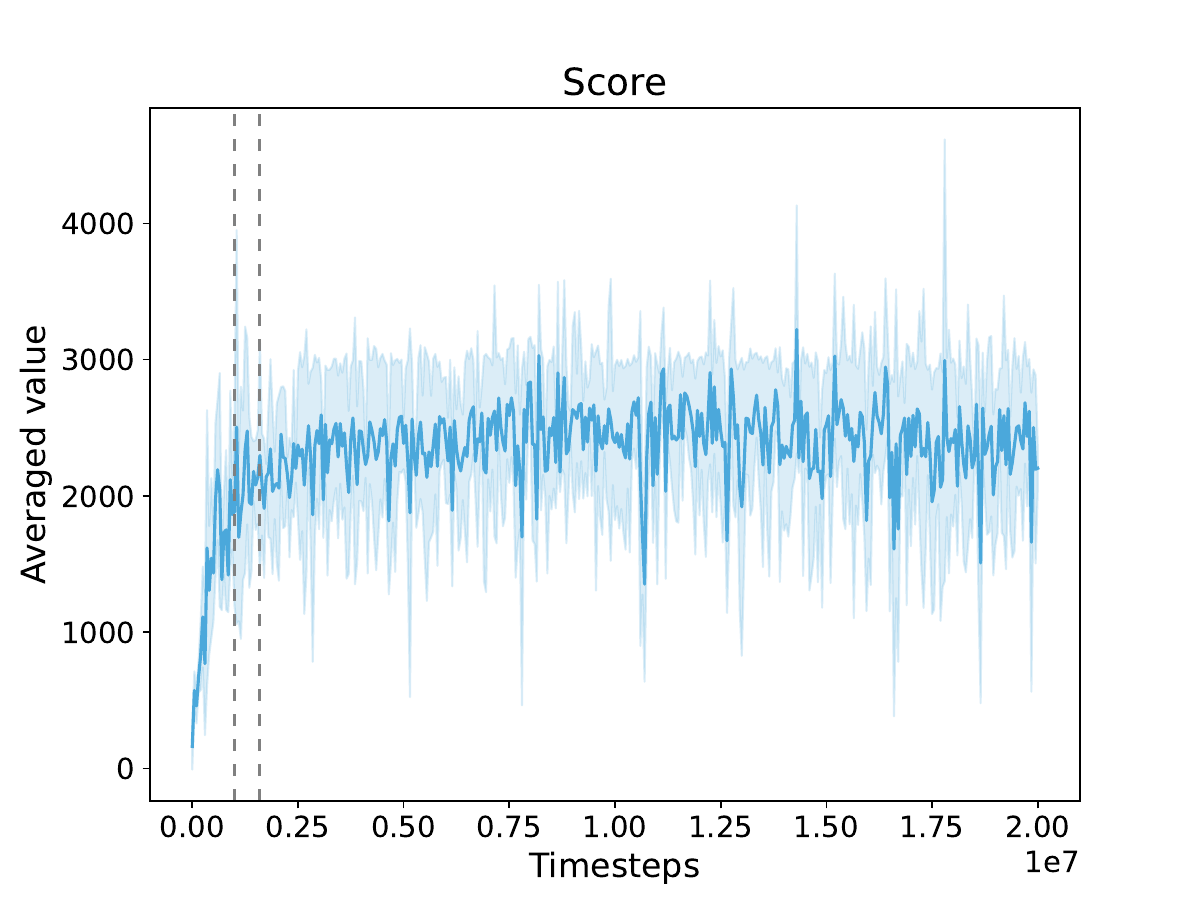}
		}
	\caption{Plots of Q function variance, estimation of Q-value, policy action entropy, episode length, number of steps with rewards and score on Atari Game MsPacman environment with discrete SAC over 10 million time steps.}
	\label{fig_example_mspacman} 
\end{figure}

We present the training process in the three aforementioned environments with deceptive rewards in Fig. \ref{fig_example_assault}-\ref{fig_example_mspacman}. It can be observed that in each case, deceptive rewards cause a rapid increase in Q variance and a decrease in policy entropy, leading the training process to fall into local optima.

\subsubsection{Comparison of Different Algorithms on Additional Atari Environments}

\begin{figure} [!t]
    \centering
    \subfigure[Assault]{
        \begin{minipage}[b]{0.3\textwidth}
	\includegraphics[width=0.98\textwidth]{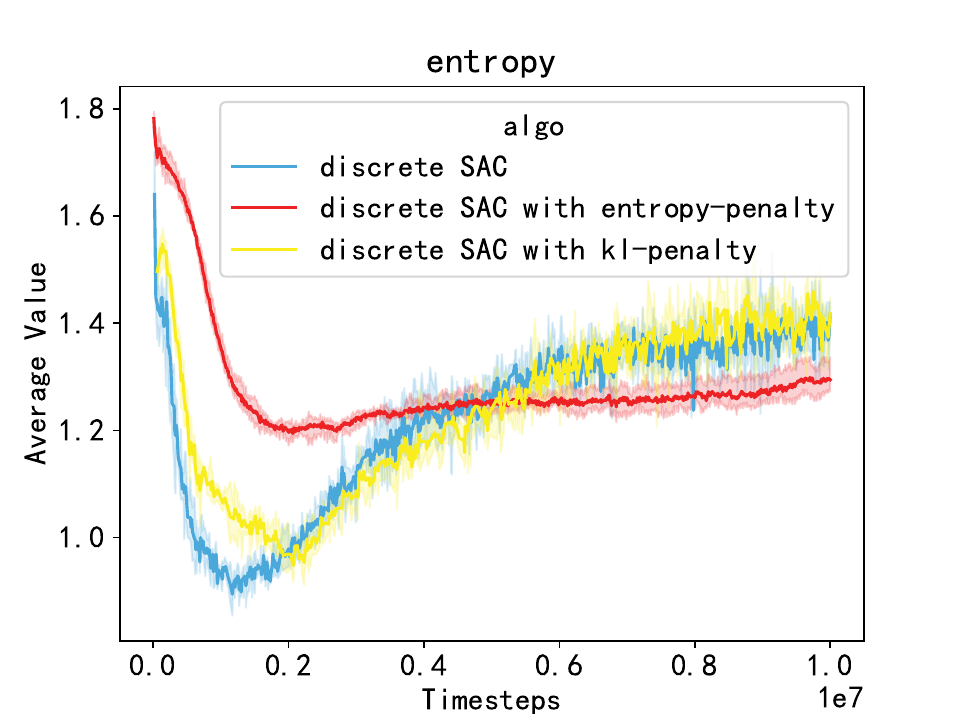} \\
        \includegraphics[width=0.98\textwidth]{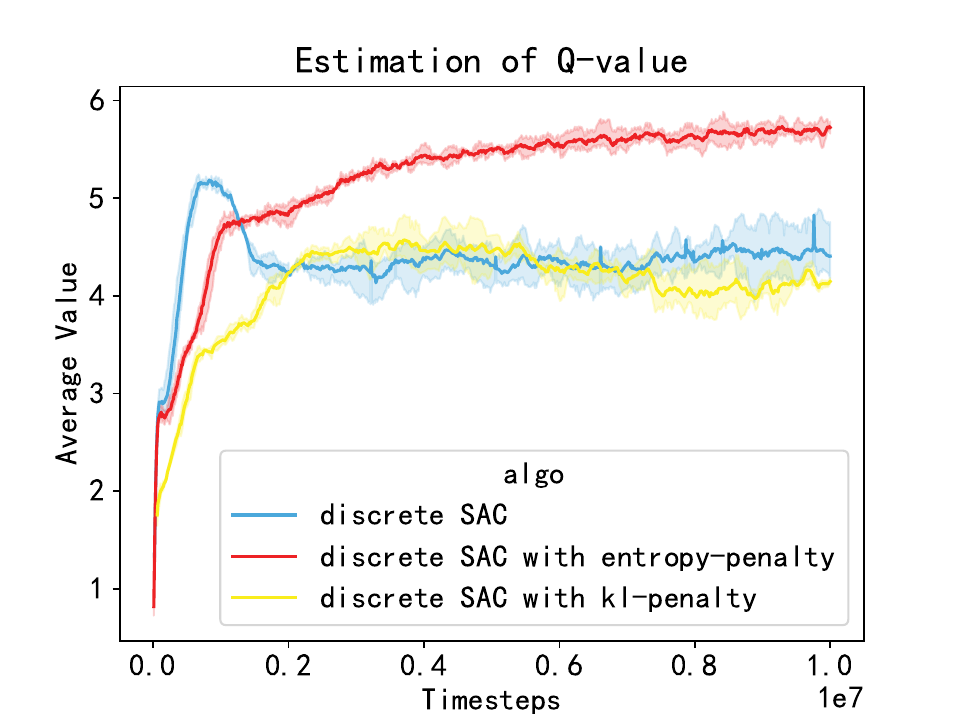} \\
        \includegraphics[width=0.98\textwidth]{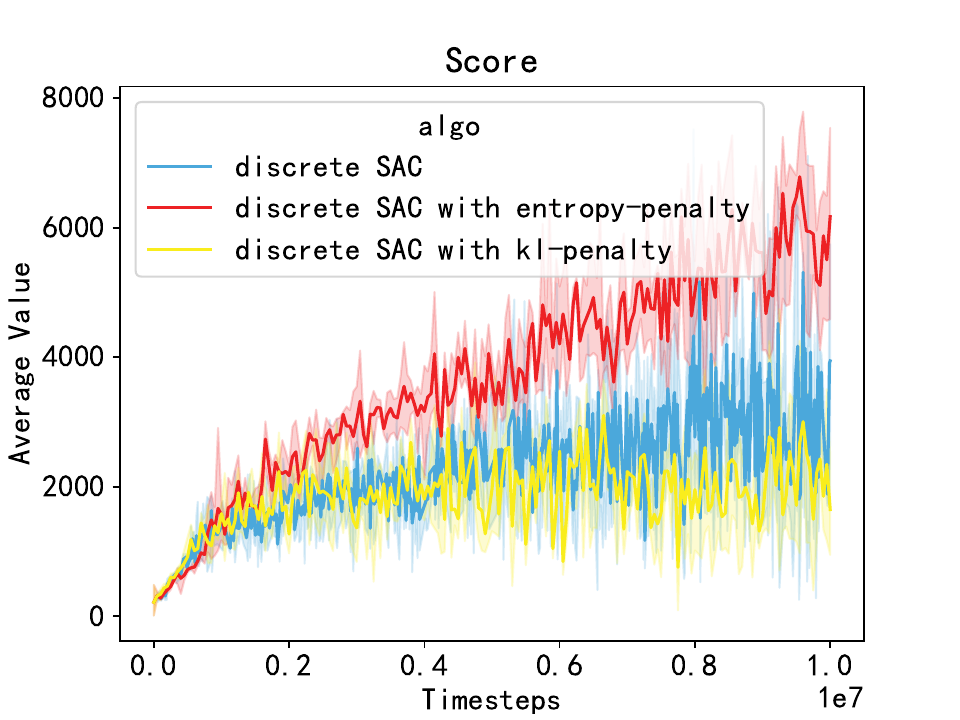}
        \end{minipage}
    }
    \subfigure[Jamesbond]{
        \begin{minipage}[b]{0.3\textwidth}
	\includegraphics[width=0.98\textwidth]{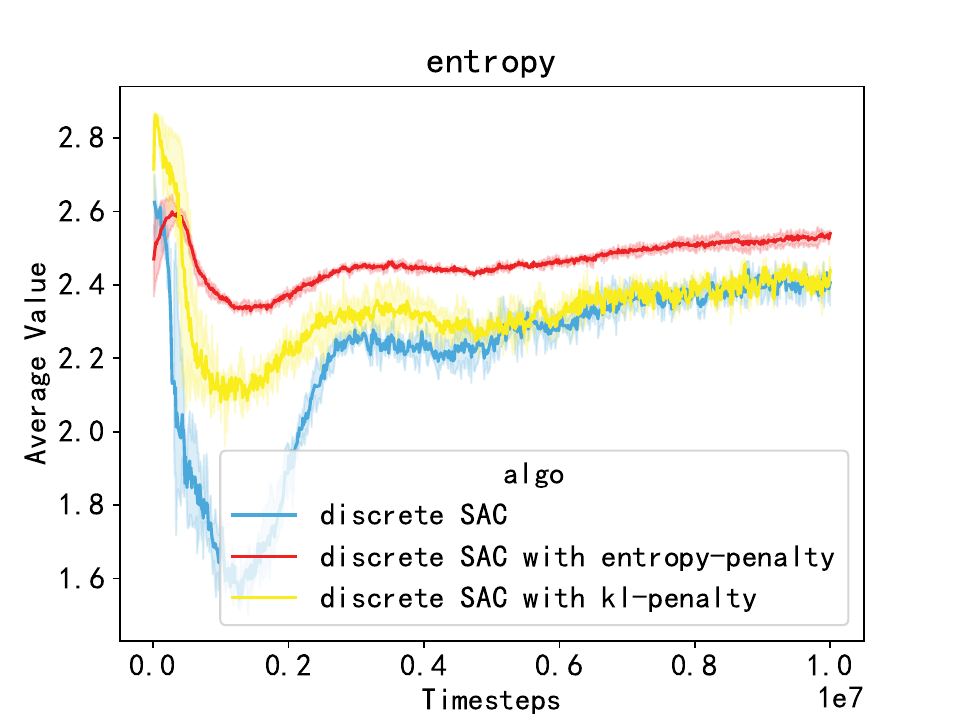} \\
        \includegraphics[width=0.98\textwidth]{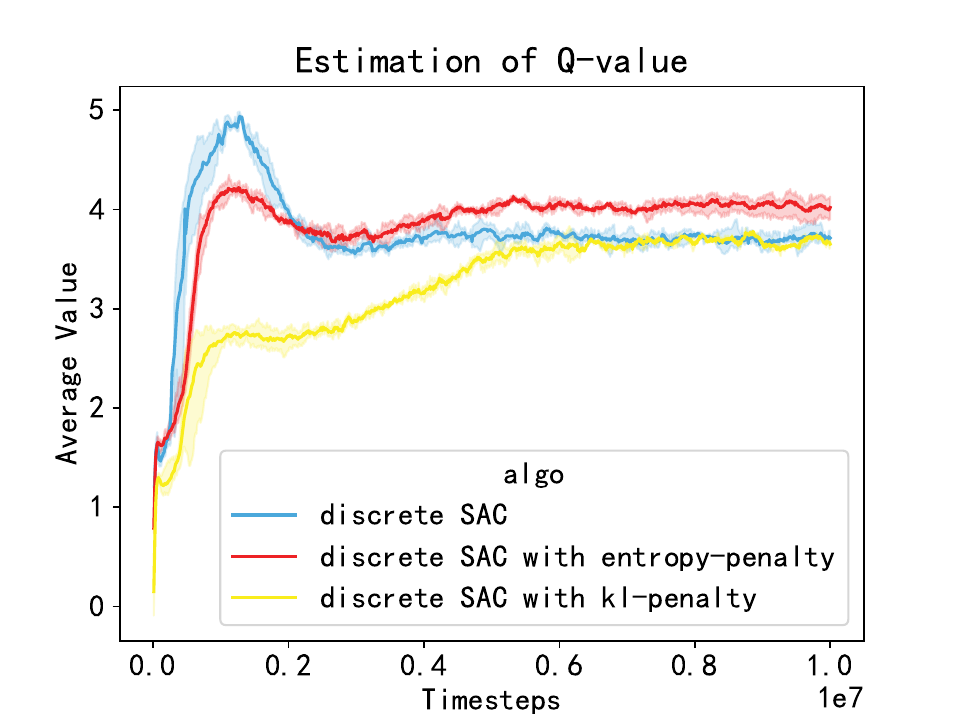} \\
        \includegraphics[width=0.98\textwidth]{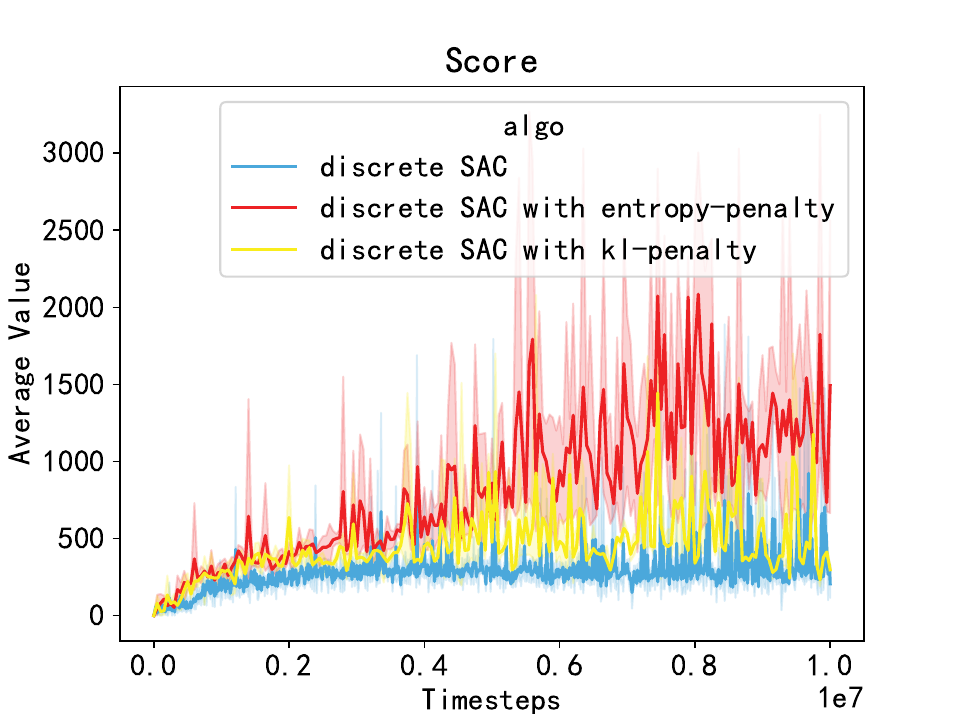}
        \end{minipage}
    }
    \subfigure[MsPacman]{
        \begin{minipage}[b]{0.3\textwidth}
	\includegraphics[width=0.98\textwidth]{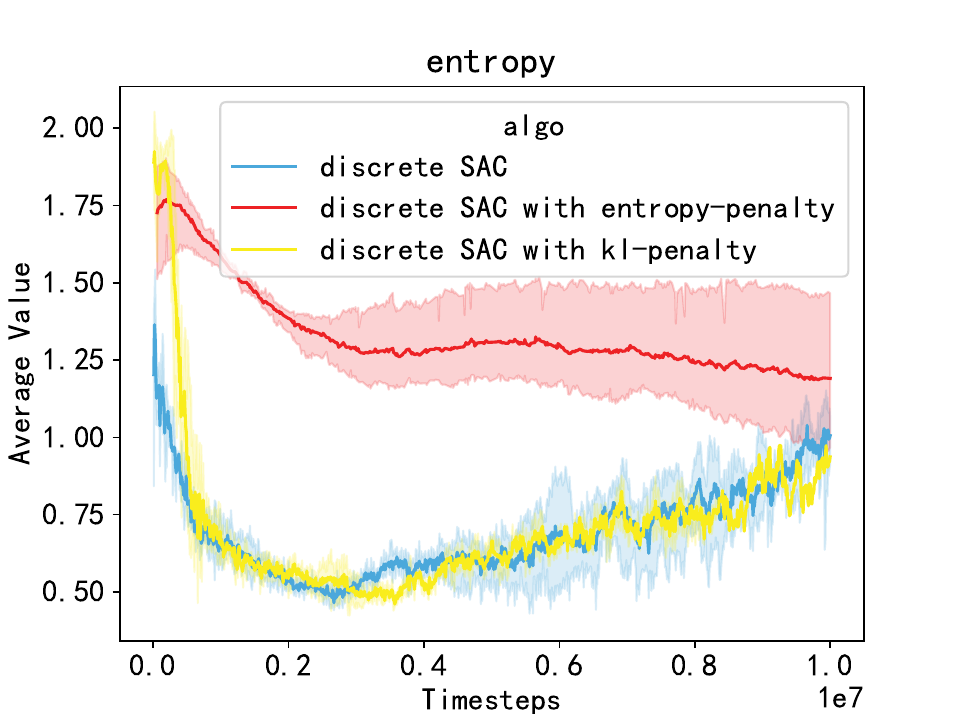} \\
        \includegraphics[width=0.98\textwidth]{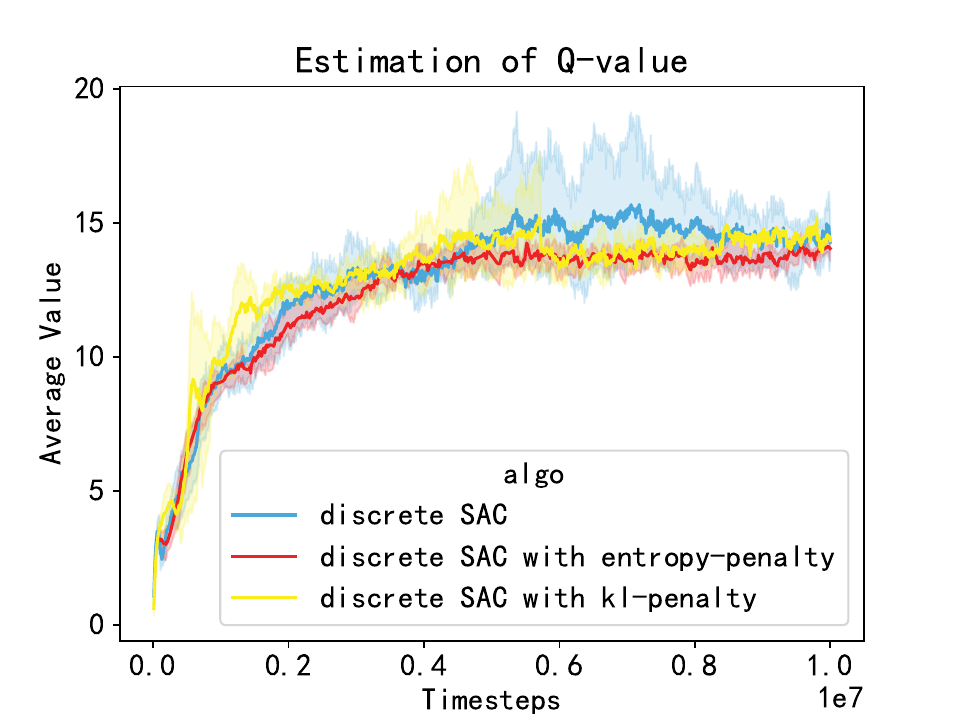} \\
        \includegraphics[width=0.98\textwidth]{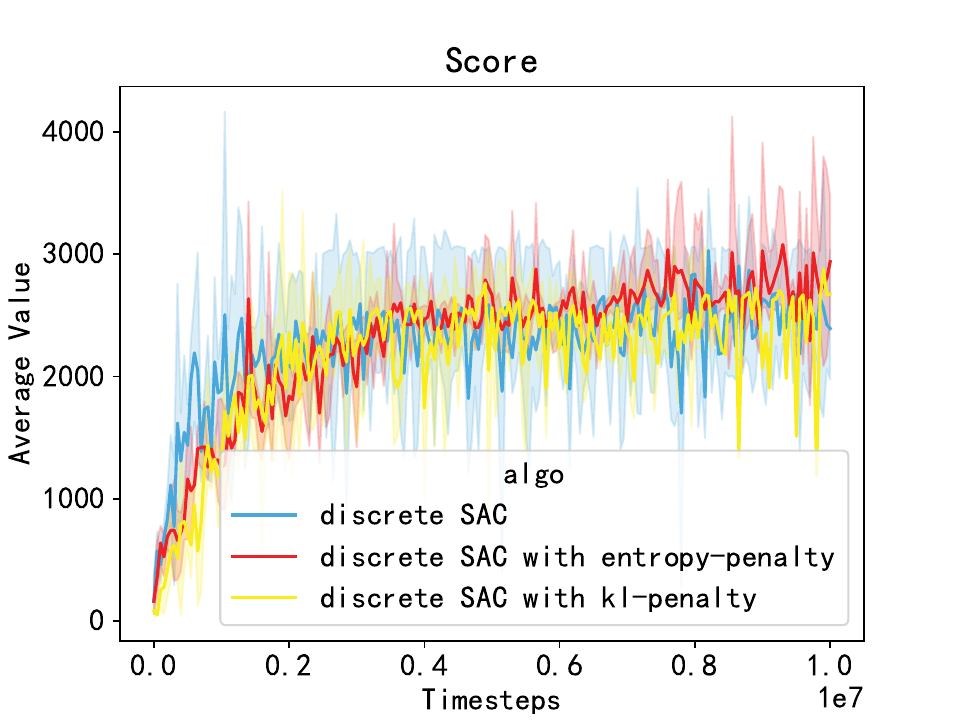}
        \end{minipage}
    }
\caption{Measuring Q function variance, policy action entropy, estimation of Q-value, and score on Atari game Assault, Jamesbond and MsPacman, comparing between discrete SAC, discrete SAC with KL-penalty and discrete SAC with entropy-penalty over 10 million time steps.}
\label{fig_comparison_additional}
\end{figure}

Here in Figure \ref{fig_comparison_additional} we provide comparative performance curves of DSAC, DSAC with entropy penalty, and DSAC with KL penalty in three additional Atari game environments: Assault, Jamesbond, and MsPacman. As shown in the results, the entropy penalty consistently offers the best early-stage regulation of entropy changes across all three environments. This regulation helps prevent the agent from falling into local optimum during the learning process, thereby improving the final score performance.

\section{Further Analysis}
\subsection{SAC Training Pattern on MuJoCo}
We only observe the failure modes in discrete SAC. The reason SAC does not exhibit these failure modes in continuous environments is twofold. First, SAC employs the reparameterization trick, fitting actions with a Gaussian distribution, allowing it to adapt to deceptive rewards without sacrificing policy diversity. Second, in continuous environments, actions that deviate slightly from the best response may have minimal impact on the outcome, whereas in discrete settings, different actions can have entirely distinct meanings. Therefore, our analysis primarily focuses on the challenges SAC faces in discrete environments.

\begin{figure} [!t]
    \centering
    \subfigure[Ant\label{fig_mujoco_ant}]{
        \begin{minipage}[b]{0.3\textwidth}
        \includegraphics[width=0.98\textwidth]{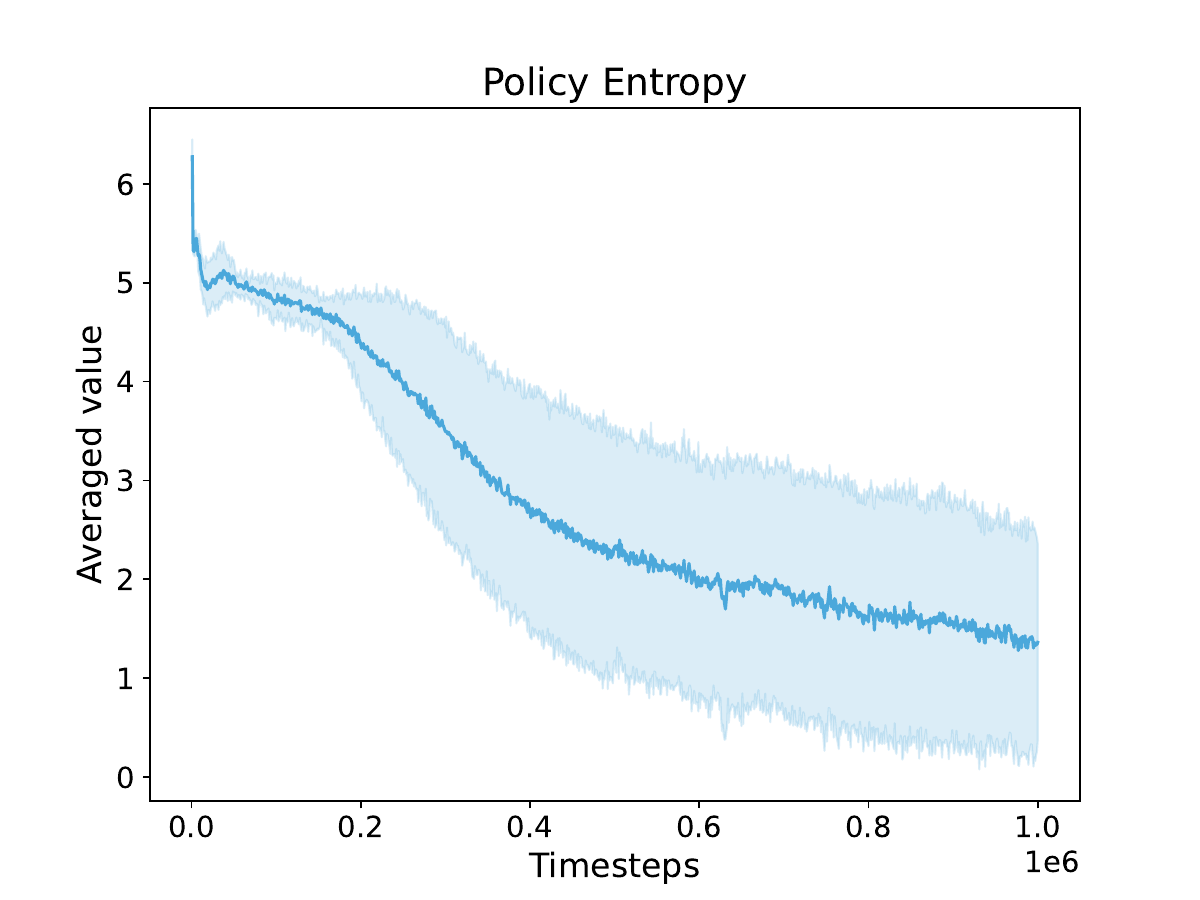}\\
	\includegraphics[width=0.98\textwidth]{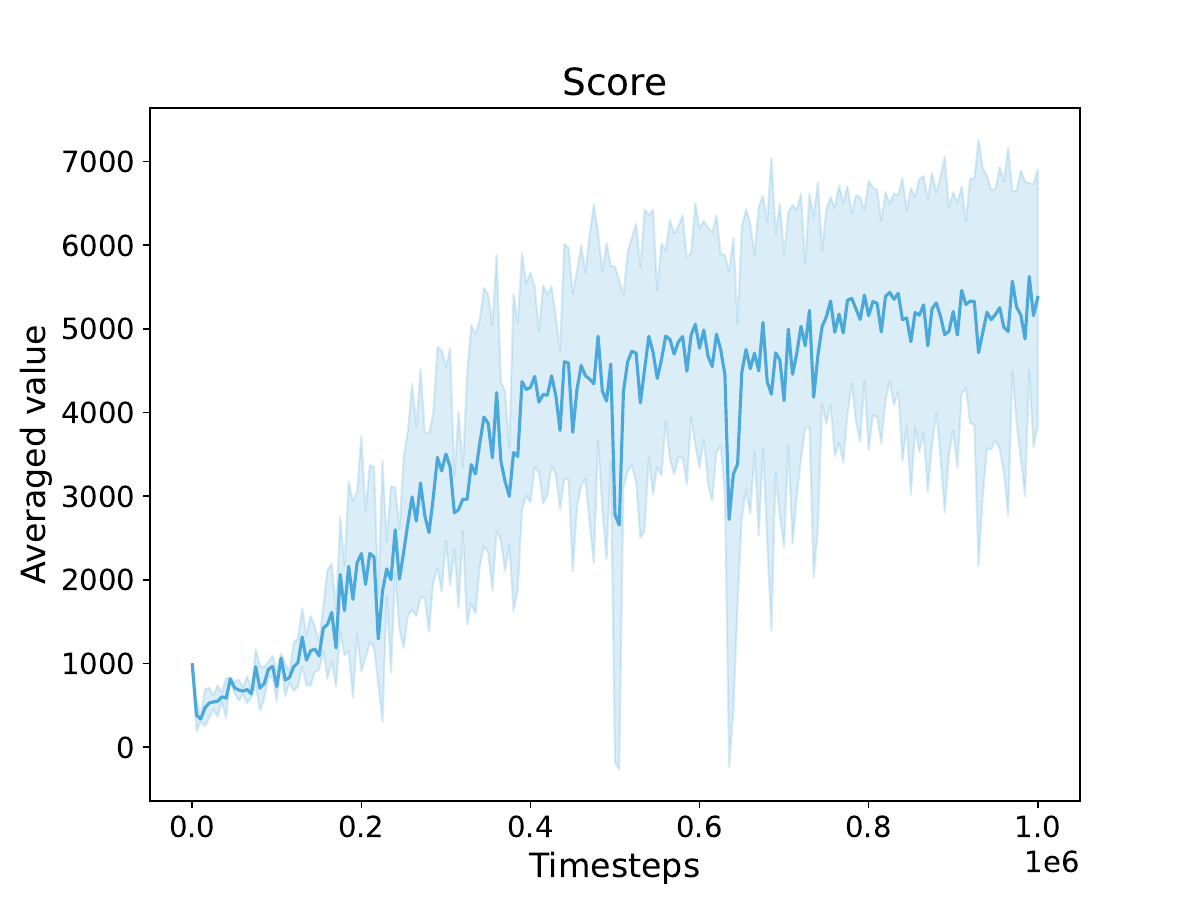}
        \end{minipage}
    }
    \subfigure[HalfCheetah\label{fig_mujoco_halfcheetah}]{
        \begin{minipage}[b]{0.3\textwidth}
        \includegraphics[width=0.98\textwidth]{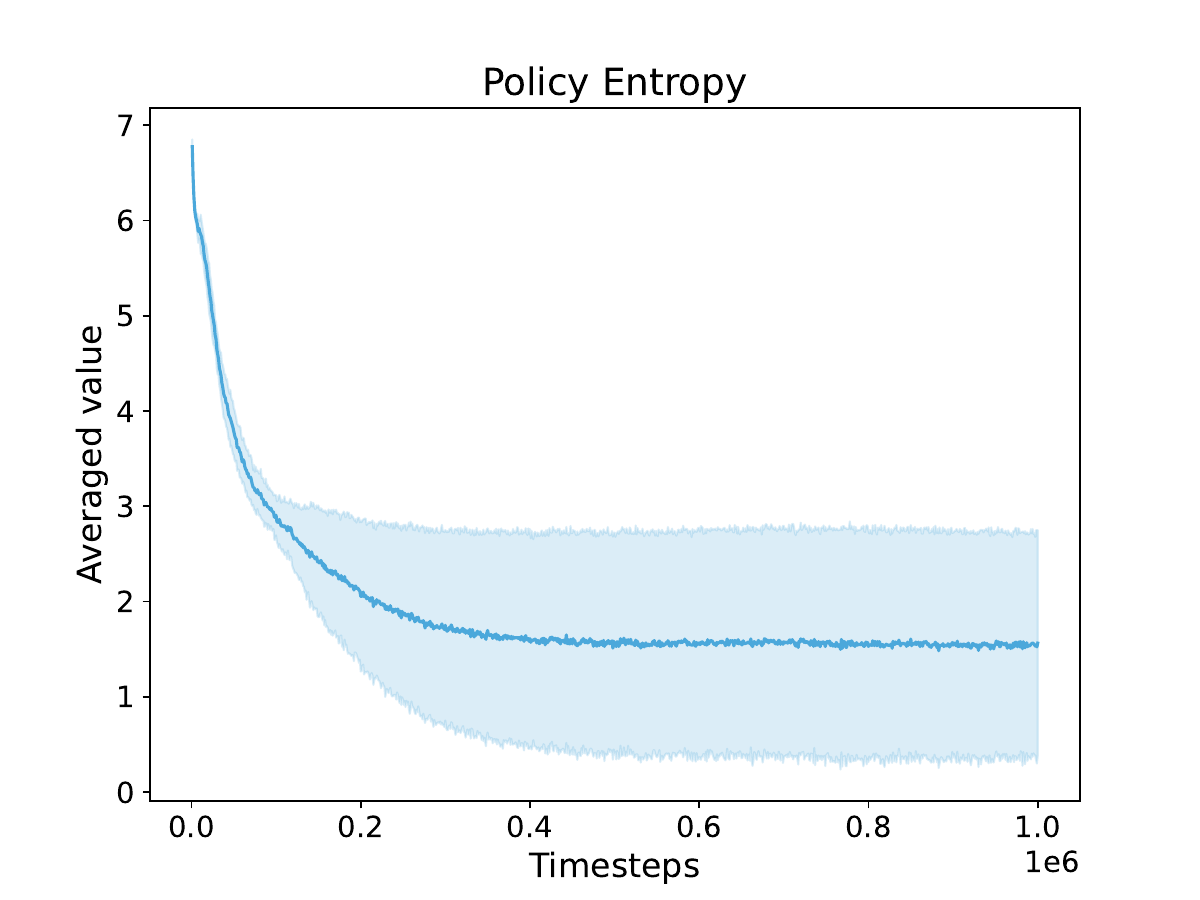}\\
	\includegraphics[width=0.98\textwidth]{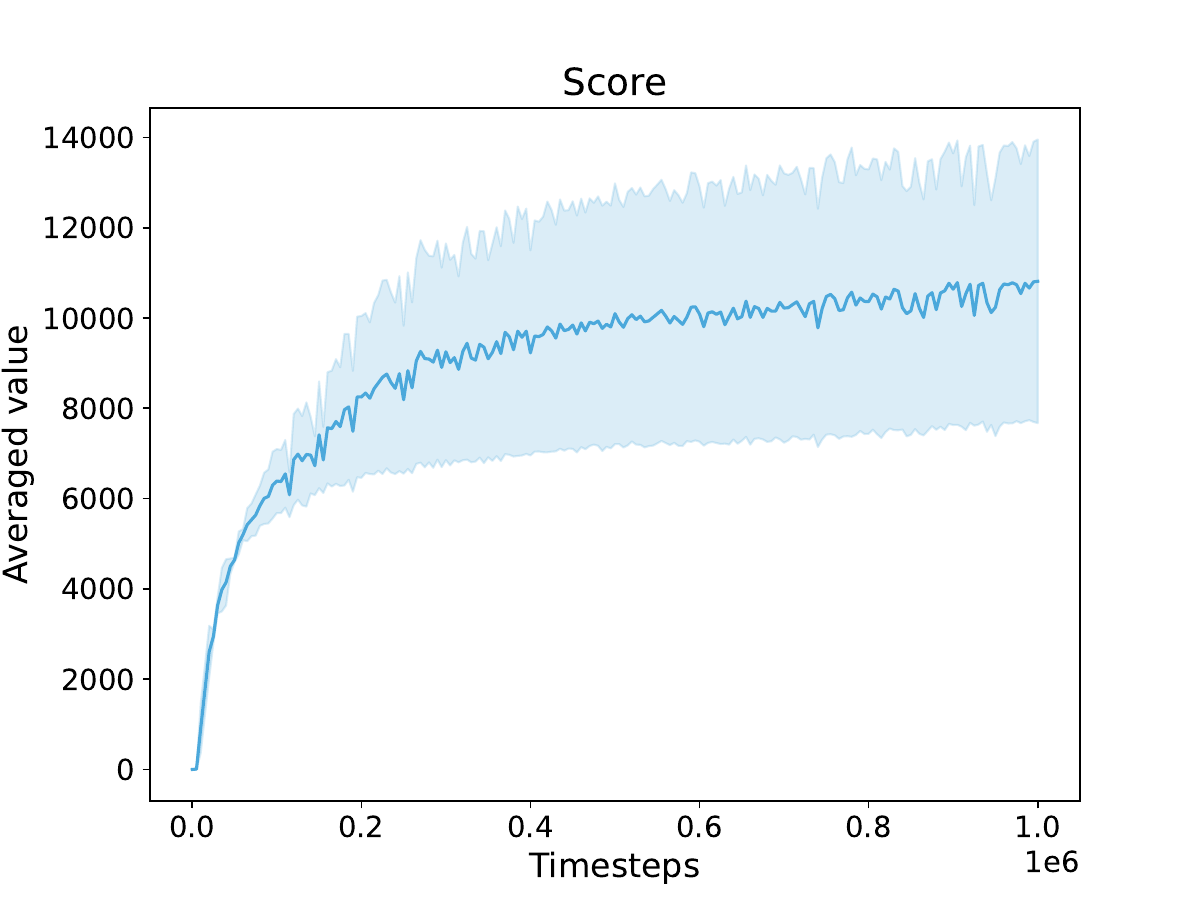}
        \end{minipage}
    }
    \subfigure[Walker2d\label{fig_mujoco_walker}]{
        \begin{minipage}[b]{0.3\textwidth}
        \includegraphics[width=0.98\textwidth]{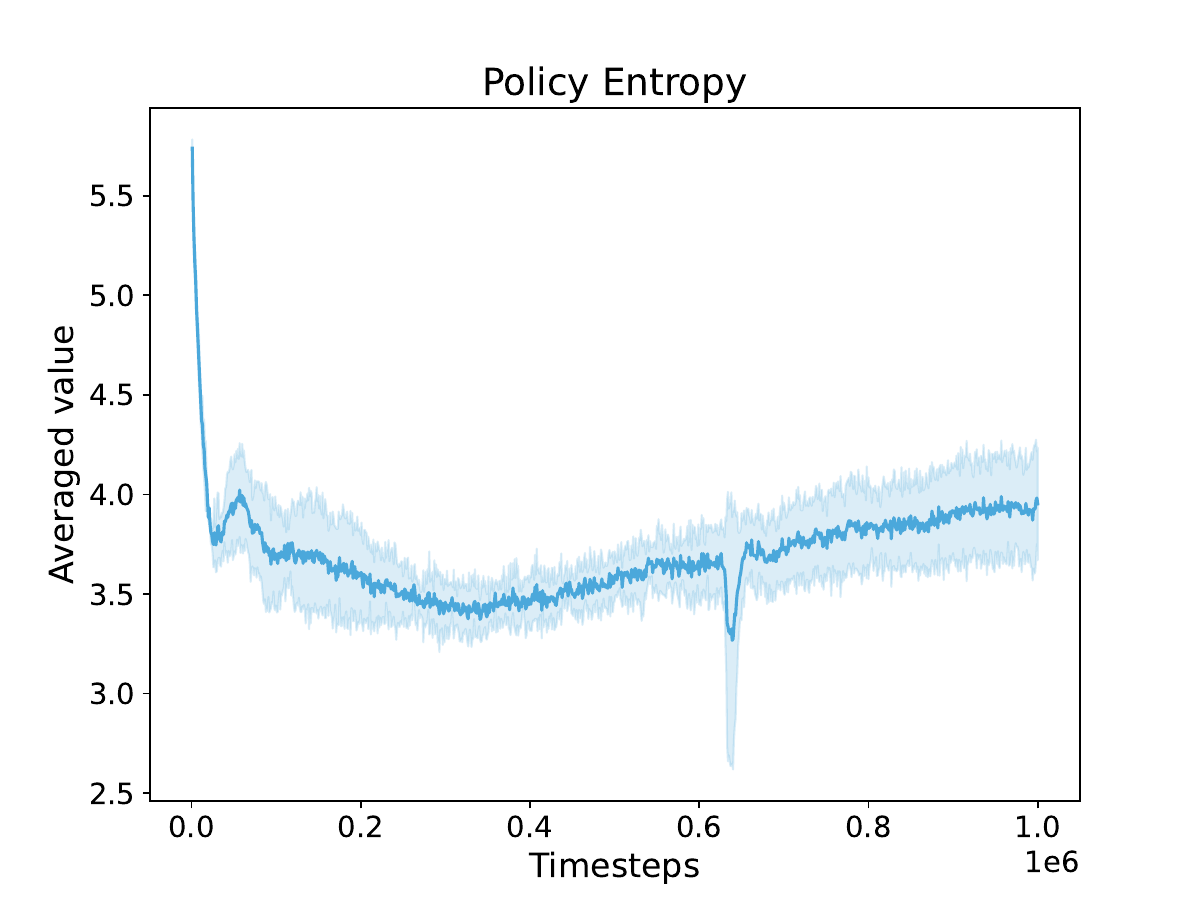}\\
	\includegraphics[width=0.98\textwidth]{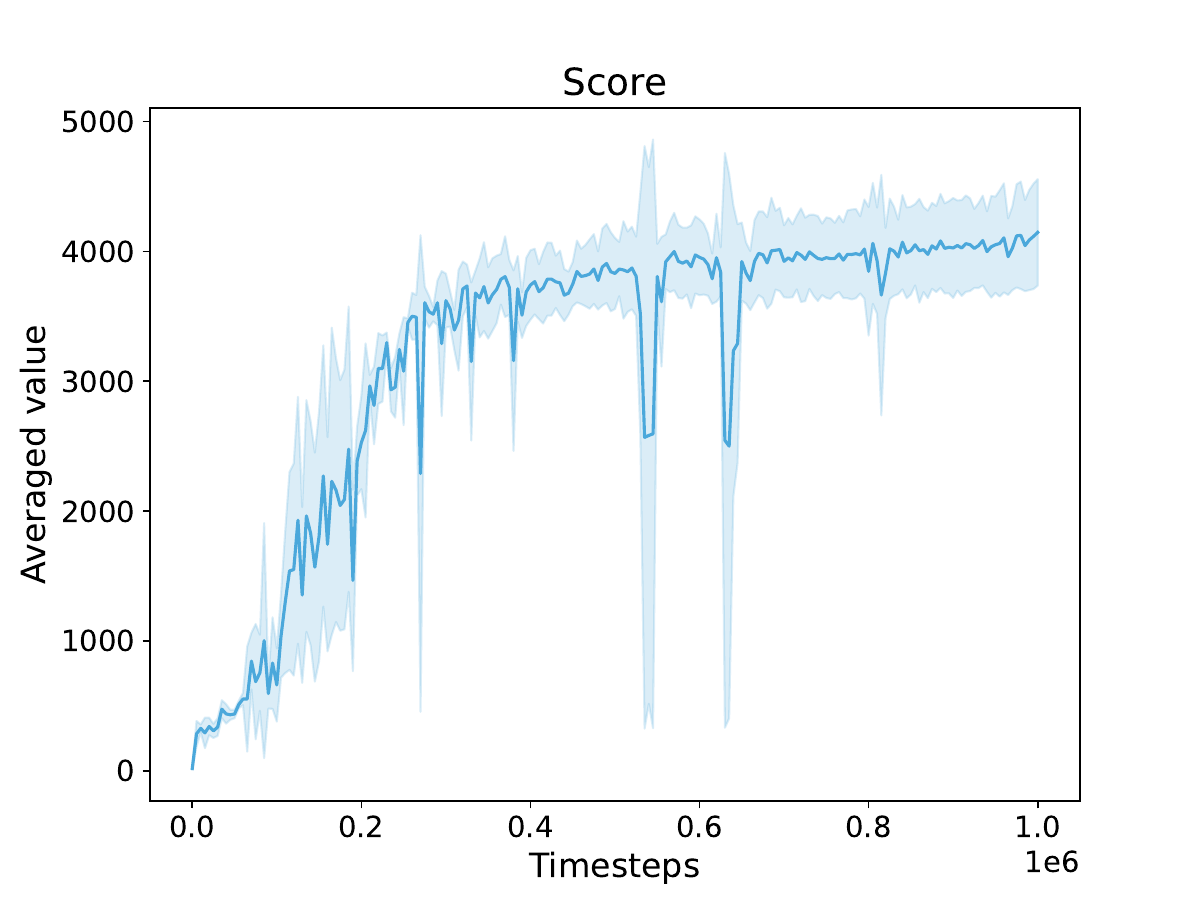}
        \end{minipage}
    }
\caption{The results of SAC in the MuJoCo environment.}
\label{fig_mujoco}
\end{figure}

To validate this point, we test SAC on three tasks of the MuJoCo environment. Results in Figure \ref{fig_mujoco} indicate that in MuJoCo, the SAC algorithm does not encounter local optimum issues; policy entropy changes are minimal and gradual, while the scores stadily increase. This suggests that SAC does not face the problems described in the paper when applied to continuous tasks.

\subsection{Hyperparameter Analysis}
\subsubsection{Different Hyperparameter Choices of DSAC}
\label{appendix_hyperparameter_analysis}
\begin{figure} [!t]
    \centering
	\subfigure[Entropy-penalty  $\beta$]{
		\includegraphics[width=0.48\textwidth]{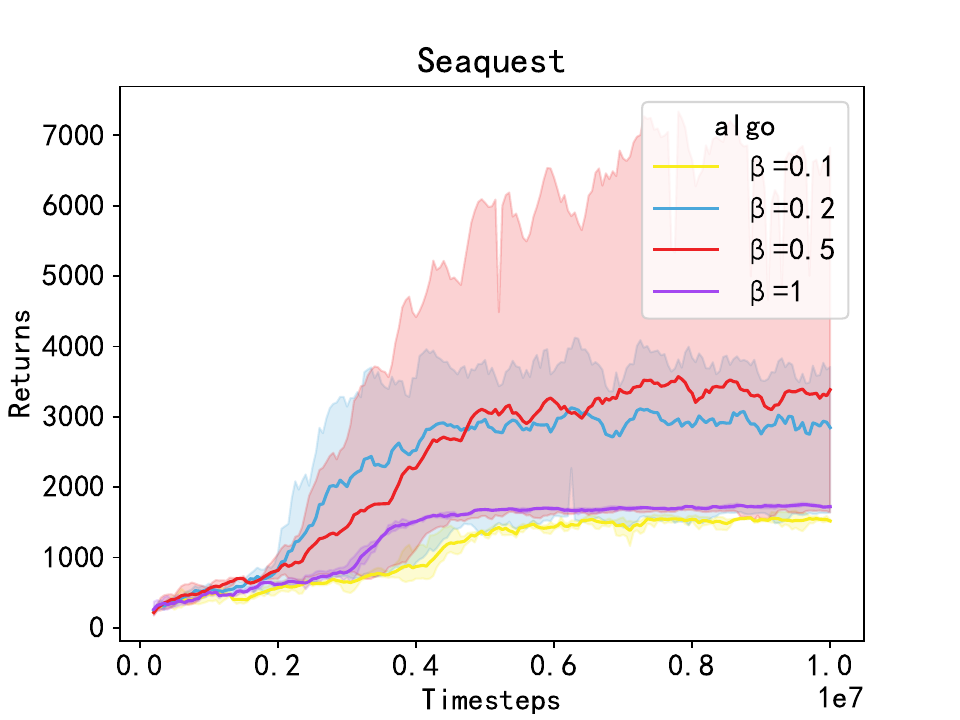}
		}
	\subfigure[Q-clip $c$]{
		\includegraphics[width=0.48\textwidth]{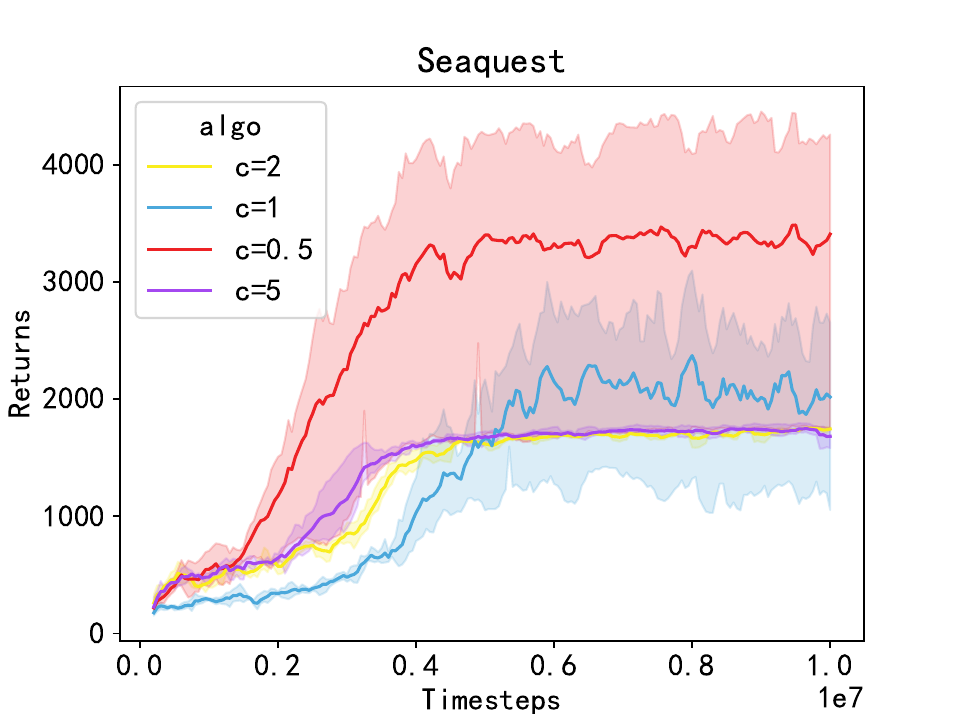}
		}

	\caption{Scores on Seaquest: a) variants entropy-penalty coefficient $\beta$ with 0.1, 0.2, 0.5 and 1. b) variants Q-clip $c$ with 0.5, 1, 2 and 5.}
	\label{fig_coefficient} 
\end{figure}

Our design method incorporates two hyperparameters, i.e., entropy-penalty coefficient $\beta$ and Q-clip range $c$. Fig.~\ref{fig_coefficient} compares various entropy-penalty coefficient $\beta$ and Q-clip range $c$ values. 
The constraint proportion of policy change is determined by the entropy-penalty coefficient $\beta$. Intuitively, an excessive penalty term will lead to policy under-optimization. 
We experiment with different $\beta$ in \{0.1, 0.2, 0.5, 1\}. 
We find that $\beta=0.5$ can effectively limit entropy randomness while improving performance.
The Q-clip constrains different ranges of Q value range $c$, and experiments with different ranges $c$ in \{0.5, 1, 2, 5\} show that 0.5 is a reasonable constraint value. 

\subsubsection{Different Choices of Clip Ratio}
\begin{figure} [ht]
    \centering

    \subfigure[Clip-Ratio]{
		\includegraphics[width=0.48\textwidth]{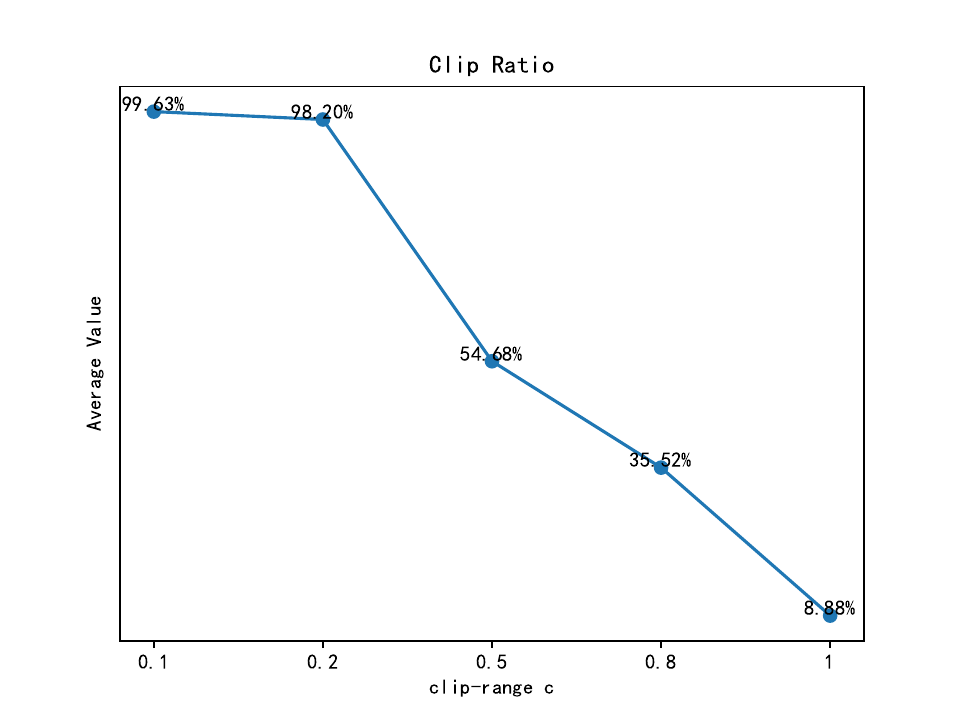}
		}
    \subfigure[Score]{
		\includegraphics[width=0.48\textwidth]{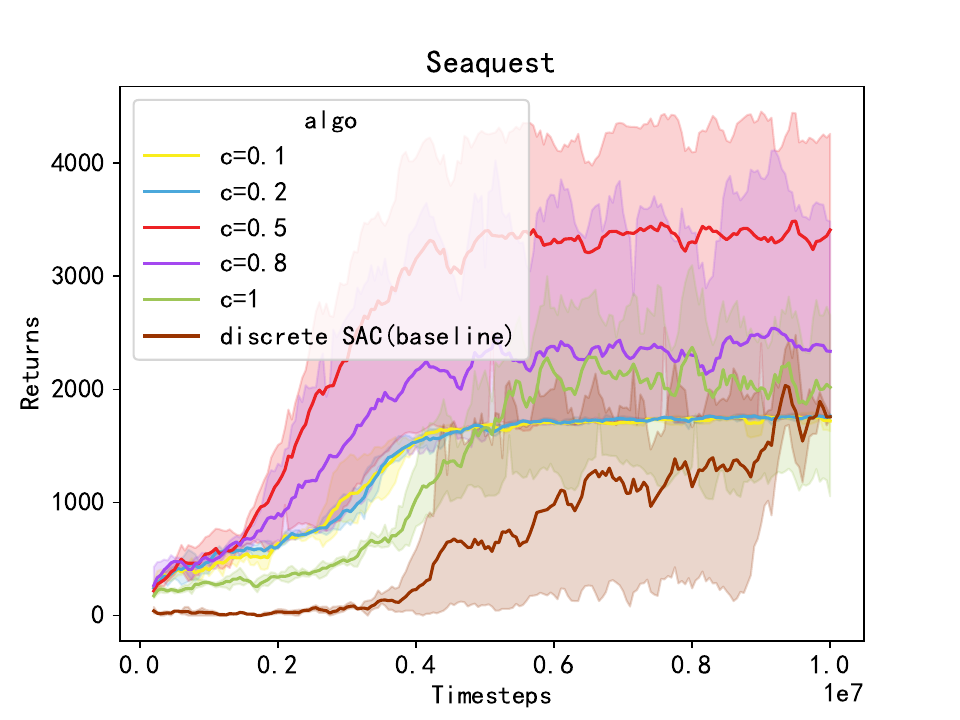}
		}
	\caption{Measuring  clip-ratio and score on Atari Game Seaquest environment with our method over 10 million time steps using variants Q-clip $c$ with 0.1, 0.2, 0.5, 0.8 and 1.0 .}
 \label{fig-c-clip-ration}
\end{figure}

In Figure \ref{fig-c-clip-ration}, we compare the clip ratio and final scores of different $c$ in our Q-clip.

\subsubsection{Various Learning Rates for Discrete SAC}
\begin{figure} [ht]
    \centering

    \subfigure[Entropy]{
		\includegraphics[width=0.31\textwidth]{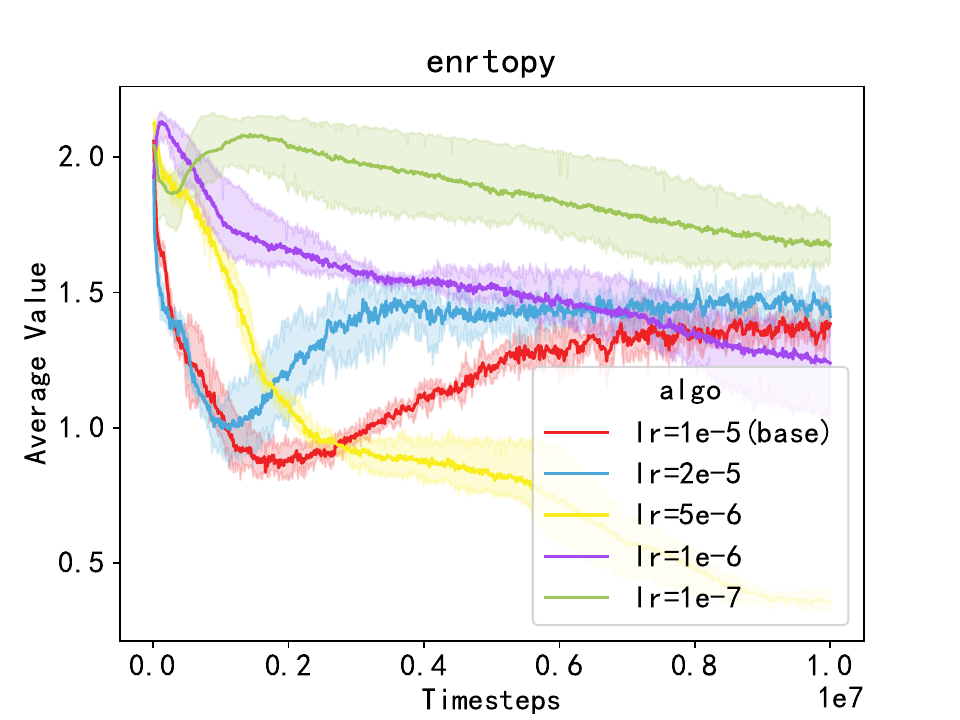}
        \label{subfig-learning-rate}
		}         
    \subfigure[Q-Value]{
		\includegraphics[width=0.31\textwidth]{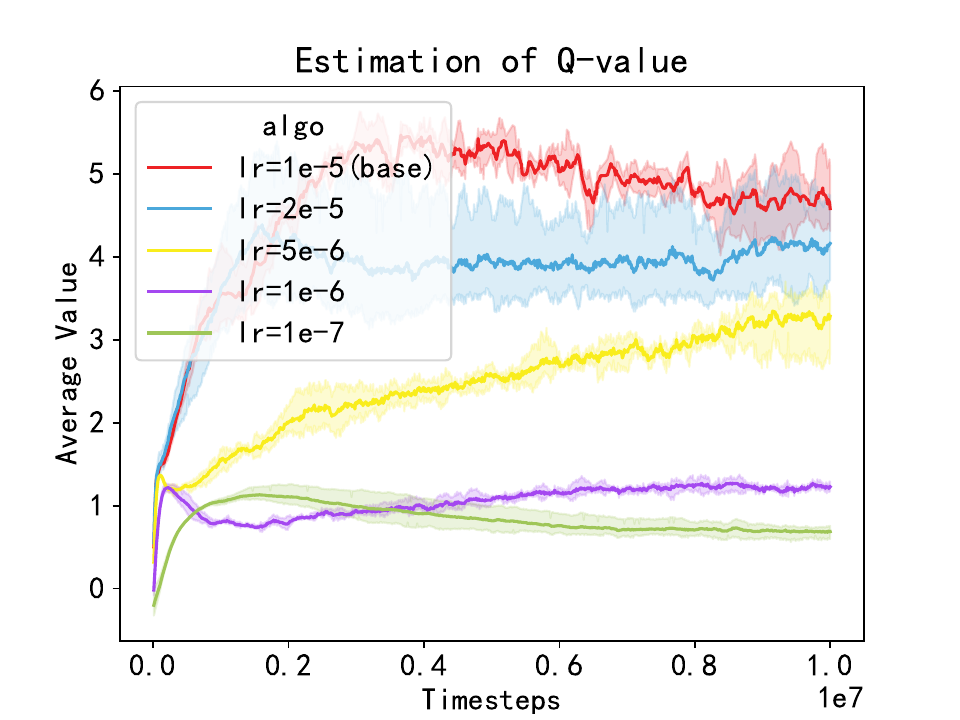}
		}
    \subfigure[Score]{
		\includegraphics[width=0.31\textwidth]{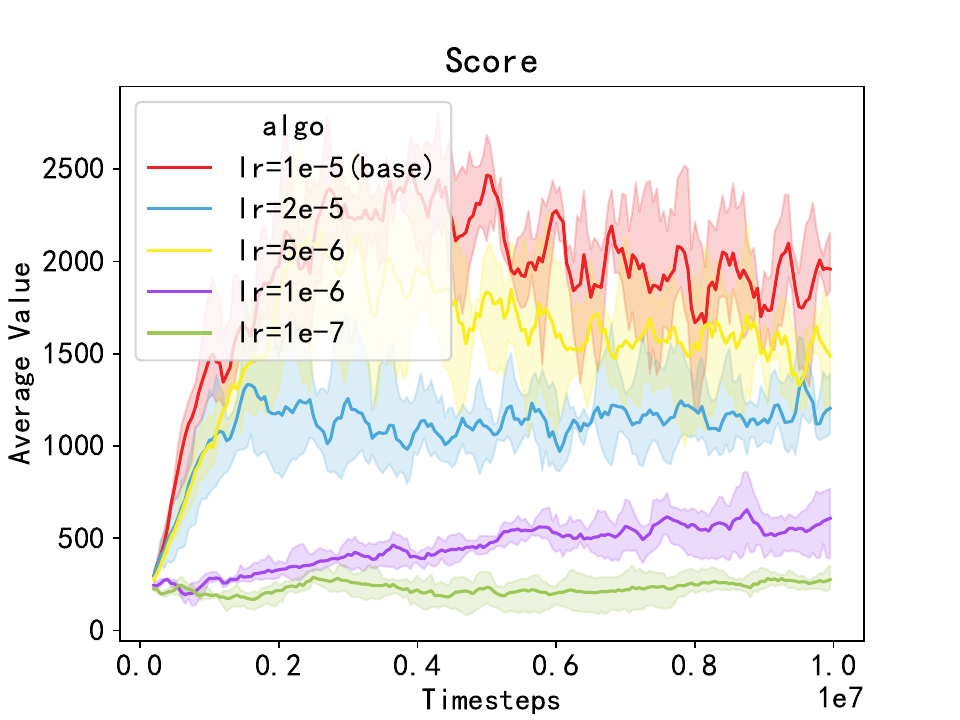}
		}
	\caption{Measuring  policy action entropy, estimation of Q-value and score on Atari Game  Asterix environment with discrete SAC  over 10 million time steps using different learning rates.}
 \label{fig-learning-rate}
\end{figure}

We introduce various learning rates for experiments on Asterix using vanilla discrete SAC in Fig. \ref{fig-learning-rate}. An excessively high learning rate leads to early convergence of entropy, while an excessively low learning rate results in insufficient optimization. The experiments show that the entropy instability issue of discrete SAC is not caused by inappropriate learning rate settings.

\subsubsection{Different Choices of Temperature $\alpha$ in Discrete SAC}
In Figure \ref{subfig_alpha}, we compare scores on Asterix by discrete SAC using variants $\alpha$ with 0.01, 0.025, 0.05, 0.075 and 0.1 over 10 million time steps.

% \begin{figure} [ht]
%     \centering

%     \subfigure[Score]{
% 		\includegraphics[width=0.50\textwidth]{images/experiments/rebuttal_alpha/Asterix_alpha_reward.pdf}
% 		}

% 	\caption{Measuring scores on Asterix by discrete SAC using variants $\alpha$ with 0.01, 0.025, 0.05, 0.075 and 0.1 over 10 million time steps.}
%  \label{fig-dsac-alpha}
% \end{figure}

\subsubsection{Comparison Across SD-SAC and DSAC}
\begin{figure} [htbp]
    \centering
	\subfigure[Discrete SAC with Different $\alpha$]{
		\includegraphics[width=0.45\textwidth]{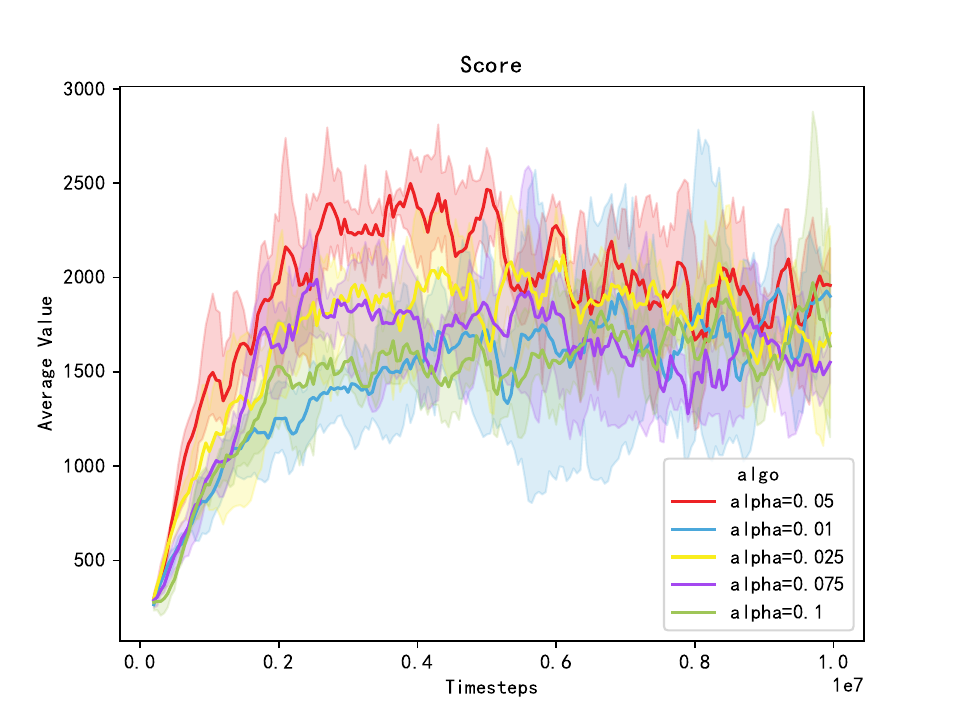}
		\label{subfig_alpha}
		}
	\subfigure[SD-SAC with Various $\beta$, and DSAC with the Best $\alpha$ and Learning Rate Combination.]{
		\includegraphics[width=0.45\textwidth]{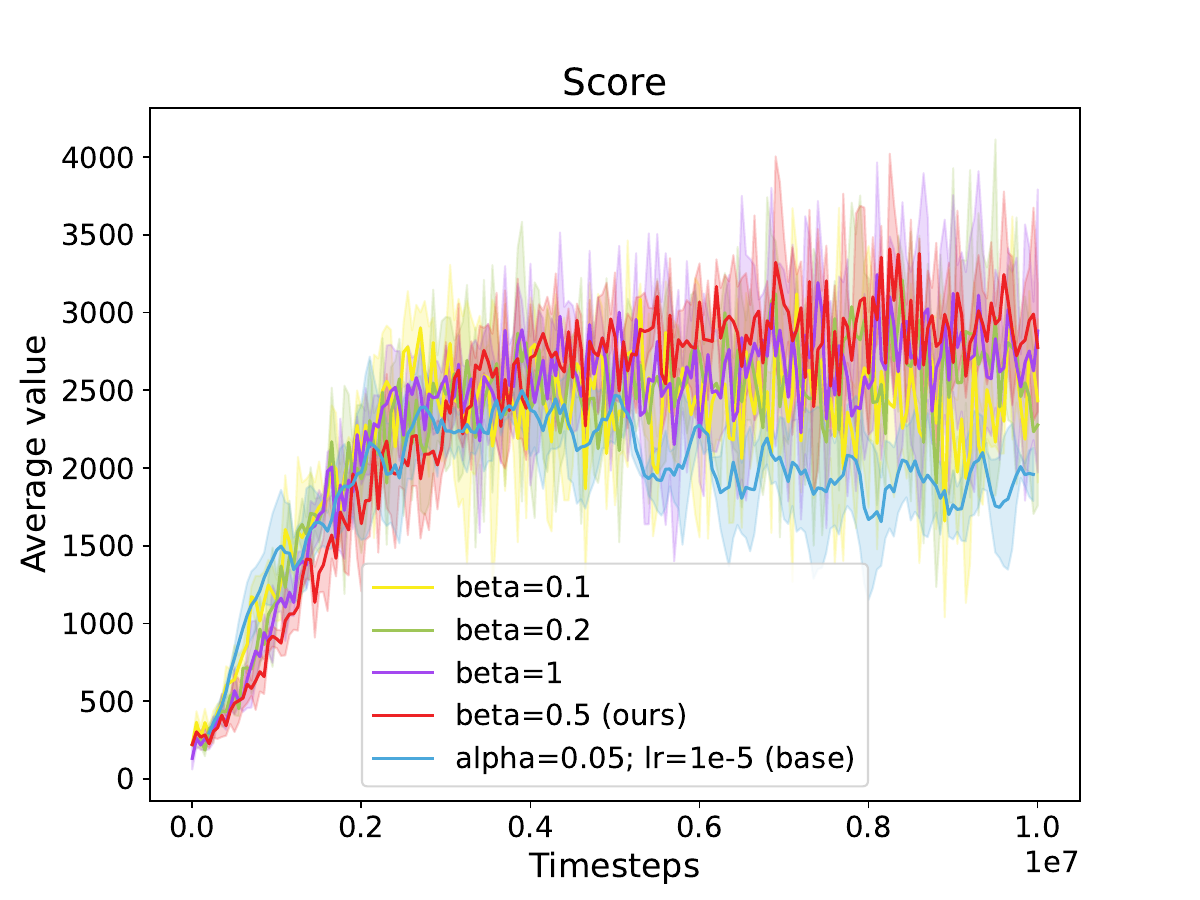}
		\label{subfig_beta}
		}
	\caption{Comparison of different hyperparameters, including $\alpha$, learning rate and $\beta$.}
	\label{fig_hyperparam_comparison} 
\end{figure}

We first determine discrete SAC's best combination of $\alpha$ (Fig. \ref{subfig_alpha}) and learning rate (Fig. \ref{subfig-learning-rate}). Then, we compare the scores between SD-SAC with different $\beta$, and DSAC with this combination ($\alpha = 0.5$; lr = 1e-5) in Figure \ref{subfig_beta}. The result shows that across various $\beta$ values, SD-SAC consistently outperforms DSAC. This indicates that the entropy penalty is a better and more balanced constraint than merely limiting the extent of policy updates.

% \begin{figure} [ht]
%     \centering
%     \subfigure[Score]{
% 		\includegraphics[width=0.50\textwidth]{images/experiments/rebuttal_hyperparam_alpha_beta/hyperparam_plot.pdf}
% 		}
% 	\caption{\textcolor{orange}{Comparing scores on Asterix between SDSAC using various $\beta$, and DSAC with the best $\alpha$ and $\alpha$ learning rate combination.}}
%  \label{fig-hyperparam_alpha_beta}
% \end{figure}

\subsection{Computation Overhead}

We test the computational speed on a machine equipped with an Intel(R) Xeon(R) Platinum 8255C CPU @ 2.50GHz with 24 cores and a single Tesla T4 GPU. The unit "it/s" represents the number of steps interacting with the environment per second. Detailed data are shown in the Table \ref{table-spped} below. The results demonstrate that our method has a 10.86\% reduction(265.41->236.58) in speed compared to the vanilla discrete SAC, while maintaining the same parameter size.
\begin{table}[h]
\centering
\caption{Computational speed our method and discrete SAC.}
\label{table-spped}
\begin{tabular}{c|c}
\toprule
         algorithm & speed \\ \hline
         discrete SAC                    				 &     265.41it/s \\
        discrete SAC + entropy-penalty           		  & 246.83it/s(-18.58) \\
        discrete SAC + avg-q + q-clip            		  & 250.27it/s(-15.14)  \\
        SD-SAC & 236.58it/s(-28.83)  \\
\bottomrule
\end{tabular}
\end{table}

\subsection{Cosine Similarity Comparison}
\begin{figure} [ht]
    \centering
    \subfigure[State Similarity]{
		\includegraphics[width=0.5\textwidth]{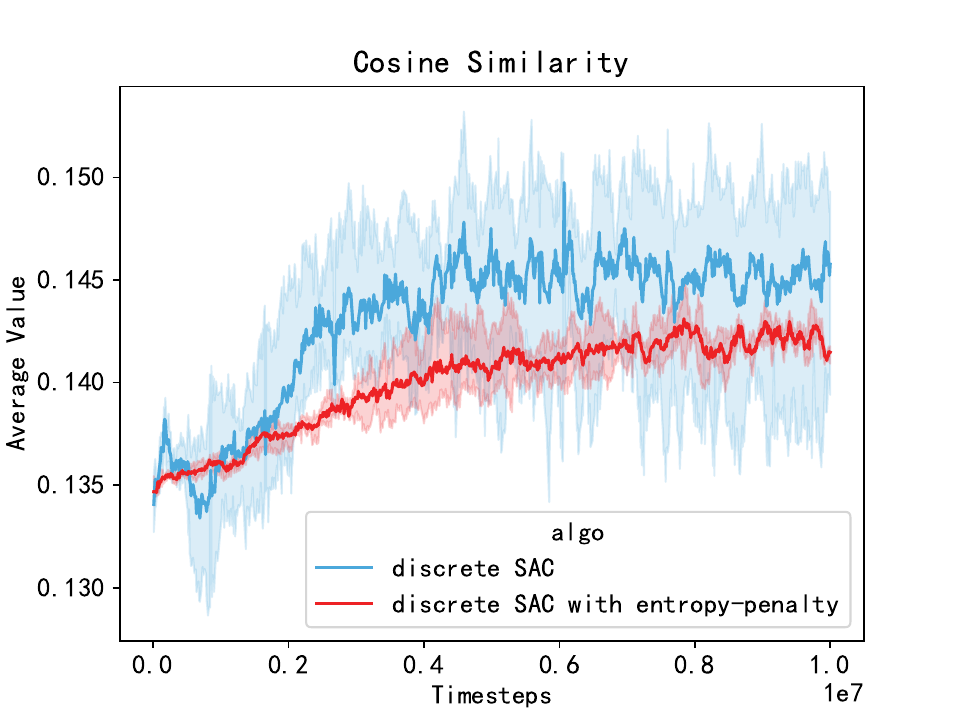}
		}
	\caption{Measuring cosine similarity of states on Atari game Asterix compared between discrete SAC and discrete SAC with entropy-penalty over 10 million time steps.}
 \label{fig-cos-simi}
\end{figure}

We visualize the changes in cosine similarity between adjacent states before and after incorporating the entropy penalty in the DSAC algorithm in Figure \ref{fig-cos-simi}. The results indicate that, following the addition of the entropy penalty, state transitions exhibit smaller and more stable changes. This observation further substantiates that the entropy penalty contributes to more stable policy updates, thereby enhancing the overall performance of the algorithm.

\bibliographystyle{tmlr}

\end{document}